%% file: main.tex
\PassOptionsToPackage{table}{xcolor}
\documentclass{article}
\usepackage{iclr2027_conference,times}

\usepackage{amsmath}
\usepackage{amssymb}
\usepackage{amsfonts}
\usepackage{algorithm}
\usepackage{algpseudocode}
\usepackage{array}
\usepackage{booktabs}
\usepackage{multirow}
\usepackage{enumitem}
\usepackage{nicefrac}
\usepackage{wrapfig}
\usepackage{graphicx}
\usepackage{subcaption}
\usepackage{tcolorbox}
\usepackage{float}
\usepackage{hyperref}
\usepackage{url}
\usepackage{cleveref}

\definecolor{ac}{HTML}{2A9D8F}
\definecolor{panelbg}{HTML}{F3FAF9}
\definecolor{paperfg}{HTML}{111111}
\definecolor{deltateal}{HTML}{69D9CC}
\definecolor{ducklingyellow}{HTML}{F4D35E}
\definecolor{softcyan}{HTML}{BFEAE5}
\definecolor{posdelta}{HTML}{16796F}
\definecolor{negdelta}{HTML}{B54747}
\floatstyle{plain}
\restylefloat{algorithm}
\newenvironment{algorithmpanel}{%
  \begin{tcolorbox}[
    colback=panelbg,
    colframe=ac,
    coltext=paperfg,
    boxrule=0.9pt,
    arc=8pt,
    left=0.35cm,
    right=0.35cm,
    top=0.28cm,
    bottom=0.28cm
  ]%
}{%
  \end{tcolorbox}%
}

\newcolumntype{C}{>{\centering\arraybackslash}p{0.95cm}}
\newtcolorbox{tealbox}[1][]{
  boxrule=0pt,
  colback=panelbg,
  coltext=paperfg,
  arc=8pt,
  left=0.45cm,
  right=0.45cm,
  top=0.35cm,
  bottom=0.35cm,
  #1
}

\title{Toward Open-World Video Segmentation over Long Horizons}

\author{Qing Su, Kaiyang Li, Yuan Zhuang, Fei Miao, and Shihao Ji\\
\normalfont University of Connecticut\\
\textbf{Code:} \href{https://github.com/QingSuML/savvy}{\nolinkurl{github.com/QingSuML/savvy}}}

\iclrfinalcopy
\hypersetup{
  hidelinks,
  pdftitle={Toward Open-World Video Segmentation over Long Horizons},
  pdfauthor={Qing Su, Kaiyang Li, Yuan Zhuang, Fei Miao, Shihao Ji},
  pdfsubject={Open-world video segmentation and granularity-agnostic evaluation}
}

\begin{document}

\maketitle
\lhead{Preprint}
\vspace{-\baselineskip}
\begin{abstract}
Long videos challenge open-world segmentation systems to continually discover objects and preserve their identities through disappearance and reappearance. We introduce \textbf{Savvy}, a zero-shot, semi-online, class-agnostic system for persistent object discovery and identity maintenance, and \textbf{OGA}, an evaluation suite that credits coherent part-level predictions when their granularity differs from the reference annotations. Savvy combines modular mask discovery, deferred admission based on accumulated evidence, and track consolidation to maintain an evolving object set. OGA allows multiple predictions to support one reference while preserving prediction IDs, pairing granularity-aware fidelity with diagnostics of identity bleeding and fragmented support. Across 142 ScanNet videos and 106 HM3D trajectories, evaluated using adaptively sampled prefixes capped at 1,500 source frames, Savvy outperforms DEVA+SAM and EntitySAM in VPQ$_\infty$, STQ, and AQ under both conventional and OGA evaluation. Conventional VPQ$_\infty$ reaches 18.44 versus DEVA+SAM's 11.52 on ScanNet and 17.31 versus 6.58 on HM3D. VIPSeg comparisons reveal conventional one-to-one VPQ's sensitivity to annotation granularity. Controlled identity-severing and flicker tests show that OGA detects temporal failures even when frame-level masks remain unchanged. Together, these results demonstrate complementary advances: Savvy improves long-horizon segmentation and association, and OGA distinguishes coherent part-level support from temporal identity failures.
\end{abstract}

\begingroup
\setlength{\textfloatsep}{10pt plus 2pt minus 2pt}
\setlength{\floatsep}{8pt plus 2pt minus 2pt}
\setlength{\intextsep}{6pt plus 2pt minus 2pt}
\captionsetup[figure]{skip=5pt}
\section{Introduction}
Promptable models such as SAM and SAM2~\citep{sam,sam2} can propose and propagate masks without first committing to a closed category vocabulary. We study the class-agnostic entity layer of \emph{open-world video segmentation} (\textbf{OVS}): the task supplies neither category names nor an initial object set. Objects must be discovered throughout a video and retain their identities across disappearance and later revisits. ScanNet exploration and HM3D looped trajectories expose these requirements through continuing object entry and repeated re-observation under changing viewpoints. Figure~\ref{fig:motivation_stats} contrasts ScanNet~\cite{scannet} with the shorter VIPSeg~\cite{vipseg} benchmark: ScanNet exhibits a heavier accumulation of seen objects and reappearance events. OVS therefore requires continuing discovery and re-association, not just propagation of an initialized object set.

Existing pipelines provide propagation and proposal fusion, but long videos also require explicit object-lifecycle decisions. DEVA~\cite{deva} fuses image proposals across short clips in its semi-online mode; EntitySAM~\cite{entitysam} emphasizes prompt refinement and discovers fewer objects than the reference trend in our long-horizon test (Figure~\ref{fig:motivation_stats}). Our focus is the lifecycle question beyond mask quality: when should a candidate be promoted or merged, and when should a returning track be accepted under its existing ID?

\begin{figure}[t]
    \centering
    \includegraphics[width=\textwidth]{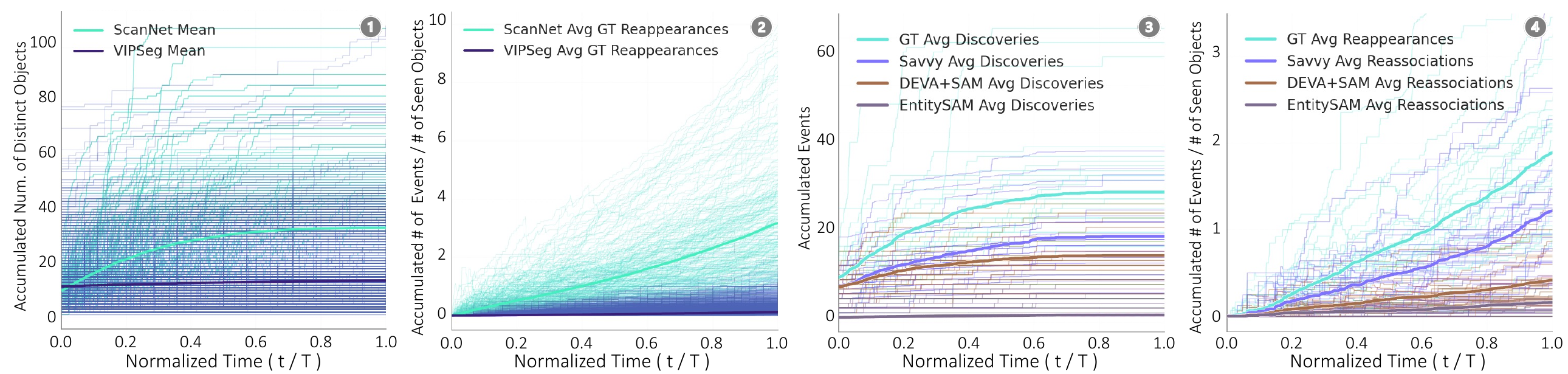}
\caption{\small \textbf{Continual discovery and re-association in long videos.}
\textbf{1:} Cumulative number of distinct objects in ScanNet and VIPSeg.
\textbf{2:} Cumulative reappearance events per seen object.
\textbf{3:} Cumulative object discoveries on ScanNet for Savvy, DEVA+SAM, and EntitySAM, compared with ground truth (GT).
\textbf{4:} Cumulative re-association events per seen object for the same methods, compared with GT reappearances.
All panels use normalized video time ($t/T$); thick curves show averages. ScanNet exhibits more object accumulation and reappearance than VIPSeg, while Savvy follows the GT discovery and reappearance trends more closely than either baseline.}
    \label{fig:motivation_stats}
\end{figure}

We introduce \textbf{Savvy}, which uses pretrained components without additional model training, to separate mask proposal, provisional admission, and revision of established identities. Accumulated evidence governs its evolving object set through a semi-online lifecycle: \emph{mask discovery} (MD) proposes candidates in unexplained regions; \emph{deferred admission} (DA) propagates them provisionally before promotion, merging, or rejection; and \emph{track consolidation} (TC) resolves conflicting, reappearing, and part--whole tracks. MD is a pluggable interface rather than a prescribed hierarchy. Our evaluated instantiation, hierarchical mask discovery (HMD), combines multi-granularity SAM proposals. Thus, candidate generation can change without making admission and identity maintenance depend on a particular mask proposer.

Evaluation poses a second, distinct challenge. VPQ~\cite{vpqvps} uses thresholded $1{:}1$ matches between predicted and reference tubes; STQ~\cite{stq} instead aggregates pixel-level associations across overlapping tracks, without a matching threshold. Both evaluate against a benchmark-defined object partition. In category-independent OVS, a coherent set of predicted tracks may partition one reference into meaningful parts, so evaluation can conflate granularity mismatch with segmentation or tracking failure. But unrestricted many-to-one credit would also reward gratuitous fragmentation. The evaluation must relax decomposition agreement while retaining ownership and temporal identity constraints.

We introduce \textbf{OGA}, a granularity-aware evaluation suite using asymmetric, reference-anchored $n{:}1$ support. Within each GA scoring scope, each prediction is assigned to at most one reference; several predictions may support the same reference. OGA retains prediction IDs so that temporal severing and distributed support remain visible to adapted STQ/VPQ scores and two-sided identity-persistence and identity-concentration diagnostics. Paired conventional and OGA scores quantify sensitivity to the reference partition.

Our primary evaluation covers all 142 ScanNet validation videos and 106 HM3D~\cite{hm3d} trajectories with instance-level references, using adaptively sampled prefixes capped at 1,500 frames. Savvy leads directly applicable baselines in long-window VPQ, STQ, and AQ under both protocols, so its advantage precedes OGA; DEVA+SAM retains higher foreground-coverage GQ. VIPSeg diagnoses granularity mismatch, ablations isolate lifecycle decisions, and controlled fragmentation tests distinguish fidelity from identity structure.

\section{Related Work}
\textbf{Propagation, Discovery, and Association.}
Cutie~\cite{cutie} and SAM2~\cite{sam2} advance propagation of specified targets; SAM-PT~\cite{sampt} and 3AM~\cite{threeam} strengthen correspondence through tracked points and geometry-aware features. DEVA~\cite{deva} couples image proposals with propagation and multi-frame fusion; EntitySAM~\cite{entitysam} and VideoSAM~\cite{videosam} address automatic entity discovery and association. MASA~\cite{masa} learns transferable instance embeddings and supports automatic detection and new-track creation. Savvy connects modular discovery to provisional admission and consolidation of returning, competing, and part--whole tracks.

\textbf{Categories and Concepts.}
OV2Seg~\cite{ov2seg} and InstFormer~\cite{openvis} extend video instance segmentation to novel category names. SAM~3~\cite{sam3} detects and tracks concept-prompted instances with delayed confirmation and duplicate suppression. Savvy targets unprompted, category-independent object-set maintenance, combining candidate evidence with subsequent revision of established identities and their histories.

\textbf{Cross-View Object Representations.}
SAI3D and Open3DIS~\cite{sai3d,open3dis} use multi-view masks to form 3D instances; ESAM~\cite{esam} maintains instances online from posed RGB-D. IGGT~\cite{iggt} jointly predicts reconstruction and instance features from RGB views, while Multiview Scene Graph~\cite{msg} associates objects and places from unposed images. Savvy shares their cross-view identity objective, producing dense video masks and persistent IDs through an explicit admission and consolidation lifecycle.

\textbf{Evaluation Across Granularities.}
VPQ~\cite{vpqvps} matches tubes one-to-one within windows; STQ~\cite{stq} aggregates full-sequence pixel associations without thresholded matching. OWTA~\cite{owta} addresses incomplete annotation through detection recall and track association. PartPQ~\cite{partpq} evaluates annotated part hierarchies, and flexible panoptic matching~\cite{instance_matching2026} permits unions of predicted fragments. OGA couples reference-anchored $n{:}1$ support with temporal support selection and retained prediction IDs. Its fidelity scores credit coherent parts, while two-sided diagnostics reveal identity bleeding and distributed reference support. Appendix~\ref{app:extended_related_work} develops these connections.

\begingroup
\setlength{\intextsep}{6pt}
\begin{figure}[H]
    \centering
    \includegraphics[width=\textwidth]{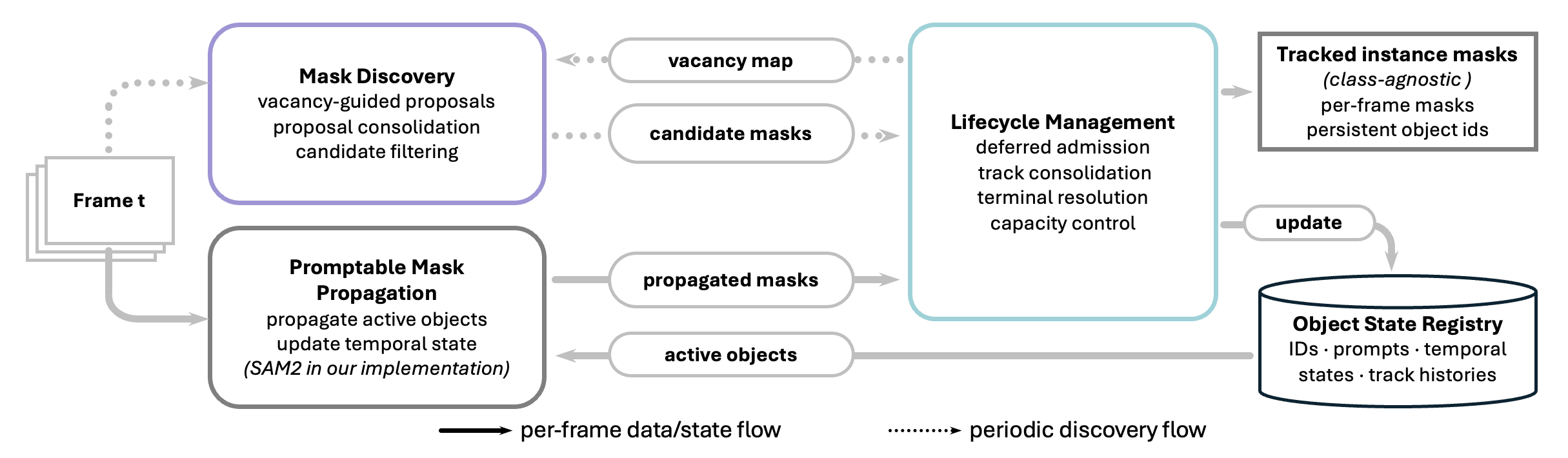}
    \caption{\small \textbf{Overview of the Savvy pipeline.} Savvy augments promptable mask propagation with periodic discovery and object lifecycle management. At each frame, a promptable propagation model propagates active objects using temporal state held in a logical object-state registry maintained throughout the video; our implementation uses SAM2. At initialization and periodically thereafter, mask discovery uses the current frame and a vacancy map of regions unexplained by the current object state to generate candidate masks. Lifecycle management integrates propagated and candidate masks through deferred admission, track consolidation, terminal resolution, and capacity control. It updates the registry and produces class-agnostic per-frame instance masks with persistent object IDs. Solid arrows denote per-frame data and state flow; dotted arrows denote the periodic discovery flow.}
    \label{fig:savvy_overview}
\end{figure}
\endgroup
\section{Methodology}
\label{sec:method}
\subsection{Savvy: Persistent Object-Set Maintenance}
\textbf{Task and state.}
Given an unprompted video $I_{1:T}$, Savvy produces class-agnostic instance masks with persistent IDs. A proposal alone does not establish a durable identity: DA decides whether a candidate should join the object set, while TC checks whether established identities remain consistent. A logical \emph{object-state registry} maintains established tracks $\mathcal{O}$, provisional tracks $\mathcal{B}$, prompts, propagation states, and histories. Processing is \emph{semi-online}: candidates propagate before admission, and later lifecycle decisions can revise stored output identities.

\textbf{Propagation and discovery cadence.}
A promptable propagator $\Psi$ updates $\mathcal{O}\cup\mathcal{B}$ at every frame; we instantiate it with SAM2. MD bootstraps established tracks on the first frame and periodically proposes candidates from unexplained regions thereafter. Later discoveries enter DA rather than becoming established immediately. Lifecycle decisions precede further discovery (Figure~\ref{fig:savvy_overview}; Algorithm~\ref{alg:savvy_main}); Figure~\ref{fig:savvy_qualitative} illustrates the resulting tracks across a revisit.

\textbf{Mask discovery (MD).}
MD accepts a frame and vacancy map and proposes masks. Our \emph{hierarchical mask discovery} (HMD) instantiation samples SAM prompts outside established, provisional, and recent-eviction masks, then consolidates multi-scale proposals. After initialization, overlap and geometric filters remove occupied support and weak remnants before candidates enter $\mathcal{B}$. Proposal generation is thus separate from identity admission (Appendix~\ref{app:hmd}).

\begingroup
\setlength{\intextsep}{6pt plus 2pt minus 2pt}
\tcbset{before skip=0pt,after skip=0pt}
\begin{algorithm}[H]
\begin{algorithmpanel}
\caption{Savvy semi-online object lifecycle}
\label{alg:savvy_main}
\small
\begin{algorithmic}[1]
\Require Video $I_{1:T}$; propagator $\Psi$; discovery operator $\mathrm{MD}$; stride $s$
\Ensure Class-agnostic instance masks with persistent IDs $\mathcal{Y}_{1:T}$
\State Bootstrap $\mathcal{O}$ and output from $\mathrm{MD}(I_1)$; $\mathcal{B}\gets\varnothing$
\Statex \emph{Predictor-track removals restart at the next frame, deferring later stages.}
\For{$t=2,\ldots,T$}
    \State Propagate $\mathcal{O}\cup\mathcal{B}$; store masks and update candidate evidence
    \State Gate returning tracks before updating appearance; arbitrate current conflicts
    \State Merge candidates with sufficient same-ID IoU hits; remap their buffered masks
    \State Apply any queued part absorption; remap its entire stored output history
    \For{each remaining candidate $z$ whose admission window has expired}
        \If{$z$ is not a residual duplicate and passes the promotion evidence gates}
            \State Promote $z$ to $\mathcal{O}$; remove it from $\mathcal{B}$ and backfill its masks
            \State Absorb contained established parts into $z$; remap their full histories
        \Else
            \State Discard $z$
        \EndIf
        \State Remove $z$ from $\mathcal{B}$
    \EndFor
    \State Prune stale predictor memory and enforce the active-track budget
    \If{discovery is due and no preceding removal deferred it}
        \State Form vacancy map; discover, filter, and register new provisional tracks
    \EndIf
\EndFor
\State Resolve remaining candidates using the same evidence gates, waiving window age
\State Drain queued absorptions and finish stored output updates
\State \Return $\mathcal{Y}_{1:T}$
\end{algorithmic}
\end{algorithmpanel}
\end{algorithm}

\begin{figure}[H]
    \centering
    \includegraphics[width=\linewidth]{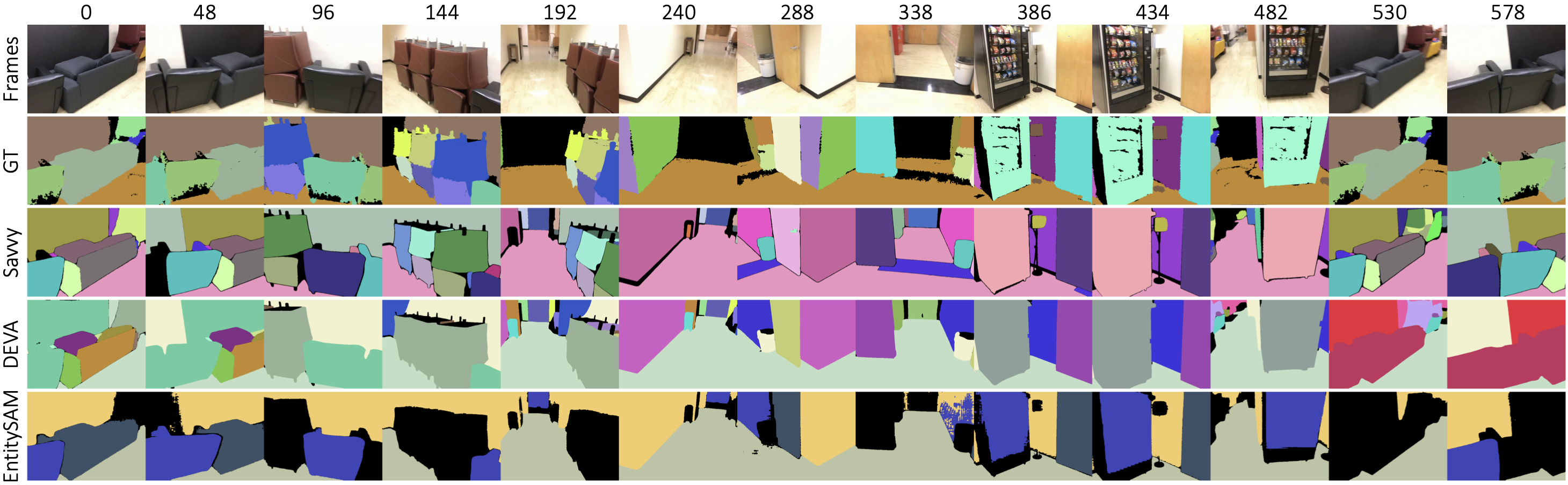}
    \caption{\small \textbf{Persistent discovery and revisits on ScanNet.} The camera leaves the seating area, observes doors and a vending machine, then returns (frames 530--578). Savvy retains detailed object partitions and consistent seating IDs; DEVA+SAM changes seating IDs. EntitySAM exhibits identity collapse: the same purple ID spans different sofas, the vending machine, and a distant corridor region. Rows show RGB, GT, Savvy, DEVA+SAM, and EntitySAM; columns show frame indices.}
    \label{fig:savvy_qualitative}
\end{figure}
\endgroup

\textbf{Deferred admission (DA).}
DA separates early duplicate resolution from delayed admission. Repeated mask-IoU agreement with the \emph{same} established ID triggers merging without waiting for expiry. At window expiry, remaining candidates with sufficient duplicate evidence are discarded; others are promoted if their visibility and any enabled size gates pass, or discarded otherwise. Promotion backfills admission-buffer masks; merging unions buffered support into the existing ID. At video end, \emph{terminal resolution} applies the same evidence tests to unfinished windows, waiving only the waiting requirement (Appendix~\ref{app:deferred_admission}).

\textbf{Track consolidation (TC).}
TC makes three distinct corrections. \emph{Reappearance gating} checks a returning track against its own appearance memory before updating it, using pooled multi-level SAM2 features. \emph{Competition arbitration} suppresses overlapping masks using appearance consistency, a presence counter, and seniority; missing appearance scores fall back directly to seniority. Both operations reject current masks while retaining the identities. In contrast, \emph{part--whole absorption}, triggered by promotion of a new whole, removes redundant established part identities and remaps their \emph{entire stored output histories}, beyond the candidate buffer (Appendix~\ref{app:track_consolidation}).

\textbf{Capacity control.}
Stale predictor memory is pruned. Capacity eviction considers established tracks before provisional ones, ordered by historical maximum area within each group. Recent eviction masks temporarily exclude discovery prompts (Appendix~\ref{app:memory_management}).

\subsection{\texorpdfstring{OGA: \underline{O}pen-world \underline{G}ranularity-\underline{A}gnostic Evaluation}{OGA: Open-world Granularity-Agnostic Evaluation}}
\textbf{Identity-first evaluation.}
OGA retains prediction IDs to distinguish persistent part-level support from temporal identity replacement. It proceeds through reference assignment, temporal support selection, fidelity scoring, and two-sided structural diagnostics.

\begin{tealbox}
\textbf{Definition 1 (GA Matching).}
Within an evaluation scope, let $\mathcal{G}$ and $\mathcal{P}$ denote reference and predicted instances. GA matching assigns each $p\in\mathcal{P}$ to \emph{at most one} $g\in\mathcal{G}$, while multiple predictions may support the same reference. This asymmetric $n{:}1$ rule permits finer decompositions without duplicating matching credit across references.
\end{tealbox}
\textbf{GA association.\ }
Let $g\in\mathcal{G}$ and $p\in\mathcal{P}$ be spatiotemporal tubes over the current evaluation scope (the full evaluated sequence for STQ or a temporal window for VPQ). All areas and intersections are measured on the dataset's valid reference domain; prediction spill into void/ignore pixels is excluded from area accounting, accommodating incomplete annotations. OGA uses the asymmetric prediction-support ratio $S(p,g)=|p\cap g|/|p|$ and defines
\begin{equation}
\mathcal{V}(p)=\left\{g\in\mathcal{G}\,\middle|\, S(p,g)\ge\tau_{\mathrm{ios}}\right\}.
\end{equation}
If $\mathcal{V}(p)=\varnothing$, $p$ remains unmatched; otherwise,
\begin{equation}
a(p)=\arg\max_{g\in\mathcal{V}(p)} S(p,g).
\end{equation}
The asymmetric criterion admits support from part of a larger reference. The one-reference rule governs GA assignment within this scope; it does not restrict the raw overlaps used by IP or the framewise relations used by IC below. GQ separately measures foreground coverage.

\textbf{Temporally coherent support.\ }
Let $\mathcal{P}(g)=\{p\in\mathcal{P}\mid a(p)=g\}$. Because several predictions may support $g$, OGA tracks which identities carry that support. Let $t_1<\cdots<t_{K_g}$ be the ordered states with non-empty assigned support and define
\begin{equation}
\begin{gathered}
A_{t_k}(g)=\{p\in\mathcal{P}(g)\mid |p_{t_k}\cap g_{t_k}|>0\},\qquad
M_{t_k}(g)=\sum_{p\in A_{t_k}(g)}|p_{t_k}\cap g_{t_k}|,\\[-2pt]
\mathcal{C}(g)=\{(t_k,A_{t_k}(g),M_{t_k}(g))\}_{k=1}^{K_g}.
\end{gathered}
\end{equation}
OGA detects identity replacement by measuring the retained support mass between consecutive states:
\begin{equation}
\rho_g^{(k)}=
\frac{\sum_{p\in A_{t_k}(g)\cap A_{t_{k+1}}(g)}
\big(|p_{t_k}\cap g_{t_k}|+|p_{t_{k+1}}\cap g_{t_{k+1}}|\big)}
{\sum_{p\in A_{t_k}(g)\cup A_{t_{k+1}}(g)}
\big(|p_{t_k}\cap g_{t_k}|+|p_{t_{k+1}}\cap g_{t_{k+1}}|\big)+\varepsilon}.
\end{equation}
Low $\rho_g^{(k)}$ means that support has largely moved to different identities. A \emph{sever} is declared when $\rho_g^{(k)}\leq\tau_{\mathrm{sever}}$, partitioning $\mathcal{C}(g)$ into fragments.

\begin{figure}[H]
    \centering
    \includegraphics[width=0.9\textwidth]{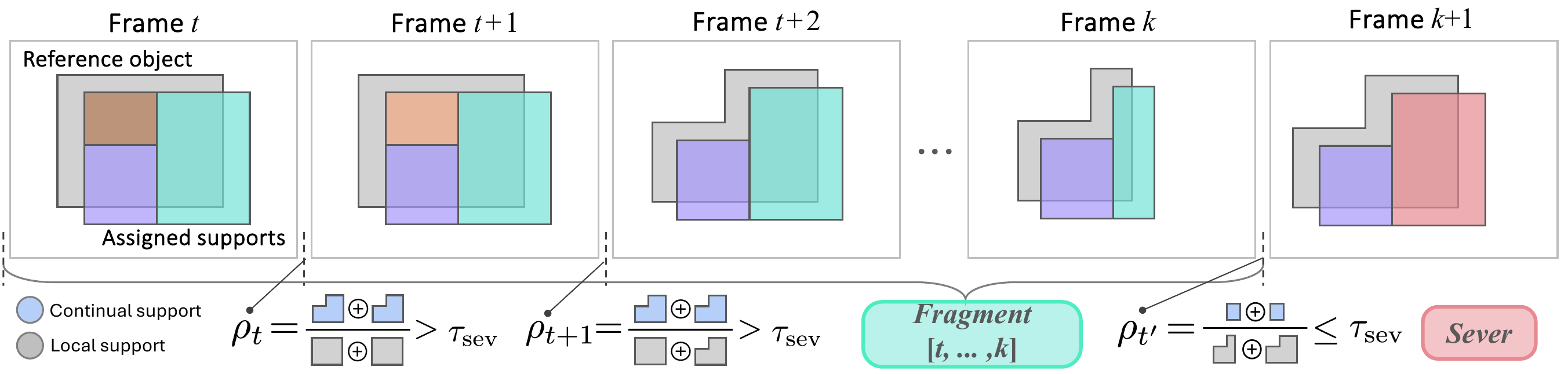}
\caption{\small \textbf{Severing under GA.} A reference may be supported by different prediction sets over time. $\rho$ measures the support mass retained by the same identities between consecutive non-empty states. Low $\rho$ severs the chain; only its highest-mass fragment is credited.}
    \label{fig:oga}
\end{figure}

Exact coherent subset selection is combinatorial, so OGA uses a deterministic \emph{dominant-fragment} surrogate. Among the sever-induced fragments $\Pi(g)$, it retains the one with greatest accumulated support mass:
\begin{equation}
F^\star(g)=\arg\max_{F\in\Pi(g)} \sum\nolimits_{(t,A_t(g),M_t(g))\in F} M_t(g),
\end{equation}
and credits only its prediction identities and non-empty support-state frames. Unsupported frames do not enter identity-transition comparisons, but their reference area remains in the scoring denominator.

\textbf{GA-adapted STQ and VPQ.\ }
For an evaluation scope $E$, let $B_g^E$ be the \emph{full} reference area of $g$. Over the dominant fragment's selected support states and prediction IDs, $D_g^E$ accumulates intersections with $g$, and $C_g^E$ accumulates valid prediction areas, including area outside $g$. The virtual-tube score combines overlap quality, active-ID stability, and an excess-area penalty:
\begin{equation}
q_g^E=\frac{D_g^E}{B_g^E+C_g^E-D_g^E+\varepsilon}\,\mathrm{Stab}_g^E
\left(\min\!\left\{1,\frac{B_g^E}{C_g^E+\varepsilon}\right\}\right)^{\!\gamma}.
\end{equation}
Here $\mathrm{Stab}_g^E$ is the mean Jaccard overlap of active identity sets at consecutive states of the dominant fragment (one for a single state), and $\gamma=1$. Using the full $B_g^E$ retains the penalty for missing support outside the selected fragment. $\mathrm{GQ}_{\mathrm{GA}}$ is the fraction of valid reference foreground covered by any prediction foreground. Suppressing the full-sequence superscript, GA-STQ combines this coverage with reference-averaged association:
\begin{equation}
\mathrm{AQ}_{\mathrm{GA}}=\frac{1}{|\mathcal{G}|}\sum_{g\in\mathcal{G}}\frac{D_g q_g}{B_g+\varepsilon},
\qquad \mathrm{STQ}_{\mathrm{GA}}=\sqrt{\mathrm{AQ}_{\mathrm{GA}}\mathrm{GQ}_{\mathrm{GA}}}.
\end{equation}
For each VPQ window $W$, OGA recomputes GA assignment, dominant fragments, and $q_g^W$. References with $q_g^W>\tau_{\mathrm{vpq}}$ form the true-positive set $T_W$; the others are false negatives. Ordinary false positives are non-void predictions outside all support fragments. If $n_g^W$ is the number of sever-induced fragments for $g$, the total false-positive count is $\mathrm{FP}_W=\mathrm{FP}_{\mathrm{ord}}^W+\sum_{g\in\mathcal{G}_W}\max(n_g^W-1,0)$. The window PQ and its average are
\begin{equation}
\mathrm{PQ}_{\mathrm{GA}}^W=
\frac{\sum_{g\in T_W}q_g^W}{|T_W|+\tfrac12\mathrm{FP}_W+\tfrac12(|\mathcal{G}_W|-|T_W|)+\varepsilon},
\quad \mathrm{VPQ}_{k}^{\mathrm{GA}}=\frac{1}{|\mathcal{W}_k|}\sum_{W\in\mathcal{W}_k}\mathrm{PQ}_{\mathrm{GA}}^W.
\end{equation}
Here $\mathcal{W}_k$ is the set of evaluation windows ($k=0$ denotes a single frame). Persistent part-level support can be credited, while severed excess support incurs false positives; IP/IC below expose residual proliferation. We use $\tau_{\mathrm{iou}}=\tau_{\mathrm{ios}}=\tau_{\mathrm{sever}}=\tau_{\mathrm{vpq}}=0.5$; Appendices~\ref{app:ga_matching}--\ref{app:ga_vpq} give the edge cases.

\textbf{Structural Diagnostics.\ }
The primary metrics measure fidelity under the relaxed decomposition; two-sided diagnostics reveal how that support is organized. \textbf{Identity persistence} (IP) measures dominant overlap mass on each axis:
\begin{equation}
\mathrm{IP}_{\mathrm{P}}(p)=\frac{\max_{g\in\mathcal{G}} |p\cap g|}{|p|},
\qquad
\mathrm{IP}_{\mathrm{G}}(g)=\frac{\max_{p\in\mathcal{P}} |p\cap g|}{|g|}.
\end{equation}
\textbf{IP(P)} measures dominant-reference purity; \textbf{IP(G)}, dominant-prediction coverage. Sequence scores scale the corresponding mean ratio by the shared discovery factor $D$.

Here $D$ is the frame mean of $\min(m_t/\max(|\mathcal{G}_t|,1),1)$, where $m_t$ counts distinct prediction IDs matching at least one reference by framewise IoU $\geq\tau_{\mathrm{iou}}$ or one-sided overlap $|p_t\cap g_t|/|p_t|\geq\tau_{\mathrm{ios}}$. The ratios use all accumulated valid pixel intersections, without IoS filtering; $\tau_{\mathrm{ios}}$ affects sequence IP only through $D$.

\textbf{Identity concentration} (IC) is instead graph-based. To keep it distinct from the tube-level assignment, define $S_t(p,g)=|p_t\cap g_t|/|p_t|$ for frames with $|p_t|>0$, and let $p\rightsquigarrow g$ denote an edge in the \emph{ever-support graph}:
\begin{equation}
p\rightsquigarrow g
\Longleftrightarrow
\exists t:\ \mathrm{IoU}(p_t,g_t)\ge\tau_{\mathrm{iou}}\ \text{or}\ S_t(p,g)\ge\tau_{\mathrm{ios}}.
\end{equation}
Let $\mathcal{G}_{\mathrm{ever}}(p)=\{g\mid p\rightsquigarrow g\}$ and $\mathcal{P}_{\mathrm{ever}}(g)=\{p\mid p\rightsquigarrow g\}$. Then
\begin{equation}
\mathrm{IC}_{\mathrm{G}}(g)=
\begin{cases}
\dfrac{1}{|\mathcal{P}_{\mathrm{ever}}(g)|}, & |\mathcal{P}_{\mathrm{ever}}(g)|>0,\\[4pt]
0, & \text{otherwise,}
\end{cases}
\qquad
\mathrm{IC}_{\mathrm{P}}(p)=
\begin{cases}
\dfrac{1}{|\mathcal{G}_{\mathrm{ever}}(p)|}, & |\mathcal{G}_{\mathrm{ever}}(p)|>0,\\[4pt]
0, & \text{otherwise.}
\end{cases}
\end{equation}
Prediction-side scores expose identity bleeding; reference-side scores expose distributed support. Pure fragments can retain perfect prediction-side scores, so both axes are necessary.

\begin{tealbox}[top=0.2cm,bottom=0.2cm]
\textbf{Worked example: parts versus identity replacement.}
One reference has equal area in four frames, with perfect foreground coverage and no competing references. In all three cases, $D=\mathrm{GQ}=\mathrm{IP}_P=\mathrm{IC}_P=1$. Scores below are normalized to $[0,1]$.
\begin{center}
\small
\setlength{\tabcolsep}{5pt}
\begin{tabular}{lccccc}
\toprule
Prediction IDs/support & GA-AQ & GA-STQ & GA-VPQ$_\infty$ & IP$_G$ & IC$_G$\\
\midrule
One exact persistent ID & $1$ & $1$ & $1$ & $1$ & $1$\\
Two persistent equal halves & $1$ & $1$ & $1$ & $1/2$ & $1/2$\\
Exact masks; IDs $a,a,a,b$ & $9/16$ & $3/4$ & $1/2$ & $3/4$ & $1/2$\\
\bottomrule
\end{tabular}
\end{center}
Persistent halves retain full fidelity; reference-side diagnostics reveal their distributed support. Replacing $a$ by $b$ severs the chain: the three-frame fragment gives $q=3/4$ and one extra-fragment FP, so VPQ$_\infty=(3/4)/(1+1/2)=1/2$ (Appendix~\ref{app:oga_worked_example}).
\end{tealbox}

\section{Experiments}
\subsection{Experimental Setup}
\textbf{Datasets.} \textbf{VIPSeg}~\cite{vipseg} diagnoses evaluation sensitivity to granularity mismatch. \textbf{ScanNet}~\cite{scannet} supplies 142 long indoor validation videos with repeated object re-observation and strong viewpoint changes. \textbf{HM3D}~\cite{hm3d} (HM3D-Sem v0.2) adds controlled looped revisits with more complete reconstructed references. HM3D-106 contains 106 scans and 293,331 rendered frames. Its 96 composition scans draw from 36 eligible validation scenes containing 569 available semantic regions. All evaluations are class-agnostic.

\textbf{Baselines.} We select DEVA+SAM and EntitySAM because both support automatic discovery throughout RGB videos and persistent mask identities without supplied category queries or target masks. We evaluate all methods using pretrained checkpoints, without additional training or fine-tuning on VIPSeg, ScanNet, or HM3D. \textbf{DEVA+SAM}~\cite{deva} combines image proposals with temporal propagation and fusion; \textbf{EntitySAM}~\cite{entitysam} uses SAM2 and DINOv2 for learned entity discovery. Appendix~\ref{app:baseline_compatibility} (Table~\ref{tab:baseline_compatibility}) distinguishes direct comparisons from task-level adaptations, whose performance depends on both the original method and the added discovery or identity-management mechanisms. Savvy and DEVA+SAM are semi-online. \textbf{SAM2}~\cite{sam2} serves as an initialized-target VIPSeg control without mid-video discovery.

\textbf{Evaluation protocol.} We score the same predictions under conventional class-agnostic evaluation ($1{:}1$ VPQ and original STQ-style association) and OGA. ScanNet and HM3D use instance-level references and adaptive sampling of the first $\min(T,1500)$ source frames per video; later frames are excluded. Scores are computed per video, macro-averaged, and scaled to 0--100. VPQ$_\infty$ uses the entire retained sampled sequence as one window, not the complete source video when truncated; finite windows probe shorter horizons. We report IP(P/G) and IC(P/G) alongside OGA fidelity. Sampling intervals, dataset conversion, and void handling appear in Appendix~\ref{app:dataset_protocol}.

\textbf{Implementation details.}
Savvy is implemented as a semi-online OVS framework with \textbf{SAM2} as the video propagation backbone. We instantiate MD with \textbf{HMD}, a \textbf{SAM}-based proposer adapted to produce multi-level masks. Appendix~\ref{app:implementation_details} summarizes the system hyperparameters, and Appendix~\ref{app:oga_hyperparameters} lists the metric thresholds used by OGA. Unless otherwise stated, the same evaluation pipeline is applied across all methods under the corresponding protocol. All experiments are performed on NVIDIA A6000.

\begin{table*}[t]
\centering
\setlength{\tabcolsep}{7pt}
\renewcommand{\arraystretch}{1.0}
\caption{\small\textbf{Conventional $1{:}1$ VPQ and OGA results on all 142 ScanNet scenes and 106 HM3D trajectories with instance-level references.} Evaluation adaptively samples the first 1,500 source frames (or the complete video if shorter). Scores are macro-averaged per video and scaled to 0--100. On prediction (P) and reference (G) axes, upward diagnostic arrows indicate greater dominant-overlap purity/coverage (IP) or identity concentration (IC). Valid persistent parts can retain perfect OGA fidelity with lower reference-side scores.}
\label{tab:scannet_hm3d}
\resizebox{\linewidth}{!}{
\begin{tabular}{cl|c|cccc|cccc}
\toprule
& & \multicolumn{1}{c|}{\textbf{Rigid $1{:}1$}} & \multicolumn{4}{c|}{\textbf{OGA fidelity $\uparrow$}} & \multicolumn{4}{c}{\textbf{OGA identity structure $\uparrow$}}\\
\cmidrule(lr){3-3} \cmidrule(lr){4-7} \cmidrule(lr){8-11}
\textbf{Data} & \textbf{Method} & \textbf{VPQ$_\infty$} & \textbf{VPQ$_\infty$} & \textbf{STQ} & \textbf{AQ} & \textbf{GQ} & \textbf{IP$_P$} & \textbf{IP$_G$} & \textbf{IC$_P$} & \textbf{IC$_G$}\\
\midrule
\multirow{3}{*}{ScanNet}
& EntitySAM & 8.92 & 10.82 & 17.65 & 6.27 & 61.61 & 24.93 & 17.31 & 9.02 & 19.19 \\
& DEVA+SAM & 11.52 & 18.53 & 43.15 & 20.35 & \textbf{95.87} & 63.21 & 50.38 & 54.82 & 25.90 \\
& \cellcolor{ac!30}Savvy (Ours) & \cellcolor{ac!30}\textbf{18.44} & \cellcolor{ac!30}\textbf{28.63} & \cellcolor{ac!30}\textbf{53.93} & \cellcolor{ac!30}\textbf{32.59} & \cellcolor{ac!30}91.11 & \cellcolor{ac!30}\textbf{75.44} & \cellcolor{ac!30}\textbf{57.92} & \cellcolor{ac!30}\textbf{62.68} & \cellcolor{ac!30}\textbf{32.03} \\
\midrule
\multirow{3}{*}{HM3D}
& EntitySAM & 1.35 & 1.46 & 5.97 & 0.74 & 64.37 & 8.36 & 5.80 & 14.46 & 15.06 \\
& DEVA+SAM & 6.58 & 8.47 & 25.26 & 6.94 & \textbf{94.81} & 35.55 & 26.11 & \textbf{74.56} & 26.57 \\
& \cellcolor{ac!30}Savvy (Ours) & \cellcolor{ac!30}\textbf{17.31} & \cellcolor{ac!30}\textbf{19.59} & \cellcolor{ac!30}\textbf{35.69} & \cellcolor{ac!30}\textbf{14.28} & \cellcolor{ac!30}91.26 & \cellcolor{ac!30}\textbf{53.15} & \cellcolor{ac!30}\textbf{37.07} & \cellcolor{ac!30}74.46 & \cellcolor{ac!30}\textbf{32.16} \\
\bottomrule
\end{tabular}
}
\end{table*}

\newpage
\subsection{Evaluation on Realistic OVS Benchmarks}
Savvy leads conventional VPQ$_\infty$ before GA support is introduced (Table~\ref{tab:scannet_hm3d}). Conventional STQ also favors Savvy over DEVA+SAM: 53.84 versus 46.66 on ScanNet and 40.35 versus 31.34 on HM3D (Tables~\ref{tab:scannet_rigid_full} and~\ref{tab:hm3d106_full}).

OGA increases VPQ$_\infty$ for each method, but HM3D STQ and AQ decrease for all three, as does EntitySAM's VPQ$_0$: changing support assignment and identity-coherence penalties need not raise scores. Savvy leads OGA VPQ$_\infty$ at 28.63 on ScanNet and 19.59 on HM3D; DEVA+SAM has higher GQ but lower long-window VPQ, STQ, and AQ. Savvy also leads all four structural diagnostics on ScanNet and three on HM3D-106, where its IC$_P$ nearly ties DEVA+SAM (74.46 versus 74.56).

Figure~\ref{fig:savvy_qualitative} shows EntitySAM reusing the purple ID across distinct sofas, the vending machine, and a distant corridor region. IP$_P$ measures the fraction of a prediction overlapping its dominant reference; IC$_P$ decreases as the number of references it ever supports grows. DEVA+SAM's seating-ID changes when the camera returns divide a reference's support across prediction IDs: IP$_G$ measures the reference fraction covered by the dominant prediction ID, while IC$_G$ decreases as more prediction IDs ever support that reference.

Tracking-only profiling on a 30-scene ScanNet minival subset gives Savvy 1.34 frame-weighted FPS and 6,631 MB mean per-sequence peak GPU memory, versus 4.31 FPS/14,484 MB for DEVA+SAM and 1.97 FPS/20,527 MB for EntitySAM. Savvy maintains a larger object set with lower peak memory, at the cost of lower throughput; Appendix~\ref{app:runtime_memory} details the timing boundary and per-method behavior. Appendices~\ref{app:scannet}--\ref{app:hm3d} describe the benchmarks, with extended results in Appendices~\ref{app:full_scannet_results} and~\ref{app:full_hm3d_results}.
\subsection{Ablation Study of Savvy}
On the 30-scene ScanNet minieval subset, Table~\ref{tab:savvy_ablation} evaluates a cumulative sequence: replace raw AMG with HMD, add deferred admission, then add track consolidation. Adjacent rows isolate each addition conditional on the preceding configuration.

With immediate admission and TC disabled, replacing SAM's default \textbf{Automatic Mask Generator (AMG)} with \textbf{HMD} raises VPQ$_\infty$ from 25.42 to 28.07 and IC$_G$ from 24.89 to 28.45. Adding \textbf{DA}, still without TC, raises these scores to 30.13 and 29.50, although VPQ$_0$ decreases slightly. Adding \textbf{TC} yields 31.31 and 31.51. Thus, each successive choice improves long-horizon VPQ and reference-side concentration in its tested configuration. The full system also leads STQ and AQ, while raw AMG has higher IC$_P$; prediction-side concentration alone does not imply compact reference-side support.

Appendix~\ref{app:additional_ablations} extends these ablations with parameter sweeps.

\begin{table}[t]
\centering
\caption{\small \textbf{Cumulative ablation on the 30-scene ScanNet minieval subset.} Read bottom to top: replace AMG with HMD, add deferred admission (DA), then add track consolidation (TC). Each adjacent comparison changes one choice. The complete system leads VPQ$_\infty$, STQ, AQ, and IC$_G$.}
\small
\setlength{\tabcolsep}{6pt}
\resizebox{0.96\linewidth}{!}{
\begin{tabular}{cc|cc|c|cccccc}
\toprule
AMG & HMD & Imm. Adm. & DA & TC & VPQ$_{\infty}$ $\uparrow$ & VPQ$_0$ $\uparrow$ & STQ $\uparrow$ & AQ $\uparrow$ & IC$_P$ $\uparrow$ & IC$_G$ $\uparrow$ \\
\midrule
& \checkmark &  &\checkmark & \checkmark & \textbf{31.31} & 58.53 & \textbf{56.02} & \textbf{34.93} & 61.16 & \textbf{31.51} \\
& \checkmark &  &\checkmark & & 30.13 & 58.56 & 55.44 & 34.44 & 58.15 & 29.50 \\
& \checkmark & \checkmark & & & 28.07 & \textbf{59.20} & 55.99 & 34.91 & 58.50 & 28.45 \\
\checkmark & & \checkmark & & & 25.42 & 58.84 & 55.66 & 34.48 & \textbf{62.15} & 24.89 \\
\bottomrule
\end{tabular}
}
\label{tab:savvy_ablation}
\end{table}

\subsection{Validating OGA: Granularity and Identity Structure}
\textbf{Granularity sensitivity.} EntitySAM's rigid VIPSeg lead lies mainly in RQ rather than SQ, consistent with its COCO panoptic supervision~\cite{entitysam} favoring annotation-compatible partitions. GA raises Savvy's VPQ$_\infty$ from 26.15 to 55.35 and EntitySAM's from 54.68 to 59.23; Savvy has higher AQ under both protocols (Appendix~\ref{app:full_vipseg_results}).

\textbf{Temporal perturbations.} Figure~\ref{fig:stress} tests five failure modes. Sever and flicker preserve frame-level masks but disrupt identity continuity, reducing long-horizon scores and reference-side concentration.

\begin{samepage}
\textbf{Persistent spatial fragmentation.} A separate 30-scene ScanNet minieval test (Appendix~\ref{app:spatial_fragmentation}) splits each track into $k\in\{1,2,4,8\}$ persistent spatial children while preserving the foreground union. At $k=8$, GQ stays at 91.59 and STQ changes only from 55.52 to 55.34, while VPQ$_\infty$ falls from 31.30 to 22.18, IP$_G$ from 58.02 to 10.65, and IC$_G$ from 31.51 to 5.00. VPQ$_0$ instead rises from 58.53 to 62.63. For long-horizon OVS, we recommend conventional/OGA VPQ$_0$, VPQ$_\infty$, and STQ with its association and spatial components, alongside IP/IC on both axes (Appendix~\ref{app:reporting_protocol}).
\end{samepage}

\begin{figure*}[t!]
    \centering
    \includegraphics[width=\linewidth]{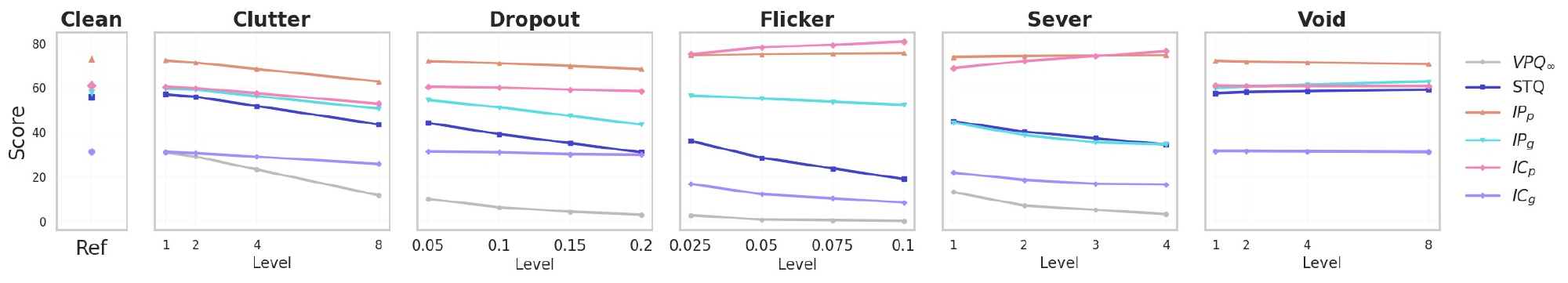}
    \small\caption{\textbf{Stress testing OGA under synthetic failure modes.} 
    OGA responds selectively to these perturbations: overlap-based fidelity drops strongly under dropout; GT-side concentration is especially sensitive to flicker and sever; and void expansion leaves the metrics nearly unchanged. Thus, OGA distinguishes different structural failure modes rather than merely tracking aggregate overlap loss.
}
\label{fig:stress}
\end{figure*}

\section{Conclusion}
We studied two linked challenges in unprompted, class-agnostic long-video segmentation: maintaining an evolving object set, and evaluating coherent predictions that differ from the reference's object partition. \textbf{Savvy} separates mask discovery, deferred admission, and track consolidation; \textbf{OGA} credits reference-anchored $n{:}1$ support while retaining temporal identity penalties and two-sided diagnostics. Paired protocols expose VPQ's granularity sensitivity on VIPSeg. On ScanNet and HM3D, Savvy leads the tested baselines in long-window VPQ, STQ, and AQ under both conventional and OGA evaluation. Ablations and controlled identity perturbations support the respective system and evaluation choices.

\textbf{Scope, Limitations, and Broader Impact.} The long-horizon quantitative evaluation covers indoor ScanNet and HM3D with two directly applicable baselines; long videos combining substantial camera motion with independently moving objects remain largely untested. Savvy studies category-independent identity maintenance rather than open-vocabulary naming, and is slower than both baselines in our tracking-only profile. Persistent object tracking may aid robotics and embodied AI but may also increase monitoring capability; deployment warrants privacy, consent, and context-specific safeguards.

\subsection*{AI use statement}
Generative AI tools assisted with reviewing relevant literature and editing manuscript
text and presentation. The authors conducted the experiments and take responsibility for the final
manuscript.

\clearpage

\endgroup
\bibliographystyle{iclr2027_conference}
\bibliography{references} %

\clearpage
\input{appendix}

\end{document}

%% file: appendix.tex
\appendix
\raggedbottom

\makeatletter
\newcommand{\appcontentssection}[2]{%
  \begingroup\sffamily\bfseries\color{ac}%
  \@dottedtocline{1}{0em}{2.1em}{\hyperref[#1]{\numberline{\ref*{#1}}#2}}{\hyperref[#1]{\pageref*{#1}}}%
  \endgroup\vspace{0.28em}%
}
\newcommand{\appcontentsitem}[2]{%
  \@dottedtocline{2}{0.9em}{3.1em}{\hyperref[#1]{\numberline{\ref*{#1}}#2}}{\hyperref[#1]{\pageref*{#1}}}%
  \vspace{0.16em}%
}
\newcommand{\appcontentssubitem}[2]{%
  \@dottedtocline{3}{2.2em}{3.6em}{\hyperref[#1]{\numberline{\ref*{#1}}#2}}{\hyperref[#1]{\pageref*{#1}}}%
  \vspace{0.16em}%
}
\makeatother
\newcommand{\appcontentsgap}{\vspace{0.7em}}
\newcommand{\ogametric}[1]{\boxed{\displaystyle #1}}
\newenvironment{appcontentspanel}{%
  \begin{tcolorbox}[
    colback=panelbg, colframe=ac, coltext=paperfg,
    boxrule=0.9pt, arc=8pt,
    left=0.5cm, right=0.5cm, top=0.42cm, bottom=0.42cm
  ]%
  \small\linespread{1.08}\selectfont
  \setcounter{tocdepth}{3}%
  \setlength{\parindent}{0pt}%
  \setlength{\parskip}{0.04em}%
  \hypersetup{hidelinks}%
}{\end{tcolorbox}}

\begin{center}
{\Large\sffamily\bfseries Appendix Contents}
\end{center}
\vspace{0.18cm}
\begin{appcontentspanel}
\appcontentssection{app:qualitative_results}{Additional Qualitative Results}
\appcontentsgap

\appcontentssection{app:extended_related_work}{Extended Related Work}
\appcontentsitem{app:rw_task_assumptions}{Task Assumptions and Open-World Benchmarks}
\appcontentsitem{app:rw_prompted_propagation}{Prompted Propagation, Memory, and Long-Term Recovery}
\appcontentsitem{app:rw_automatic_discovery}{Automatic Discovery and Learned Instance Association}
\appcontentsitem{app:rw_open_vocabulary}{Open-Vocabulary and Concept-Prompted Segmentation}
\appcontentsitem{app:rw_cross_view}{Cross-View Association and Scene Representations}
\appcontentsitem{app:rw_part_granularity}{Part Granularity and Flexible Instance Matching}
\appcontentsitem{app:rw_temporal_evaluation}{Temporal Evaluation and Two-Sided Identity Diagnostics}
\appcontentsitem{app:baseline_compatibility}{Method Interfaces and Comparison Requirements}
\appcontentsgap

\appcontentssection{app:savvy_details}{Details of Savvy}
\appcontentsitem{app:savvy_pipeline}{Pipeline Overview}
\appcontentsitem{app:hmd}{MD Instantiation: Hierarchical Mask Discovery}
\appcontentsitem{app:deferred_admission}{Deferred Admission and Transient Buffering}
\appcontentsitem{app:track_consolidation}{Track Consolidation}
\appcontentssubitem{app:appearance_consistency}{Appearance Self-Consistency Gating}
\appcontentssubitem{app:competition_arbitration}{Competition Arbitration}
\appcontentssubitem{app:part_whole_absorption}{Part-Whole Absorption}
\appcontentsitem{app:memory_management}{Memory Management and Active-Track Control}
\appcontentsitem{app:implementation_details}{Implementation Details and Hyperparameters}
\appcontentsitem{app:runtime_memory}{Runtime and Memory Profiling}
\appcontentsgap

\appcontentssection{app:oga_details}{Details of OGA Evaluation}
\appcontentsitem{app:oga_overview}{Overview}
\appcontentsitem{app:ga_matching}{Granularity-Agnostic Matching}
\appcontentsitem{app:ga_stq}{GA-Adapted STQ}
\appcontentsitem{app:ga_vpq}{GA-Adapted VPQ}
\appcontentsitem{app:oga_worked_example}{Worked OGA Example}
\appcontentsitem{app:dominant_fragment}{Dominant Fragment Selection}
\appcontentsitem{app:mwcs}{Dominant Fragment as a Practical Relaxation of Coherent Support Selection}
\appcontentsitem{app:structural_diagnostics}{Structural Diagnostics: Persistence and Concentration}
\appcontentsitem{app:identity_persistence}{Identity Persistence}
\appcontentsitem{app:identity_concentration}{Identity Concentration}
\appcontentsitem{app:support_matrix}{Prediction--Reference Support Matrix}
\appcontentsitem{app:temporal_diagnostics}{Additional Temporal Diagnostics}
\appcontentsitem{app:oga_hyperparameters}{Evaluation Hyperparameters}
\appcontentsitem{app:reporting_protocol}{Recommended Reporting Protocol}
\end{appcontentspanel}
\clearpage
\begin{center}
{\Large\sffamily\bfseries Appendix Contents \mdseries\large (continued)}
\end{center}
\vspace{0.18cm}
\begin{appcontentspanel}
\appcontentssection{app:dataset_protocol}{Dataset and Evaluation Protocol}
\appcontentsitem{app:vipseg}{VIPSeg}
\appcontentsitem{app:scannet}{ScanNet}
\appcontentsitem{app:hm3d}{HM3D}
\appcontentsitem{app:class_agnostic_conversion}{Class-Agnostic Conversion}
\appcontentsitem{app:void_handling}{Void and Ignore-Region Handling}
\appcontentsitem{app:frame_sampling}{Frame Sampling and Evaluation Windows}
\appcontentsgap

\appcontentssection{app:granularity_mismatch}{Analysis of Granularity Mismatch}
\appcontentsitem{app:vipseg_qualitative_matching}{Qualitative Failure of Rigid 1:1 Matching on VIPSeg}
\appcontentsitem{app:class_agnostic_insufficient}{Why Class-Agnostic Evaluation Alone Is Insufficient}
\appcontentsgap

\appcontentssection{app:additional_results}{Additional Experimental Results}
\appcontentsitem{app:full_vipseg_results}{Full VIPSeg Results}
\appcontentsitem{app:full_scannet_results}{Full ScanNet Results}
\appcontentsitem{app:full_hm3d_results}{Full HM3D Results}
\appcontentsitem{app:hm3d_granularity_robustness}{Robustness to HM3D Annotation Granularity}
\appcontentsitem{app:additional_ablations}{Additional Ablation Studies}
\appcontentsgap

\appcontentssection{app:stress_tests}{Synthetic Stress-Test Protocol and Results}
\appcontentsitem{app:spatial_fragmentation}{Simultaneous Spatial Fragmentation}
\appcontentsgap

\appcontentssection{app:oga_visualization_protocol}{OGA Visualization Protocol}
\appcontentsitem{app:support_matrix_visualization}{Support Matrices}
\appcontentsitem{app:ic_typology}{Identity-Concentration Typology}
\appcontentsgap

\appcontentssection{app:failure_analysis}{Failure Cases and Limitations}
\appcontentsitem{app:failure_cases}{Failure Cases}
\appcontentsitem{app:limitations}{Limitations}
\appcontentsgap

\appcontentssection{app:ood_epickitchens}{Out-of-Domain Stress Test on Egocentric Activity Video}
\end{appcontentspanel}

\clearpage
\section{Additional Qualitative Results}
\label{app:qualitative_results}
\label{app:scannet_qualitative}

This section provides additional qualitative results on ScanNet and HM3D, focusing on long-horizon scene-centric videos with strong ego-motion, clutter, object disappearance, and repeated re-observation. These examples complement the quantitative results in Sec.~\ref{app:full_scannet_results} and Sec.~\ref{app:full_hm3d_results}.

\begin{figure}[H]
    \centering
    \includegraphics[width=\linewidth]{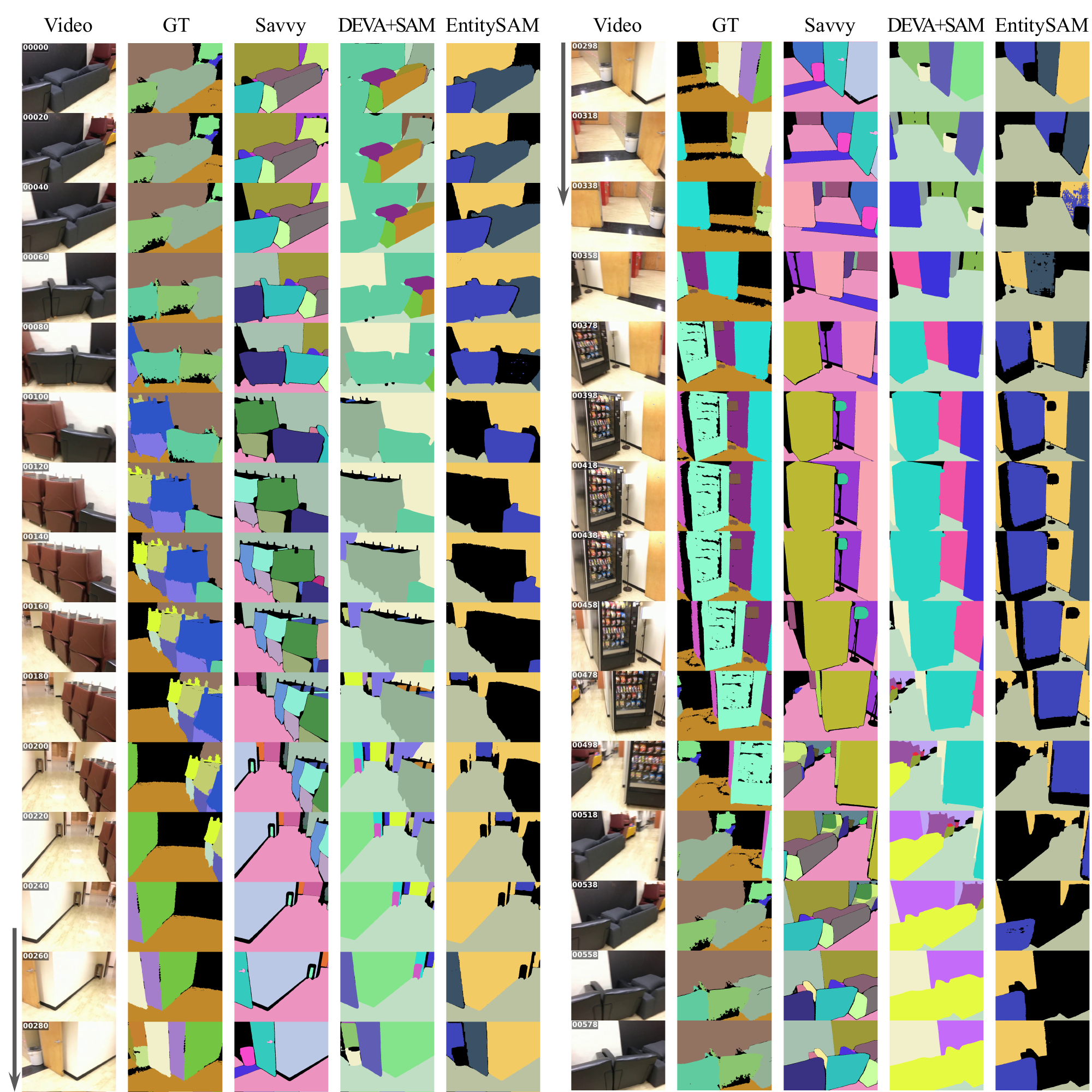}
    \caption{\small
\textbf{Long-horizon qualitative results on a ScanNet sequence (1).}
The sequence is split into two vertical blocks from left to right, following the same video over time.
Early frames show couches and chairs, the camera then moves through a corridor-like region with doors, trash bins, and a vending machine, and later returns to the seating area.
Savvy maintains a more detailed and consistent object decomposition across these transitions: chairs, couch regions, doors, bins, vending-machine surfaces, walls, and floor regions remain separated with relatively stable identities as the camera viewpoint changes.
DEVA+SAM often produces broad foreground coverage, but local object structure is unstable: chair backs and seats blob into nearby furniture, object boundaries bleed into floors or walls, and several thin or adjacent structures are absorbed into large regions.
EntitySAM reuses the same purple ID across different sofas, the vending machine, and a distant corridor region, demonstrating cross-object identity bleeding alongside coarse partitions and limited new-object discovery.
This example illustrates the long-horizon OVS challenge: good performance requires not only segmenting visible foreground, but also expanding the object set when new objects appear, preserving fine object boundaries, and maintaining identities when the camera leaves and later revisits the same scene regions.
}
\label{fig:scannet_qualitative_1}
\end{figure}
\clearpage

\begin{figure}[t]
    \centering
    \includegraphics[width=\linewidth]{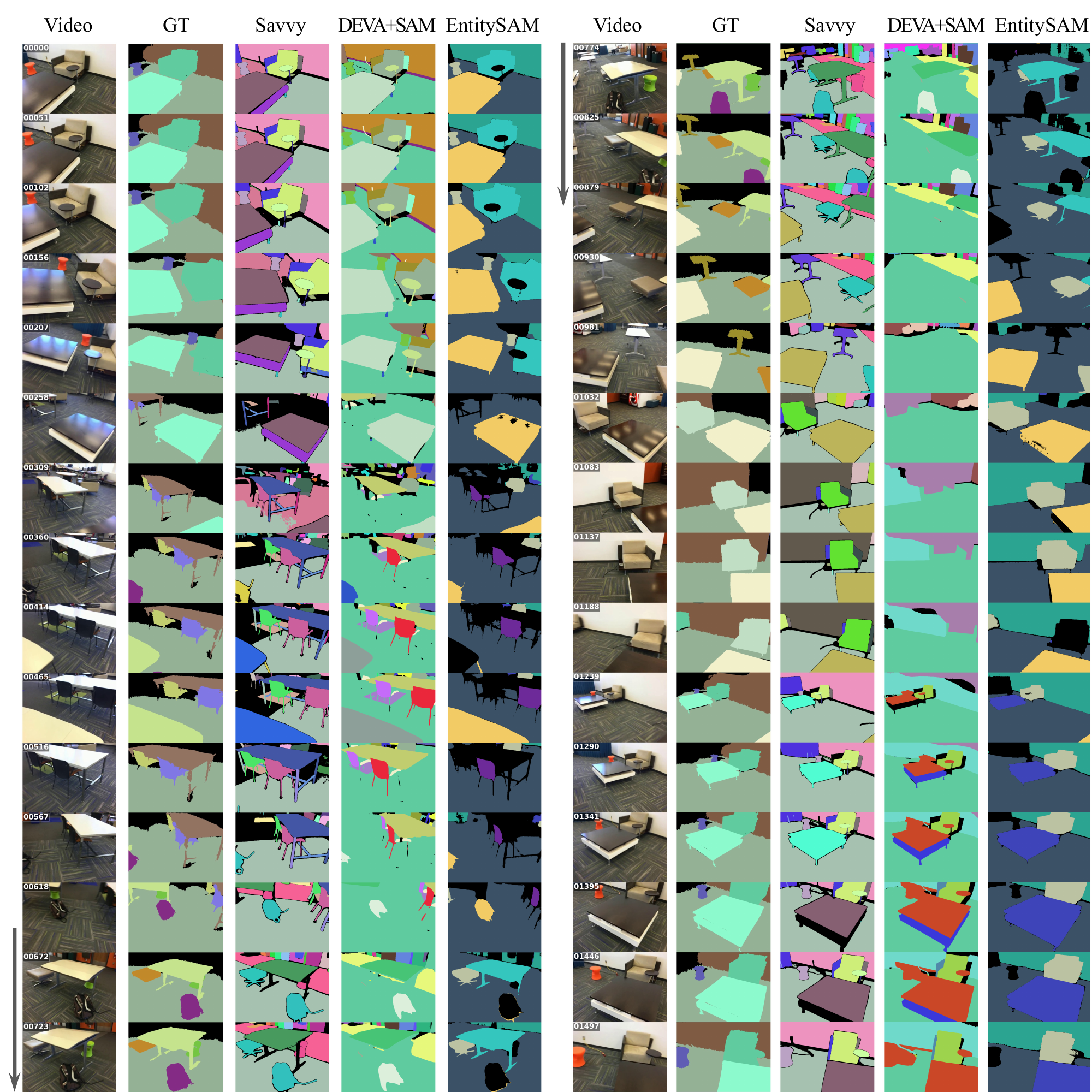}
    \caption{\small
\textbf{Long-horizon qualitative results on a ScanNet sequence (2).}
The sequence is split into two vertical blocks from left to right, following the same video over time.
The camera first observes a seating/table area, then moves through a wider room region with multiple tables and chairs, and later revisits furniture from different viewpoints.
Savvy maintains a more detailed and temporally consistent object decomposition across the sequence: tables, chairs, sofa/couch regions, stools, and surrounding floor/wall structures remain separated as the camera moves and revisits previously seen areas.
DEVA+SAM often provides broad foreground coverage, but its local structure is unstable: nearby objects blob together, chair legs and table boundaries bleed into adjacent surfaces, and thin structures are frequently smeared or inconsistently split across frames.
EntitySAM produces coarse partitions that appear stable at the surface level, but its new-object discovery is limited; newly encountered chairs, tables, and small furniture pieces are often missed or absorbed into large regions, and existing identities may drift onto visually different objects instead of creating new object identities.
}
\label{fig:scannet_qualitative_2}
\end{figure}
\clearpage

\begin{figure}[t]
    \centering
    \includegraphics[width=\linewidth]{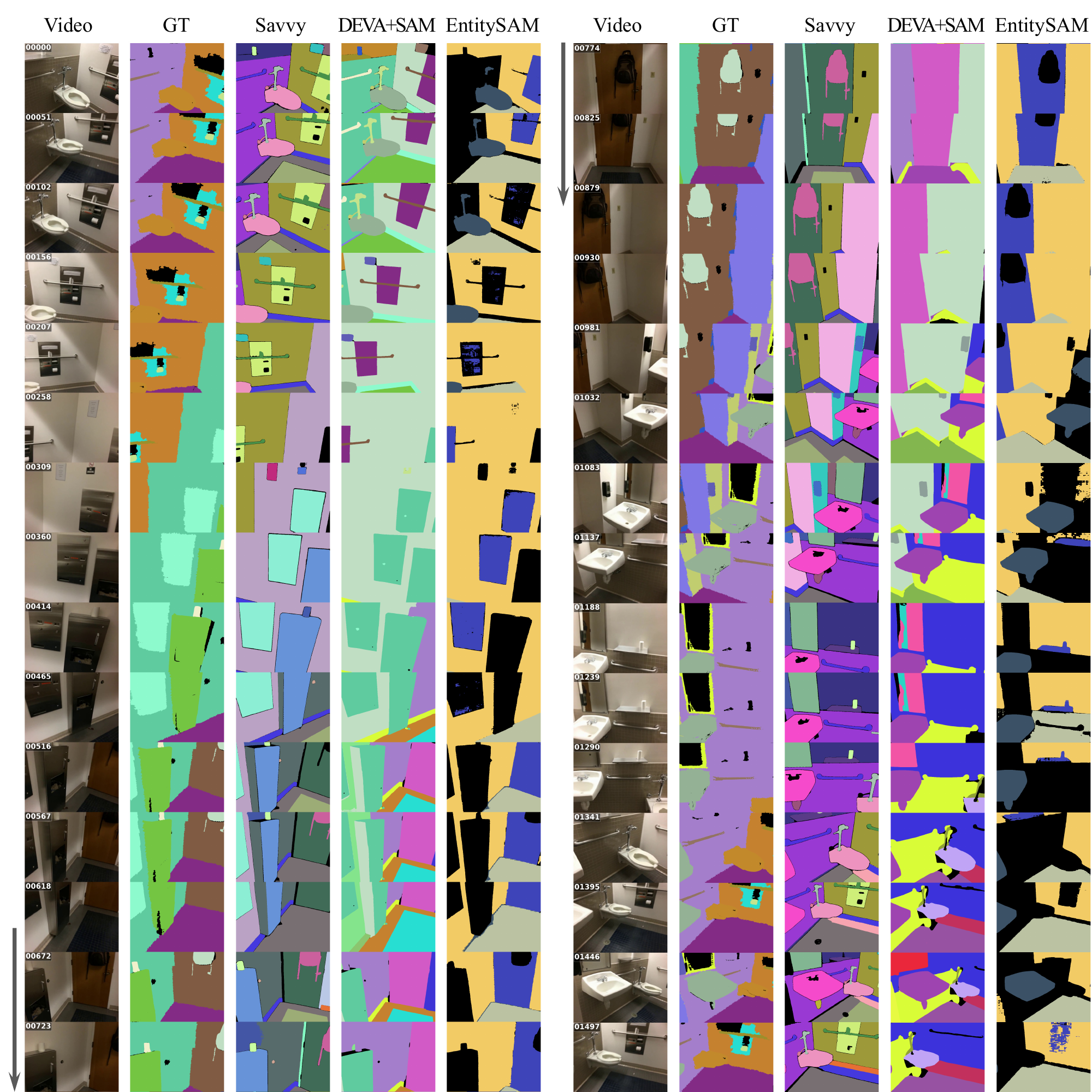}
    \caption{\small
\textbf{Long-horizon qualitative results on a ScanNet sequence (3).}
The sequence is split into two vertical blocks from left to right, following the same video over time.
The camera first observes a sink, wall fixtures, counter regions, and nearby floor/wall surfaces, then turns around towards the door region with hanging backpack, and later revisits the bathroom fixtures from a different viewpoint.
Savvy maintains a detailed and temporally consistent object decomposition across these large viewpoint changes: the sink basin, counter surface, wall-mounted fixtures, door, floor, wall regions, and small objects remain better separated and are re-associated when the camera returns.
DEVA+SAM often segments broad foreground regions, but its local structure is unstable: sink and counter regions bleed into surrounding walls, thin fixtures are inconsistently preserved, and large wall/floor regions absorb nearby objects across frames.
EntitySAM produces much coarser partitions with limited new-object discovery; large surfaces remain dominant, while small fixtures, sink details, and corridor objects are frequently missed, absorbed into background-like regions, or assigned inconsistent identities as the viewpoint changes.
}
\label{fig:scannet_qualitative_3}
\end{figure}
\clearpage

\begin{figure}[t]
    \centering
    \includegraphics[width=\linewidth]{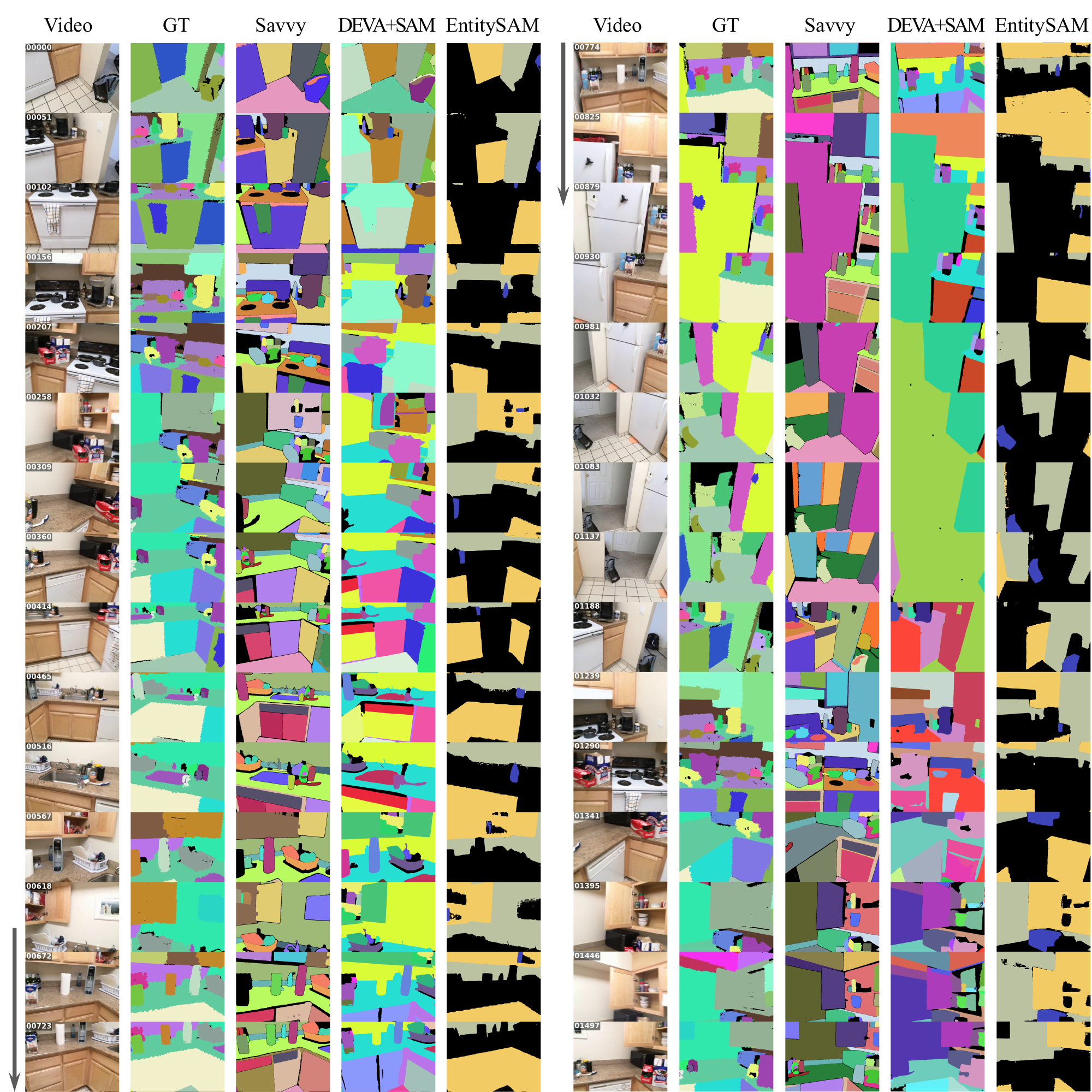}
    \caption{\small
\textbf{Long-horizon qualitative results on a cluttered ScanNet sequence (4).}
The sequence is split into two vertical blocks from left to right, following the same video over time.
The camera first observes cabinets, a stove, countertop objects, and floor regions, then moves toward a refrigerator and doorway area, and later returns to the kitchen workspace from different viewpoints.
Savvy maintains a detailed object decomposition in this cluttered scene: cabinets, countertop surfaces, stove regions, bottles, small kitchen items, refrigerator surfaces, door regions, and floor/wall structures remain more separated across the long trajectory.
DEVA+SAM captures many foreground regions but shows frequent local bleeding and blobbing: countertop objects merge into cabinets or counters, adjacent surfaces are inconsistently split, and small objects are often absorbed into larger regions as the camera moves.
EntitySAM produces much coarser partitions and limited new-object discovery; large surfaces such as cabinets, walls, and appliances dominate, while small kitchen objects and newly observed structures are frequently missed or assigned to broad existing regions.
}
\label{fig:scannet_qualitative_4}
\end{figure}

\begin{figure}[t]
    \centering
    \includegraphics[width=\linewidth]{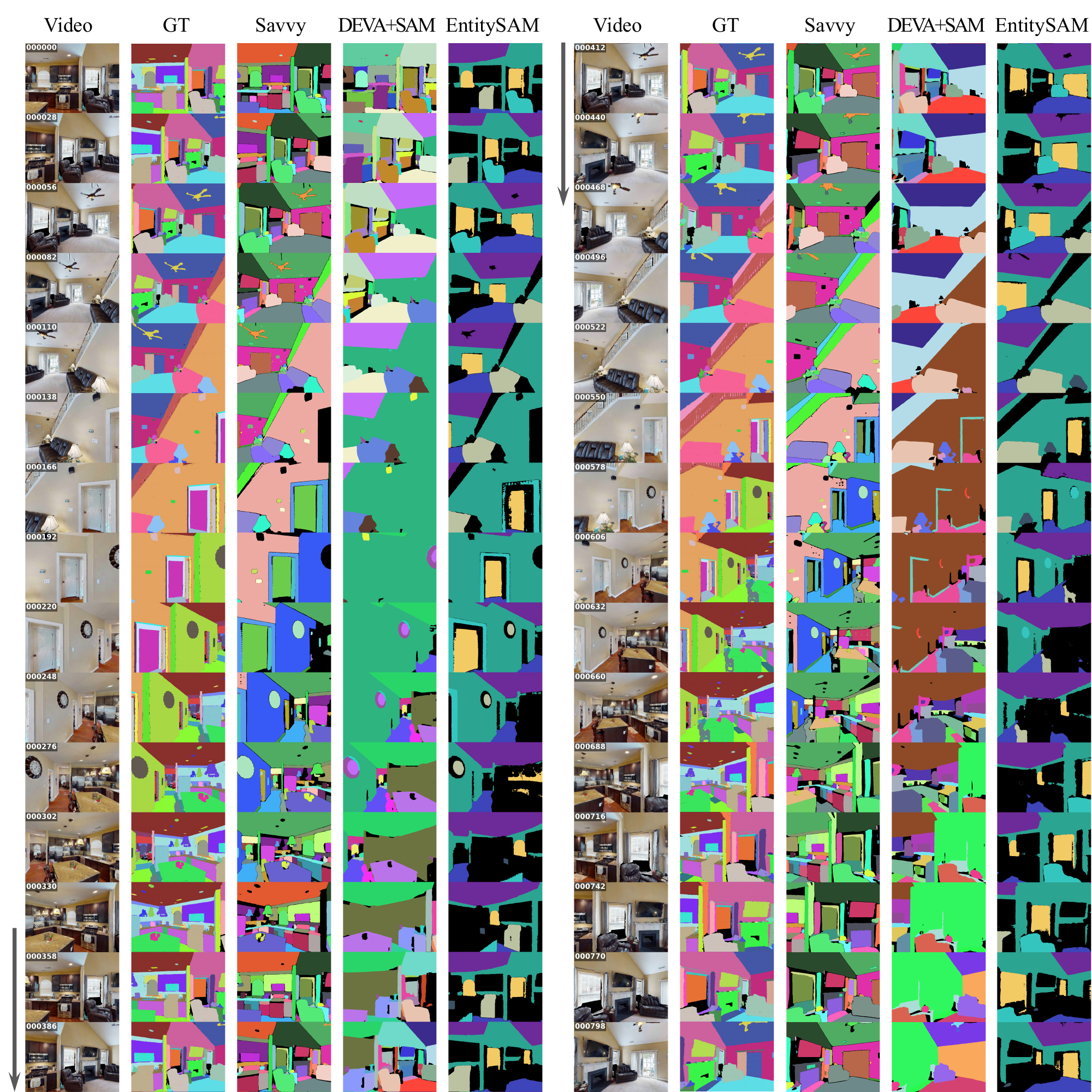}
    \caption{
\textbf{Long-horizon qualitative results on an HM3D indoor home sequence (1).}
The sequence is split into two vertical blocks from left to right, following the same video over time.
The camera moves through a living-room and kitchen area with large viewpoint changes, repeated observations of furniture, walls, doors, ceiling structures, counters, cabinets, and small household objects.
Savvy maintains a more detailed and temporally consistent scene decomposition across the trajectory: major room structures remain separated, while smaller objects and furniture regions are repeatedly recovered as the viewpoint changes.
DEVA+SAM provides broad coverage but shows unstable local structure, with large surfaces and furniture regions frequently bleeding into one another or changing decomposition across frames.
EntitySAM produces coarser partitions with limited object-set expansion; large surfaces are often stable, but newly observed furniture, cabinets, counters, and small objects are frequently missed as separate identities.
This example shows that the trends observed on ScanNet also hold in HM3D: long-horizon OVS requires not only foreground coverage, but also persistent discovery, stable granularity, and re-association across repeated scene revisits.
}
\label{fig:hm3d_qualitative_home}
\end{figure}

\begin{figure}[t]
    \centering
    \includegraphics[width=\linewidth]{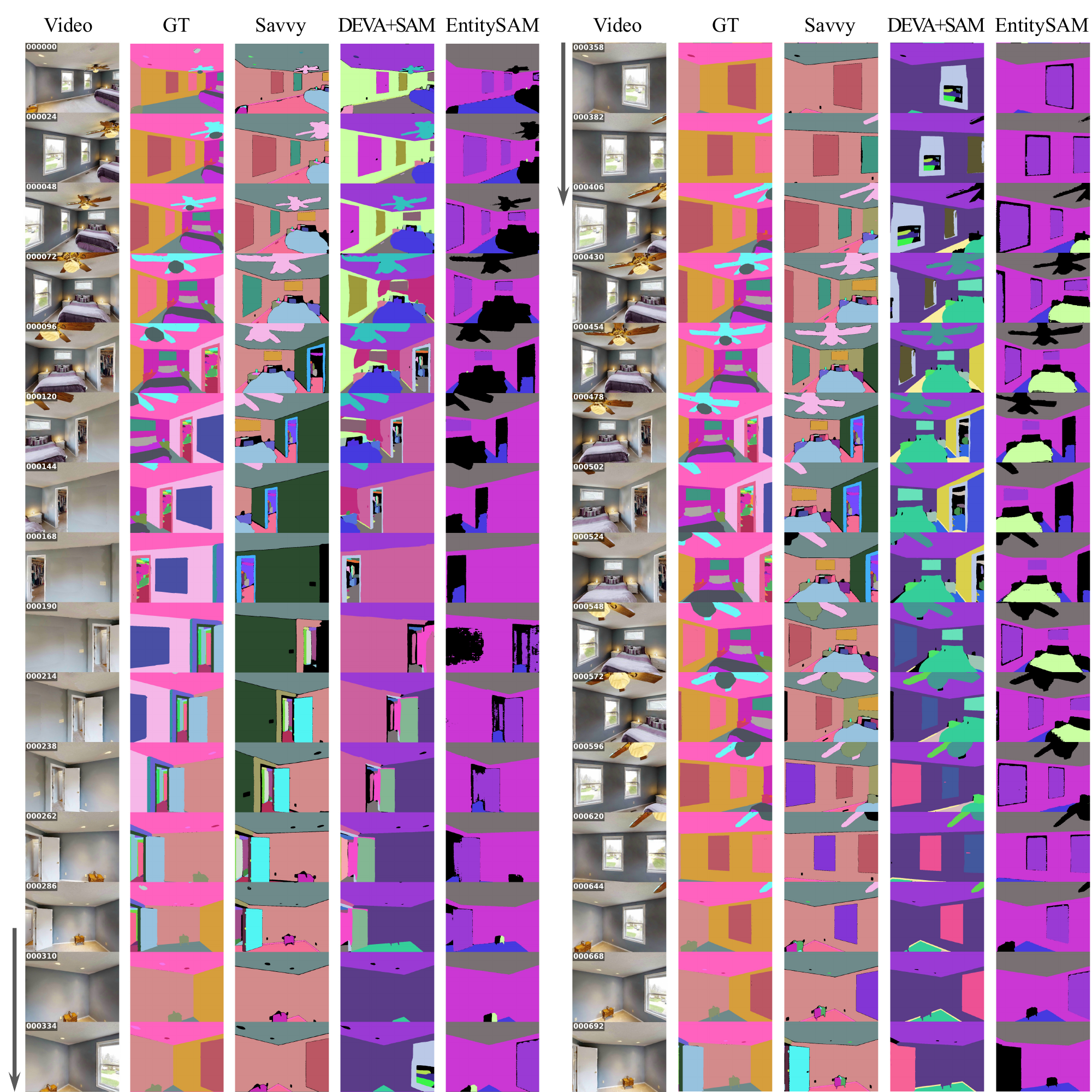}
    \caption{
\textbf{Long-horizon qualitative results on an HM3D bedroom sequence (2).}
The sequence is split into two vertical blocks from left to right, following the same video over time.
The camera moves around a bedroom with repeated observations of walls, windows, doors, ceiling fan, bed, lamps, pillows, and small room objects under large viewpoint changes.
Savvy maintains a relatively detailed and temporally consistent decomposition of both large room structures and smaller objects: wall and ceiling regions remain separated, while the bed, fan, door, and nearby objects are repeatedly recovered as the camera leaves and revisits the same regions.
However, window consistency remains challenging for all methods; the same physical window can be rediscovered with a new identity or switch identity across revisits.
DEVA+SAM provides broad coverage but shows severe local blobbing, bleeding, deformation, and unstructured fragmentation, especially around furniture, wall/ceiling boundaries, and window regions.
EntitySAM produces coarser partitions with limited object-set expansion; large surfaces remain dominant, while smaller objects such as the ceiling fan, lamp, pillows, and door/window details are often missed as separate identities or absorbed into broad regions.
This example further supports the HM3D trend: long-horizon OVS requires stable object-set maintenance across repeated views, not only coarse foreground coverage of large indoor surfaces.
}
\label{fig:hm3d_qualitative_bedroom}
\end{figure}

\begin{figure}[t]
    \centering
    \includegraphics[width=\linewidth]{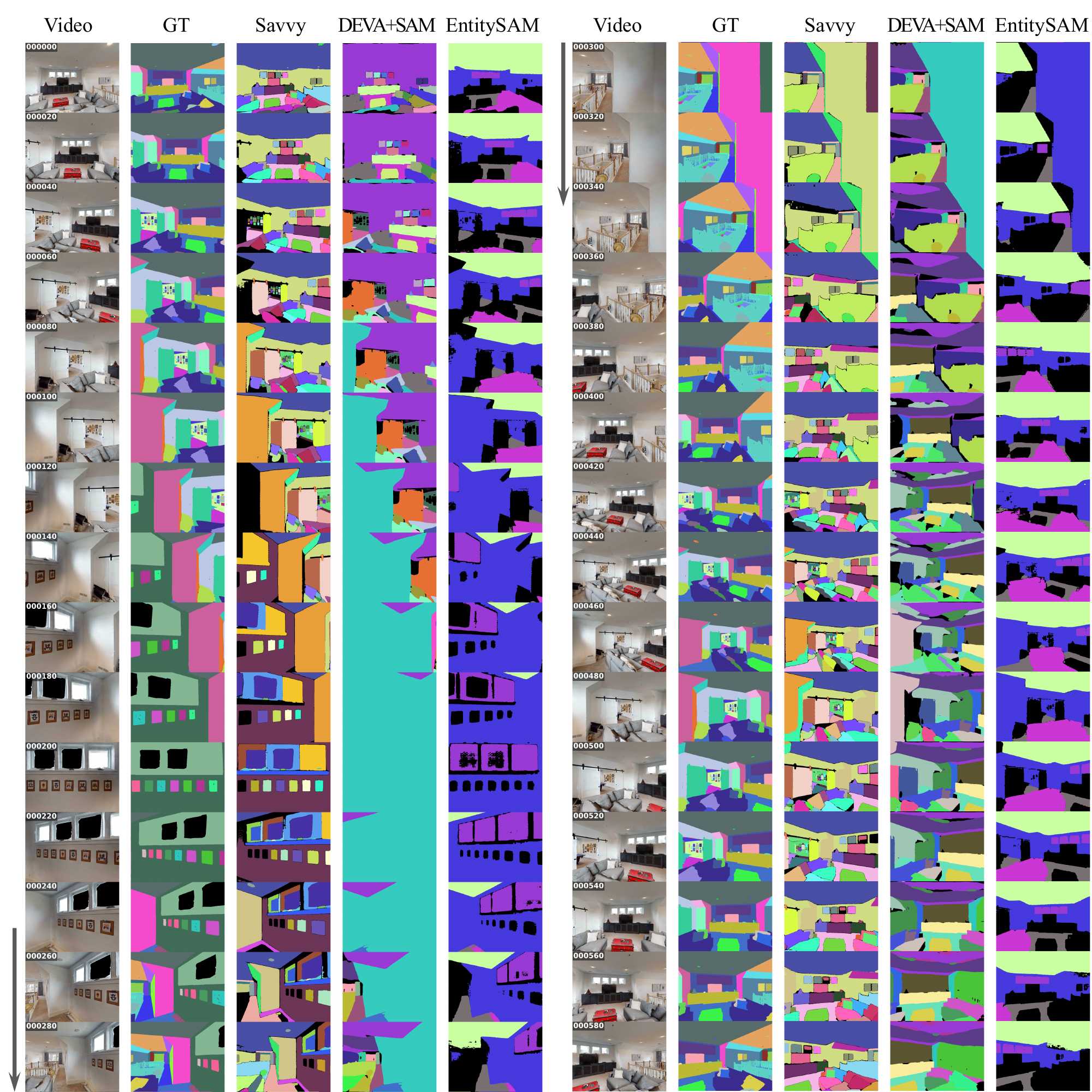}
    \caption{
\textbf{Long-horizon qualitative results on an HM3D living-room sequence.}
The sequence is split into two vertical blocks from left to right, following the same video over time.
The camera first observes a living-room area with couches, windows, wall decorations, tables, and shelves, then moves through a stairway/corridor view before returning to the living-room region from different viewpoints.
Savvy maintains a relatively detailed and temporally consistent object decomposition across these large viewpoint changes: couches, pillows, tables, shelves, framed wall objects, windows, and room surfaces are repeatedly recovered and kept more separated over time.
DEVA+SAM provides broad coverage but shows strong local instability, including blobbing, boundary bleeding, deformation of furniture regions, and inconsistent fragmentation of shelves, wall decorations, and couch/table areas.
EntitySAM produces much coarser partitions with limited object-set expansion; large surfaces and dominant furniture regions remain visible, but many smaller objects such as frames, pillows, shelves, and table-top items are missed as separate identities or absorbed into broad regions.
This example highlights a common HM3D challenge: long-horizon OVS must preserve detailed object structure while the camera alternates between wide room views, close-up wall/shelf views, and revisits to previously observed living-room regions.
}
\label{fig:hm3d_qualitative_livingroom}
\end{figure}
\clearpage

\section{Extended Related Work}
\label{app:extended_related_work}

\subsection{Task Assumptions and Open-World Benchmarks}
\label{app:rw_task_assumptions}
Open-world video understanding spans several distinct questions: how targets are specified, whether categories were seen during training, whether outputs require semantic names, when objects enter the tracked set, and how identities persist across observations. UVO~\cite{uvo} evaluates automatic class-agnostic segmentation alongside initialized object tracking. BURST~\cite{burst} organizes exemplar-guided and class-guided tasks, including an open-world setting; its exemplar prompts can arrive at each object's first appearance. These benchmarks establish complementary forms of openness. Savvy studies an externally unprompted video task: the evaluated HMD generates its own SAM point prompts, while the system decides which proposals become persistent identities. Our ScanNet and HM3D evaluation emphasizes continuing entry and re-observation during scene exploration, with category names removed from the output requirement.

\subsection{Prompted Propagation, Memory, and Long-Term Recovery}
\label{app:rw_prompted_propagation}
Semi-supervised video object segmentation studies reliable propagation of specified targets through occlusion and appearance change. STM~\cite{stm} retrieves information from space--time memory, AOT~\cite{aot} learns identity-aware association, and XMem~\cite{xmem} organizes memory across temporal scales. Cutie~\cite{cutie} combines pixel memory with object queries and compact object memory to improve object-level consistency. SAM2~\cite{sam2} provides promptable mask propagation with streaming memory. These mechanisms support the propagation stage of a persistent object system; the complementary system-level question is how newly observed regions enter and subsequently alter its object inventory.

Correspondence can also be strengthened by tracked points or geometry-aware features. SAM-PT~\cite{sampt} tracks positive and negative point prompts, uses visible trajectories to prompt SAM, and can resample points from predicted masks as the target's visible surface changes. 3AM~\cite{threeam} fuses MUSt3R's geometry-aware features with SAM2 and learns cross-view correspondence for prompted targets. Its inference inputs are RGB and target prompts, without externally supplied camera poses or depth. This makes it relevant to viewpoint changes and revisits even though its target specification differs from Savvy's autonomous discovery.

Long-video recovery also depends on which observations enter memory and when competing hypotheses are resolved. SAM2Long~\cite{sam2long} retains a bounded set of target-specific segmentation paths and selects a final path using accumulated confidence. DAM4SAM~\cite{dam4sam} separates recent target appearance from anchor memories that help resolve distractors and re-detect targets. Savvy operates at the level of object-set membership and identity: candidates accumulate evidence before promotion, merging, or rejection, and established tracks undergo reappearance checking, competition arbitration, and part--whole consolidation. Its lifecycle makes these decisions explicit around the chosen propagation backbone.

\subsection{Automatic Discovery and Learned Instance Association}
\label{app:rw_automatic_discovery}
DEVA~\cite{deva} decouples task-specific image segmentation from class-agnostic temporal propagation and fuses proposals across short clips in its semi-online mode. Both Savvy and our evaluated DEVA+SAM baseline are semi-online. Savvy's admission stage propagates candidates provisionally, resolves repeated same-ID duplicate agreement, and promotes or rejects the remaining candidates from accumulated evidence. Subsequent consolidation operates on established identities, including the stored histories of absorbed parts.

EntitySAM~\cite{entitysam} removes explicit user prompting and discovers entities that appear later in a video. VideoSAM~\cite{videosam} uses frozen RADIO features for association and trains autoregressive object prompts to stabilize granularity across frames. MASA~\cite{masa} learns transferable instance embeddings from SAM regions in augmented static images. Its SAM-based configuration uses a learned detection head to propose objects automatically, obtains masks, and matches embeddings to update tracks or create new ones; its association pipeline also removes duplicate proposals. VideoCUPS~\cite{videocups} trains a scene-centric video panoptic model on pseudo-labels generated from estimated motion, depth, and self-supervised visual features, providing an unsupervised route to instance discovery and association. These systems establish discovery and association as connected tasks. Savvy develops an explicit lifecycle for deciding when a discovered region merits a durable identity and when later evidence warrants merging or revising established tracks. VideoSAM's pursuit of consistent predicted granularity also complements OGA's question of how to evaluate a coherent partition that differs from the reference partition.

\subsection{Open-Vocabulary and Concept-Prompted Segmentation}
\label{app:rw_open_vocabulary}
OV2Seg~\cite{ov2seg} extends instance segmentation and tracking to category names beyond the training vocabulary. InstFormer, introduced for OpenVIS~\cite{openvis}, combines class-agnostic proposals with instance-level vision--language representations and temporal instance modeling. Such methods connect category generalization with video association. Savvy isolates the entity layer: it maintains masks and identities without requiring their names, allowing semantic interpretation to remain a separate downstream task.

SAM~3~\cite{sam3} unifies detection, segmentation, and tracking of instances specified by noun phrases or image exemplars. Its video system combines a detector with a memory-based tracker, delays confirmation of new tracks, removes duplicate tracks, and suppresses uncertain masks while retaining confirmed track state. Savvy couples unprompted discovery with provisional evidence accumulation and later consolidation of established tracks, including part--whole absorption that revises their stored output histories.

\subsection{Cross-View Association and Scene Representations}
\label{app:rw_cross_view}
Reconstruction-based instance segmentation reconciles observations in a shared spatial representation. SAI3D~\cite{sai3d} starts from a scene point cloud and posed RGB-D images, builds affinities between geometric primitives using multi-view SAM masks, and progressively merges primitives into 3D instances. Open3DIS~\cite{open3dis} grows regions under 2D mask guidance and hierarchically merges them across views, combining the resulting masks with proposals from a 3D network. OpenMask3D~\cite{openmask3d} separates class-agnostic 3D proposals from multi-view CLIP image-feature aggregation for open-vocabulary retrieval. These approaches connect image-level evidence to scene-level instances; Savvy produces per-frame masks and persistent IDs directly from RGB video.

Online scene methods maintain persistent object state as observations arrive. ESAM~\cite{esam} lifts SAM masks from posed RGB-D streams into geometry-aware queries, refines 3D masks, and associates current and accumulated instances using geometric, contrastive, and semantic similarities. Its matching stage merges matched masks and registers unmatched ones as new instances. ConceptGraphs~\cite{conceptgraphs} incrementally associates and fuses observations into an object-centric 3D map; after processing the image sequence, it adds captions and relations for language-based querying and planning. Savvy shares the need to reconcile repeated observations. Its later discoveries propagate provisionally before promotion, merging, or rejection, followed by consolidation of the established video track set.

A complementary line infers geometry or connectivity directly from RGB views. IGGT~\cite{iggt} jointly predicts camera parameters, depth, point maps, and instance-discriminative features from RGB views; jointly clustering the predicted instance features across views produces consistent instance masks. Multiview Scene Graph~\cite{msg} builds a topological graph from unposed images, learning place and object representations and maintaining an object memory bank through similarity-based updates and registration. IGGT couples instance correspondence with reconstruction over an available view set, while Multiview Scene Graph represents place and object connectivity. Savvy addresses continued discovery and semi-online identity revision in dense video segmentation. Together, these lines show how persistent entities can be represented through masks, geometric instances, or graph nodes, with different inputs and output objectives.

\subsection{Part Granularity and Flexible Instance Matching}
\label{app:rw_part_granularity}
Part-level segmentation makes the choice of decomposition explicit. PartSAM~\cite{partsam} predicts 3D part masks from shape representations using point prompts; its automatic Segment-Every-Part mode samples prompts and filters masks with a granularity-controlling suppression threshold. Meaningful predicted regions can therefore lie below the object scale chosen by a reference annotation. For video, OGA evaluates temporal support from finer predicted regions without requiring semantic part labels.

PartPQ~\cite{partpq} evaluates an annotated object--part hierarchy: parent segments follow panoptic matching, and semantic part overlap determines quality within matched parents. Closely related work by \citet{instance_matching2026} formalizes flexible panoptic matching with independently constrained graph degrees, including many-to-one matching that scores the union of predicted fragments. OGA carries the granularity question into temporal identity evaluation. It retains supporting prediction IDs, selects temporally coherent support through identity-sensitive severing, and pairs fidelity with overlap-based and graph-based diagnostics. A persistent subdivision can receive full fidelity while its distribution across prediction identities remains visible.

\subsection{Temporal Evaluation and Two-Sided Identity Diagnostics}
\label{app:rw_temporal_evaluation}
Panoptic quality (PQ)~\cite{pq} matches non-overlapping image segments one-to-one using a strict IoU threshold; VPQ~\cite{vpqvps} applies this principle to tubes within temporal windows. HOTA~\cite{hota} instead matches detections one-to-one in each frame, evaluates association over their full tracks, and averages across localization thresholds. STQ~\cite{stq} avoids thresholded assignment: its association term uses pixel overlaps across all intersecting prediction--reference track pairs over the full sequence, separately from semantic segmentation quality. VPQ, HOTA, and STQ already evaluate temporal identity consistency. OGA changes how coherent finer decompositions receive fidelity credit while preserving identities for temporal and structural evaluation. We retain conventional STQ and VPQ alongside OGA to make this protocol difference measurable.

OWTA~\cite{owta} replaces HOTA's detection accuracy with detection recall while retaining association penalties for matched tracks, and requires non-overlapping output masks. This accommodates wholly unmatched tracks that may correspond to unlabeled objects; BURST~\cite{burst} adopts it for open-world tracking. Annotation coverage and reference granularity are distinct issues. OGA restricts overlap accounting to the valid reference domain, while its many-to-one rule permits several predicted regions to support an annotated object. Ignored-pixel handling also appears in PQ~\cite{pq}; collective part support is a separate evaluation choice.

Segmentation covering~\cite{segcover} is a directed, area-weighted best-region overlap score, while variation of information~\cite{vi} compares partitions through conditional entropies that capture splits and merges. Applied independently to frames, these measures omit temporal identity. They can also operate on spatiotemporal partitions, capturing overlap-based fragmentation, but a contingency table alone does not encode the order of support states. OGA makes temporal support selection explicit and assigns each prediction to at most one reference within each GA fidelity scoring scope. Several prediction IDs may support the same reference, and their identities remain available to the two-sided IP/IC diagnostics.

The resulting distinction is between fidelity and the organization of support. In the worked example, two persistent equal parts retain perfect OGA fidelity while reference-side IP and IC decrease because support is distributed across two identities. Replacing an identity over time instead reduces temporal fidelity. Higher diagnostic scores indicate greater dominant-overlap purity or coverage and greater identity concentration; lower reference-side scores can describe a valid persistent decomposition. Joint interpretation of fidelity and diagnostics thus separates coherent part support, identity turnover, and prediction-side bleeding across references.

\clearpage
\subsection{Method Interfaces and Comparison Requirements}
\label{app:baseline_compatibility}

Task-level adaptation changes the system being evaluated: adding discovery or identity-management mechanisms yields a composite pipeline whose performance reflects both the published method and the added design. These modifications differ from routine data loading or output formatting; adapted comparisons require documenting and validating the added mechanisms.

\begin{table}[H]
\centering
\small
\setlength{\tabcolsep}{3pt}
\renewcommand{\arraystretch}{1.15}
\caption{\textbf{Published tracking and segmentation interfaces.} Inputs distinguish supplied target information from internally generated prompts. Temporal procedures identify how each cited configuration constructs masks and identities.}
\label{tab:baseline_compatibility}
\begin{tabular}{@{}>{\raggedright\arraybackslash}p{0.20\linewidth}>{\raggedright\arraybackslash}p{0.24\linewidth}>{\raggedright\arraybackslash}p{0.23\linewidth}>{\raggedright\arraybackslash}p{0.28\linewidth}@{}}
\toprule
\textbf{Method} & \textbf{Input and external specification} & \textbf{Object entry} & \textbf{Output and temporal procedure} \\
\midrule
Savvy & RGB; no supplied targets & Periodic discovery and provisional admission & Mask IDs; semi-online admission and later identity revision \\
\addlinespace[3pt]
DEVA+SAM\newline\cite{deva} & RGB; no supplied targets & Periodic SAM proposals & Mask IDs; short-clip consensus and propagation \\
\addlinespace[3pt]
EntitySAM\newline\cite{entitysam} & RGB; no supplied targets & Automatic entity queries & Mask IDs; single-pass propagation \\
\addlinespace[3pt]
MASA with SAM\newline\cite{masa} & RGB; internally generated box prompts & Automatic proposals and new-track creation & Potentially overlapping instance tracks; online matching \\
\addlinespace[3pt]
VideoSAM\newline\cite{videosam} & RGB; internally generated point prompts & Grid initialization and periodic re-detection & Mask tracklets; sequential propagation and memory-based association \\
\addlinespace[3pt]
SAM2 PVS\newline\cite{sam2} & RGB; target points, boxes, or masks & Target specification by prompts & Target masklets; streaming propagation \\
\addlinespace[3pt]
SAM 3 PCS\newline\cite{sam3} & RGB; concept text or image exemplars & New instances matching the supplied concept & Mask IDs; detection, propagation, delayed confirmation \\
\addlinespace[3pt]
OV2Seg / InstFormer\newline\cite{ov2seg,openvis} & RGB; vocabulary names for classification & Class-agnostic per-frame proposals & Mask IDs; memory-query / history-based association \\
\addlinespace[3pt]
ESAM\newline\cite{esam} & Posed RGB-D; no supplied target instances & Unmatched instances registered in a 3D map & Accumulated 3D masks; online instance fusion \\
\bottomrule
\end{tabular}
\end{table}

\paragraph{Comparison coverage.}
DEVA+SAM and EntitySAM are evaluated representatives of proposal--propagation fusion and learned entity discovery; SAM2 serves as the initialized-target VIPSeg control. MASA supports overlapping object/part masks and reports the absence of long-term memory~\cite[Appendices H, K]{masa}; a dense long-horizon comparison therefore requires explicit overlap resolution and revisit handling. VideoSAM couples RADIO correspondence, cycle-ack propagation with memory, trained autoregressive prompts, and periodic re-detection~\citep{videosam}. A matched comparison must preserve this full inference pipeline.

\paragraph{Task and output differences.}
OV2Seg and InstFormer generate class-agnostic instance masks and associate them across frames~\citep{ov2seg,openvis}; comparison with our dense outputs must specify category-independent track selection and overlap resolution into one prediction ID per pixel. SAM2's automatic data-engine procedure seeds masklets from a first-frame point grid~\citep{sam2}; continuing discovery additionally requires finding later arrivals and deciding which receive new IDs. SAM~3 searches for instances of supplied concepts~\citep{sam3}, making the query set part of the task specification. ESAM uses depth and camera poses in addition to RGB and outputs accumulated 3D instances~\citep{esam}; evaluating these as per-frame masks requires a defined 3D-to-2D projection.

\clearpage
\section{Details of Savvy}
\label{app:savvy_details}
\subsection{Pipeline Overview}
\label{app:savvy_pipeline}

Savvy (\textbf{S}egment \textbf{A}ny \textbf{V}ideo and e\textbf{V}er\textbf{Y}thing) is implemented as a semi-online object-set maintenance pipeline. Its object-state registry is a logical collection of track records maintained throughout the video. Each object has a persistent identity, prompts and propagation state, an appearance record consisting of an active descriptor and a historical appearance bank, and a lifecycle status indicating whether it is provisional or established. A promptable video model propagates registered tracks; our implementation uses SAM2. The registry is updated as objects are discovered, admitted, consolidated, or evicted. The following sections detail discovery, deferred admission, track consolidation, and long-video system controls.

\textbf{Propagate--verify--discover.} At initialization, Savvy bootstraps the object set either from external proposals or by querying a pluggable mask-discovery (MD) module on the first frame. Each initial mask is registered with the SAM2 video predictor as a new object, after which the system enters a propagate--verify--discover loop. Starting from the current frame, SAM2 propagates all active objects forward. For each propagated frame, Savvy stores the predicted masks, updates object presence and miss streaks, applies appearance consistency checks when objects reappear after long absences, resolves conflicts among overlapping established tracks, and periodically invokes MD in regions not already explained by the active set. Our experiments instantiate MD with hierarchical mask discovery (HMD; Section~\ref{app:hmd}).

\textbf{Deferred admission.} Initial seed masks are established immediately. Later discoveries enter a provisional buffer, where propagation accumulates visibility and duplicate-agreement evidence. Sufficient repeated agreement triggers early merging into an established identity; it does not wait for the window to expire. Remaining candidates are promoted or discarded at expiry according to their evidence. Terminal resolution applies the same evidence checks to unfinished windows at video end. These separate decisions avoid treating every discovery as a new established identity.

\textbf{Track consolidation.} Reappearance gating checks a returning track against its own appearance memory; competition arbitration suppresses a conflicting mask in the current frame. Both retain the track identity for later propagation. Part--whole absorption has a different effect: when a newly promoted whole contains established parts, it removes those part identities and remaps their complete stored output histories. Thus, observation rejection, frame-level competition, and permanent identity absorption are distinct lifecycle operations.

\textbf{Long-video system controls.} Old non-conditioning SAM2 outputs are pruned, stale masks can be stripped from predictor-side frame outputs, and an active-object budget limits predictor state. Capacity eviction considers established tracks first, then provisional tracks, ordered by increasing historical maximum area within each group. Available masks of recently evicted tracks temporarily exclude discovery prompts. These controls limit active predictor state. The complete output history is retained, and appearance-bank size can be bounded separately.

\subsection{MD Instantiation: Hierarchical Mask Discovery}
\label{app:hmd}

MD is Savvy's architectural interface for proposing masks in regions not explained by the current object set; it does not prescribe a particular segmenter or proposal hierarchy. Our experiments instantiate this interface as \emph{Hierarchical Mask Discovery} (HMD). Rather than running an image segmenter over the full frame and directly registering every proposal, HMD identifies unexplained regions, samples prompts only there, generates multi-granularity SAM proposals, distills them through hierarchical merging, and passes only surviving candidates to deferred admission.

\paragraph{Unexplained-region prompting.}
At a discovery frame, Savvy constructs an occupied-region mask from the current established tracks. In the default non-exploratory mode, transient tracks are also included so that recently discovered candidates are not immediately rediscovered. Recently evicted tracks can further be retained as temporary suppression regions. If the occupied mask covers almost the entire frame, discovery is skipped. Otherwise, HMD samples a regular grid of prompt points from the remaining unexplained region and queries the image-level segmenter only at those locations.

\paragraph{Multi-granularity SAM proposals.}
HMD uses an adapted SAM1 automatic mask generator. For each prompt point, SAM produces multiple candidate masks through its multimask output. Instead of flattening these outputs into a single proposal set, HMD keeps their scale structure and groups them into small-, medium-, and large-level proposal sets, denoted by $\mathcal{M}^{\textsc{S}}$, $\mathcal{M}^{\textsc{M}}$, and $\mathcal{M}^{\textsc{L}}$. Each proposal is filtered by SAM's predicted IoU and stability score, and is assigned a quality score used for downstream competition.

\paragraph{Hierarchical mask merging.}
The core of HMD is a survival-based merging operator. Given two proposal sets $\mathcal{M}_1$ and $\mathcal{M}_2$, let $\mathcal{M}=\mathcal{M}_1\cup\mathcal{M}_2$ be the combined candidate set. Each mask $m_k\in\mathcal{M}$ has binary support $\chi_k(x)\in\{0,1\}$ and quality score $q_k$. HMD assigns each pixel to the highest-quality proposal covering it:
\begin{equation}
    A(x)=\arg\max_{k:m_k\in\mathcal{M}} q_k \chi_k(x),
\end{equation}
with pixels not covered by any proposal assigned to background.

After this winner-take-all arbitration, a proposal is retained only if it keeps enough of its original support. Let
\begin{equation}
    \Omega_k = \{x \mid \chi_k(x)=1\}, 
    \qquad
    \widehat{\Omega}_k = \{x \mid A(x)=k\}.
\end{equation}
The retained area ratio is
\begin{equation}
    r_k = \frac{|\widehat{\Omega}_k|}{|\Omega_k|}.
\end{equation}
Proposal $m_k$ survives if $r_k \geq \tau_{\mathrm{surv}}$ and $|\widehat{\Omega}_k|>0$; otherwise, its assigned pixels are released to background. This filtering suppresses unstable fragments that are almost entirely overwritten by stronger overlapping masks, while preserving proposals that retain sufficient unique support.

HMD applies this operator recursively across proposal scales:
\begin{align}
    \mathcal{M}^{\textsc{sm}} &= \operatorname{Merge}(\mathcal{M}^{\textsc{s}}, \mathcal{M}^{\textsc{m}}; \tau_{\mathrm{surv}}), \\
    \mathcal{M}^{\textsc{hmd}} &= \operatorname{Merge}(\mathcal{M}^{\textsc{sm}}, \mathcal{M}^{\textsc{l}}; \tau_{\mathrm{surv}}).
\end{align}
The resulting $\mathcal{M}^{\textsc{hmd}}$ defines the image-level proposal partition. At initialization, its nonempty masks are registered directly as established seeds. At later discovery steps, they undergo the candidate-survival and geometric filters below before entering deferred admission. If only one proposal group remains and no cross-group merge is needed, the implementation rasterizes it using area-prioritized overlap arbitration.

\paragraph{Mask quality score.}
For each SAM proposal, HMD uses a quality score that combines SAM's predicted mask quality with a log-scaled area prior:
\begin{equation}
    q_k = \left(1+\log(1+|\Omega_k|)\right)c_k,
\end{equation}
where $c_k$ is SAM's predicted IoU score and $|\Omega_k|$ is the mask area. The logarithmic area term discourages small high-confidence texture fragments from dominating larger stable regions, while remaining sublinear so that large masks do not automatically suppress meaningful fine-scale proposals. In practice, this score is used in the pixel-level arbitration described above. \paragraph{Scale-dependent scoring.}
The recursive formulation permits different reliability scores or survival thresholds at different scale transitions. The default implementation uses the same confidence-area quality score at both merge stages.

\paragraph{Candidate survival against active tracks.}
After initialization, HMD proposals are filtered before provisional registration. This filtering uses the raw propagated masks and the current transient masks as pixel-level vetoes. A candidate keeps only unclaimed pixels and must retain at least a fraction $\tau_{\mathrm{cand}}$ of its proposal area. This veto differs from the prompt-exclusion mask: recent-eviction masks restrict prompt locations but are not added to the pixel-level survival veto. The hierarchy threshold $\tau_{\mathrm{surv}}$ and candidate-survival threshold $\tau_{\mathrm{cand}}$ are separate controls, both with implementation default $0.1$.

\paragraph{Additional filters.}
Savvy applies two lightweight geometric filters before inserting a discovered mask as a transient track. First, sliver-like masks are removed using compactness and bounding-box aspect-ratio checks. Second, a border-delay rule suppresses masks whose centroid lies near the image boundary, where objects are often only partially visible and likely to produce unstable identities. Candidates that pass these checks are inserted as transient tracks and must still pass deferred admission before becoming established objects.

\begin{algorithm}[t]
\begin{algorithmpanel}
\caption{HMD Proposals and Post-Initialization Candidate Filtering}
\label{alg:hmd}
\begin{algorithmic}[1]
\Require Frame $I_t$; segmenter $\Phi$; object registry; raw propagated masks;
\Statex \hspace{\algorithmicindent} hierarchy and candidate thresholds $\tau_{\mathrm{surv}},\tau_{\mathrm{cand}}$
\Ensure Candidate masks $\mathcal{C}_t$
\Statex
\State Build $U_t^{\mathrm{prompt}}$ from accepted, provisional, and recent-eviction masks
\If{$U_t^{\mathrm{prompt}}$ nearly covers the frame}
    \State \Return $\emptyset$
\EndIf
\State Sample prompt points $\mathcal{P}_t$ from $\overline{U_t^{\mathrm{prompt}}}$
\State $(\mathcal{M}^{\text{\textsc{s}}}, \mathcal{M}^{\text{\textsc{m}}}, \mathcal{M}^{\text{\textsc{l}}}) \gets \Phi(I_t,\mathcal{P}_t)$
\Statex \emph{Empty proposal groups bypass their corresponding merge.}
\State $\mathcal{M}^{\text{\textsc{sm}}} \gets \operatorname{Merge}(\mathcal{M}^{\text{\textsc{s}}}, \mathcal{M}^{\text{\textsc{m}}}; \tau_{\mathrm{surv}})$
\State $\mathcal{M}^{\text{\textsc{hmd}}} \gets \operatorname{Merge}(\mathcal{M}^{\text{\textsc{sm}}}, \mathcal{M}^{\text{\textsc{l}}}; \tau_{\mathrm{surv}})$
\If{initialization}
    \State \Return nonempty proposal masks for established seed registration
\EndIf
\State Form $U_t^{\mathrm{surv}}$ from raw propagated and current provisional masks
\State $\mathcal{C}_t \gets \emptyset$
\For{$m \in \mathcal{M}^{\text{\textsc{hmd}}}$}
    \State $m' \gets m \setminus U_t^{\mathrm{surv}}$
    \If{$|m'|>0$, $|m'|/|m| \geq \tau_{\mathrm{cand}}$, and $m'$ passes geometric filters}
        \State $\mathcal{C}_t \gets \mathcal{C}_t \cup \{m'\}$
    \EndIf
\EndFor
\State \Return $\mathcal{C}_t$
\end{algorithmic}
\end{algorithmpanel}
\end{algorithm}

\subsection{Deferred Admission and Transient Buffering}
\label{app:deferred_admission}

Open-world discovery is noisy: an image-level proposal may correspond to a genuinely new object, a transient fragment, a partially visible boundary object, or a duplicate view of an already tracked entity. Savvy therefore separates \emph{discovery} from \emph{admission}. After bootstrap, new candidates produced by HMD are not immediately added to the established object set. Instead, they are first inserted as transient tracks and must pass a deferred-admission test before receiving permanent status.

\paragraph{Transient track state.}
Each transient track stores a candidate identity, its entry frame, visibility evidence, maximum observed area, and duplicate-agreement statistics with established tracks. Let $z$ denote a transient track. During its deferred-admission window, Savvy records
\begin{equation}
    v(z), \qquad a_{\max}(z), \qquad h(z), \qquad b(z),
\end{equation}
where $v(z)$ is the number of visible frames, $a_{\max}(z)$ is the maximum mask area observed during the window, $h(z)$ is the number of repeated agreement hits with its best matched established object, and $b(z)$ is the identity of that best matched established object if one exists.

\paragraph{Duplicate agreement checking.}
At each frame, Savvy compares visible transient tracks against visible established tracks. For a transient $z$ and established track $o$, agreement is measured by mask IoU:
\begin{equation}
    \mathrm{IoU}_t(z,o)
    =
    \frac{|M_{z,t}\cap M_{o,t}|}{|M_{z,t}\cup M_{o,t}|}.
\end{equation}
The best visible established match is eligible when its IoU is at least $\tau_{\mathrm{agree}}$. A first eligible match initializes the stored target and hit count; further hits for the same ID accumulate, without requiring consecutive frames. No eligible match leaves the evidence unchanged. A different target replaces the stored one only when its current IoU is strictly greater than $1.25$ times the historical peak IoU for the stored target; switching resets the count to one.

As soon as $h(z)\geq\tau_{\mathrm{hit}}$, the candidate is merged into $b(z)$, \emph{without waiting for buffer expiry}. Its buffered output masks are unioned into the established identity, while the established predictor memory is retained and the transient predictor state is removed. The implementation defaults to $\tau_{\mathrm{agree}}=0.5$ and $\tau_{\mathrm{hit}}=5$.

\paragraph{Expiry-based admission.}
After agreement-driven merging, a remaining transient is eligible for resolution when $t-t_{\mathrm{entry}}\geq B$. Define $G_{\mathrm{area}}(z)$ to require every enabled absolute or image-relative maximum-area threshold; with both thresholds disabled it is always true. The ordered expiry decision is
\begin{equation}
    z \rightarrow
\begin{cases}
\text{discard residual duplicate}, 
& \text{if } b(z)\neq \varnothing \text{ and } h(z)\geq \tau_{\mathrm{hit}},\\
\text{promote}, 
& \text{else if } v(z)\geq \tau_{\mathrm{vis}} \text{ and } G_{\mathrm{area}}(z),\\
\text{discard}, 
& \text{otherwise}.
\end{cases}
\end{equation}
The first branch prevents a candidate with residual duplicate evidence from being promoted; expiry itself does not invoke a merge. The second admits candidates with sufficient observed support, and the last removes weak proposals. The implementation defaults are $B=30$ and $\tau_{\mathrm{vis}}=3$; both promotion-area gates are disabled by default. Visibility is accumulated from nonempty propagated masks, rather than counting registration itself as an observation.

\paragraph{Backfilling promoted tracks.}
When a transient is promoted, Savvy backfills its admission-buffer masks into the final prediction sequence. This prevents the object from appearing only after the end of the deferred-admission window. In other words, the buffer delays the identity decision, not the temporal support credited to a valid object. If the transient is instead merged into an established object, its buffered masks can be remapped to the established identity when they provide additional support. A transient contributes only its admission-buffer history; this differs from established-part absorption, which remaps the part's entire stored output history (Sec.~\ref{app:track_consolidation}).

\paragraph{Terminal resolution.}
After the last video frame, terminal resolution reuses the expiry decision for every unresolved candidate, waiving \emph{only} the elapsed-window requirement. Duplicate evidence still has discard priority, and visibility and any enabled area gates still apply. The handler repeats until pending absorptions and deferred candidate resolutions have been drained. It neither unconditionally promotes candidates nor reruns the agreement-merging stage. For a terminally promoted track, presence/absence counters use its actual trailing buffered observations rather than assuming a full-window history.

\paragraph{Role of deferred admission.}
Immediate admission creates a permanent object identity from a single image-level proposal. In long videos, this causes rapid identity growth from clutter, partial views, and duplicate discoveries. Deferred admission turns object registration into an evidence-accumulation process: candidates are promoted, merged, or discarded according to their accumulated visibility, duplicate agreement, and any enabled area tests. This design is especially important in long-horizon OVS, where the number of potential proposals grows over time and greedy admission can lead to severe ID fragmentation.

\subsection{Track Consolidation}
\label{app:track_consolidation}

Deferred admission reduces noisy object creation, but long-horizon OVS still faces a second challenge: the established object set itself can become inconsistent over time. Reappearing objects may drift away from their historical appearance, multiple established tracks may compete for the same region, and part-level and whole-level tracks may coexist redundantly. Savvy addresses these issues through \emph{track consolidation}, which continuously regularizes the active object set after tracks have been promoted.

Track consolidation is applied to established tracks, rather than transient candidates. This distinction is important: deferred admission decides whether a new candidate deserves a persistent identity, while track consolidation decides whether existing persistent identities remain mutually consistent and useful. In Savvy, consolidation has three components.

First, \textbf{appearance self-consistency gating} checks whether a returning object remains compatible with its own appearance record. The check precedes the appearance update, so a rejected reappearance cannot validate itself by entering the record first. This is a within-ID consistency check, not a search over other identities for a new correspondence.

Second, \textbf{competition arbitration} resolves high-overlap conflicts among established tracks. When both appearance scores exist, a clear score difference decides the winner; otherwise their presence counters and then seniority break the tie. If either appearance score is missing, arbitration uses seniority directly. Only the losing frame's mask is suppressed, not its propagation identity.

Third, \textbf{part--whole absorption} is triggered when a newly promoted whole contains redundant established parts under the geometry and containment tests below. Their stored masks are unioned into the whole identity across the output history, and their predictor states are removed. The reverse situation is not handled by an unconditional containment rule: a new candidate overlapping an established whole follows discovery filtering and the IoU-based duplicate-agreement path. Containment alone does not guarantee its identity will be merged.

Together, these mechanisms make the established object set self-regularizing. Rather than relying on SAM2 propagation alone, Savvy continually checks whether object identities remain appearance-consistent, mutually non-redundant, and structurally compact. This is essential in long videos, where small local errors can otherwise accumulate into persistent ID duplication, drift, or fragmentation.

\subsubsection{Appearance Self-Consistency Gating}
\label{app:appearance_consistency}

Appearance self-consistency gating is designed to prevent long-gap identity drift. In long-horizon videos, an object may disappear for many frames and later reappear under a different viewpoint. A propagation backbone may also re-activate an object on an unrelated region with a plausible mask. Savvy therefore does not accept every reappearance purely from propagation. Instead, it checks whether the current observation is compatible with the track's own historical appearance record.

For each established object $o$, Savvy maintains an active exponential moving average (EMA) and a bank of committed appearance descriptors. Nonempty accepted observations update the active EMA. When the absence streak reaches $B$ frames, the EMA is committed to the bank and cleared; shorter gaps do not end the appearance episode. The bank therefore aggregates episodes rather than storing one descriptor per frame, but its size is unbounded unless a bank-size limit is configured.

Given a mask $M_{o,t}$, Savvy computes its bounding box and expands it by a factor of $1.5$ for local context; masks occupying more than $15\%$ of the image use the tight box. It averages features within the corresponding box at each available SAM2 FPN level and concatenates the pooled vectors:
\begin{equation}
    \phi_{o,t}
    =
    \operatorname{Concat}_{\ell}
    \operatorname{GAP}
    \left(
        F^{\ell}_{t}
        \big[
        B_{\ell}(M_{o,t})
        \big]
    \right),
\end{equation}
where $F^{\ell}_{t}$ is the $\ell$-th backbone feature level at frame $t$, $B_{\ell}(M_{o,t})$ is the mask bounding box mapped to that feature resolution, and $\operatorname{GAP}$ denotes global average pooling.

Within an appearance episode, the first descriptor initializes the EMA and subsequent accepted observations update it:
\begin{equation}
    e_{o,t}
    =
    \alpha e_{o,t-1} + (1-\alpha)\phi_{o,t}.
\end{equation}
Let $\mathcal{M}_o$ contain the committed bank descriptors and, when present, the active EMA. The comparison score is
\begin{equation}
    s(o,t)
    =
    \max_{e\in\mathcal{M}_o}
    \operatorname{cos}
    \left(
        \phi_{o,t}, e
    \right).
\end{equation}
For a track returning after at least $B$ absent frames, this score is computed \emph{before} updating the active EMA. If a score exists and $s(o,t)<\tau_{\mathrm{app}}$, the current mask is suppressed and its miss streak continues. Otherwise, including when no comparison memory is available, the observation is accepted, the miss streak is reset, and the active EMA is updated. Subsequent competition uses the updated appearance state.

This gating is applied only when an established track reappears after a long enough absence. It therefore does not over-constrain normal short-term propagation, where appearance may change smoothly from frame to frame. Instead, it targets the failure mode most harmful to long-horizon OVS: an old identity being incorrectly revived on a different object or background region after a long temporal gap.

\subsubsection{Competition Arbitration}
\label{app:competition_arbitration}

Even after deferred admission, established tracks may occasionally compete for the same image region. This can happen when two identities become redundant, when a previous object drifts into a nearby region, or when SAM2 propagation produces overlapping masks for objects with similar visual support. Savvy resolves these conflicts through competition arbitration, a frame-level suppression rule applied only among established tracks.

For each frame $t$, let $\mathcal{O}_t$ denote the set of established tracks with non-empty masks. Savvy considers pairs of established tracks $(o_i,o_j)$ whose masks overlap substantially:
\begin{equation}
    \operatorname{IoU}_t(o_i,o_j)
    =
    \frac{|M_{i,t}\cap M_{j,t}|}{|M_{i,t}\cup M_{j,t}|}.
\end{equation}
If $\operatorname{IoU}_t(o_i,o_j) < \tau_{\mathrm{comp}}$, the two tracks are allowed to coexist. Otherwise, Savvy treats the pair as a competition for the same region and suppresses one of the two masks at frame $t$.

Let $s_i,s_j$ be the appearance scores against committed descriptors and any active EMA, and let $\ell_i,\ell_j$ be presence counters. Counters increment on accepted presence and reset on absence or suppression; ordinary promotion initializes its counter to $B$, whereas terminal promotion uses observed trailing support. For a competing pair, the suppressed identity is
\begin{equation}
    L(o_i,o_j)=
    \begin{cases}
    \arg\min_{o\in\{o_i,o_j\}} s_o,
    & \text{if } s_i,s_j \text{ exist and } |s_i-s_j|\geq \delta,\\[3pt]
    \arg\min_{o\in\{o_i,o_j\}} \ell_o,
    & \text{if } s_i,s_j \text{ exist},\ |s_i-s_j|<\delta,\ \ell_i\neq \ell_j,\\[3pt]
    \arg\max_{o\in\{o_i,o_j\}} \operatorname{id}(o),
    & \text{otherwise}.
    \end{cases}
\end{equation}
The first branch prefers the larger appearance score. The second applies only when both scores exist but are close, and prefers the larger presence counter. The last branch prefers the smaller (older) ID, including whenever either appearance score is unavailable.

This arbitration rule is intentionally conservative. It does not globally merge identities or permanently delete the losing track; instead, it suppresses the losing mask in the current frame and updates its miss streak. This allows a track to recover later if it becomes spatially separated and appearance-consistent again. Thus, competition arbitration removes frame-level duplicate support without prematurely destroying potentially valid long-term identities.

\subsubsection{Part-Whole Absorption}
\label{app:part_whole_absorption}

Open-world segmentation does not assume a single fixed semantic granularity. As a result, Savvy may encounter both part-level and whole-level hypotheses for the same physical entity. For example, a chair seat may be discovered before the full chair, or a cabinet door may be tracked before the entire cabinet becomes visible. Without additional consolidation, both tracks may remain active, causing redundant identities and unnecessary competition. Savvy addresses this through part-whole absorption.

Absorption tests established parts against a \emph{newly promoted} whole. This directional rule differs from early transient duplicate resolution, which requires repeated IoU agreement. In particular, a small contained candidate need not meet that IoU threshold merely because an established whole covers it.

For a newly promoted whole $o_w$, established candidates are considered in descending mask area. A pair must have whole-to-part area ratio at least $1.5$ and either bounding-box IoU at least $0.1$ or intersection divided by the part's bounding-box area at least $0.3$. Passing pairs are tested by pixel-level containment:
\begin{equation}
    \operatorname{IoS}(o_p,o_w)
    =
    \frac{|M_{p,t}\cap M_{w,t}|}{|M_{p,t}|}.
\end{equation}
The established track $o_p$ is considered a contained part of $o_w$ if
\begin{equation}
    \operatorname{IoS}(o_p,o_w) \geq \tau_{\mathrm{contain}}.
\end{equation}
When this condition is satisfied, $o_p$ is absorbed into $o_w$: its entire stored output history is remapped into the whole identity, and the part identity is removed from the active object set. This retrospective identity remapping is not restricted to the admission buffer; it also relabels the part's earlier output masks.

This operation is intentionally asymmetric. The gates restrict absorption to substantially covered tracks smaller than the newly promoted whole; mere overlap is insufficient. This decision uses projected mask geometry rather than semantic part labels.

In implementation, containment is checked after promotion of a transient into an established track. If multiple contained parts are found, Savvy absorbs them sequentially across pipeline restarts to keep the SAM2 object state consistent.

\subsection{Memory Management and Active-Track Control}
\label{app:memory_management}

Long-horizon OVS requires maintaining an expanding object set over many frames. Without explicit control, the number of active tracks and stored propagation states can grow quickly, leading to excessive memory use and repeated rediscovery of weak objects. Savvy therefore includes lightweight memory management and active-track control mechanisms. These mechanisms make continued object discovery computationally practical.

\paragraph{SAM2 memory pruning.}
Savvy prunes non-conditioning predictor entries older than $t-n_{\mathrm{mem}}s_{\mathrm{mem}}-1$, using the propagator's memory length and temporal stride. It also strips stored prediction-mask tensors from non-conditioning entries more than ten frames old while retaining their other memory features. These predictor-side operations are distinct from the stored output history used for retrospective identity remapping.

\paragraph{Active-track budget.}
When the active SAM2 object count exceeds its budget, Savvy evicts established tracks first and provisional tracks only if necessary. Within each group, eviction proceeds in increasing maximum historical mask area, rather than current area. This preserves information that a currently small mask may have had substantial earlier support. Capacity eviction removes predictor state and bookkeeping but preserves already stored established output masks.

\paragraph{Recent-eviction suppression.}
If a mask is available at the eviction frame, Savvy retains it as a prompt-exclusion region until $t+3s$, where $s$ is the discovery stride. This short cooldown discourages immediate rediscovery. It restricts prompt locations, not all pixels of subsequently generated proposals, and is separate from candidate-survival filtering.

\paragraph{Adaptive discovery density.}
Savvy optionally adapts the SAM grid density used by HMD. In cluttered scenes, dense prompting may produce excessive fragmented proposals, while in cleaner scenes a denser grid can improve discovery coverage. Savvy therefore tracks the recent number of generated masks and reduces the grid density when proposal counts become too high; it gradually restores the density when the scene becomes less cluttered. This feedback mechanism reduces mask explosions without disabling discovery altogether.

\paragraph{Role in the overall pipeline.}
These controls limit active predictor state rather than the complete output archive. Alongside deferred admission, they manage the computation associated with continued discovery; they do not impose a fixed video duration or change the evaluation's identity definitions.

\subsection{Implementation Details and Hyperparameters}
\label{app:implementation_details}

Savvy is implemented as a semi-online pipeline built on a SAM2 video predictor for mask propagation. Its generic MD interface is instantiated with HMD, using an adapted SAM1 automatic mask generator for image-level proposals. We use SAM1 ViT-H for HMD and SAM2.1 Hiera-Large as the video propagation backbone. The SAM2 predictor maintains object-specific propagation states, while HMD is queried periodically to propose new candidate masks in unexplained regions. Unless otherwise stated, all methods are evaluated with the same output format and the same OGA evaluation pipeline. All experiments are performed on NVIDIA A6000 GPUs. We report detailed runtime and memory profiling in Sec.~\ref{app:runtime_memory}.

Savvy propagates the active object set, checks returning and competing tracks, resolves duplicate candidates, and then handles expiry and absorption. Memory control precedes discovery. A merge, absorption, discard, or capacity eviction restarts propagation at the next frame, deferring any later stages of the current frame; discovery is therefore stride-scheduled but not unconditional. Semi-online output revisions have two scopes: transient promotion or merging contributes admission-buffer history, whereas established-part absorption remaps the part's entire stored output history. Terminal resolution handles unfinished windows after the final frame.

Table~\ref{tab:savvy_hyperparams} reports the experimental settings.

Algorithm~\ref{alg:savvy_inference} preserves this restart-aware ordering while abstracting predictor reindexing. $\mathcal{O}$ denotes established tracks, $\mathcal{B}$ provisional tracks with buffered evidence, $\mathcal{A}$ appearance memory, and $\mathcal{Y}$ stored output. $\mathrm{MergeDuplicates}$ updates agreement evidence and applies early merges. $\mathrm{ResolveCandidates}$ first drains a queued absorption if present; otherwise it evaluates candidate expiry, backfills promotions, checks newly promoted wholes for contained parts, and discards rejected candidates. It returns true when a predictor-track removal requires a restart. At termination, its finalization flag waives candidate age but retains every evidence test. Absorption always remaps the complete stored output history; candidate resolution uses only buffered masks.

\subsection{Runtime and Memory Profiling}
\label{app:runtime_memory}

We profile Savvy and both long-horizon baselines on the same 30-scene stratified ScanNet minival subset (12,803 frames). The Savvy run uses its full configuration, including HMD, SAM2 propagation, deferred admission, consolidation, and memory control. Frame-weighted throughput uses per-frame tracking time; Savvy timing includes mask propagation and CUDA synchronization. Loading, saving, visualization, and evaluation are excluded. Across sequences $s$, we report
\begin{equation}
    \mathrm{FPS}_{\mathrm{fw}}
    =
    \frac{\sum_s \text{processed frames}_s}
    {\sum_s \text{tracking seconds}_s}.
\end{equation}

Table~\ref{tab:matched_efficiency} reports the matched-scene comparison. Savvy processes more masks per frame and maintains the largest final object set, reflecting its sustained-discovery objective. It is slower than DEVA+SAM and EntitySAM, but has lower mean per-sequence peak GPU memory: 6,631 MB versus 14,484 MB and 20,527 MB, respectively. Its frame-weighted FPS is lower than its mean per-sequence FPS because object-heavy scenes take disproportionately long. DEVA+SAM timings are restricted to the same scenes; EntitySAM object counts are computed from its output instance maps. GPU memory measurements may include framework allocation overhead.

\begin{table}[t]
\centering
\small
\setlength{\tabcolsep}{5.5pt}
\caption{\textbf{Tracking-only profile on the 30-scene ScanNet minival subset.} All methods cover the same 12,803 frames. GPU memory is the mean per-sequence peak, not the maximum across scenes.}
\label{tab:matched_efficiency}
\resizebox{\linewidth}{!}{%
\begin{tabular}{lrrrrr}
\toprule
\textbf{Method} & \textbf{Frame-wt. FPS $\uparrow$} & \textbf{Mean seq. FPS $\uparrow$} & \textbf{Masks/frame} & \textbf{Final object set} & \textbf{Mean peak GPU (MB) $\downarrow$} \\
\midrule
EntitySAM & 1.97 & 1.97 & 3.71 & 4.57 & 20,527 \\
DEVA+SAM & \textbf{4.31} & \textbf{5.51} & 13.40 & 91.03 & 14,484 \\
Savvy (Ours) & 1.34 & 1.84 & \textbf{16.51} & \textbf{123.47} & \textbf{6,631} \\
\bottomrule
\end{tabular}}
\end{table}

\begin{table}[t]
\centering
\small
\setlength{\tabcolsep}{5.5pt}
\caption{Savvy experimental hyperparameters.}
\label{tab:savvy_hyperparams}
\resizebox{0.94\textwidth}{!}{%
\begin{tabular}{lll}
\toprule
Parameter & Value & Role \\
\midrule
Discovery stride $s$ & 3 frames & Scheduled image-level discovery opportunities \\
SAM grid size & 32$\times$32 & Default prompt grid for HMD \\
HMD survival threshold $\tau_{\mathrm{surv}}$ & 0.1 & Minimum retained area after proposal competition \\
Border delay & 0.05 & Delays mask discovery\\
Transient window $B$ & 30 frames & Deferred-admission duration \\
Duplicate IoU threshold & 0.5 & Transient--established agreement test \\
Duplicate agreement hits $\tau_{\mathrm{hit}}$ & 5 & Repeated hits needed for duplicate resolution \\
Visible frames $\tau_{\mathrm{vis}}$ & 3 & Minimum evidence for transient promotion \\
Appearance EMA $\alpha$ & 0.9 & Descriptor smoothing within an appearance episode \\
Appearance threshold $\tau_{\mathrm{app}}$ & 0.1 & Reappearance self-consistency gate \\
Competition IoU $\tau_{\mathrm{comp}}$ & 0.5 & Established-track arbitration (ScanNet/HM3D) \\
Competition margin $\delta$ & 0.1 & Minimum appearance-score gap for clear decision \\
Part-whole containment $\tau_{\mathrm{contain}}$ & 0.7 & Minimum containment for absorption \\
Part-whole area ratio & 1.5 & Minimum whole/part area ratio \\
Max active objects & 300 & Active SAM2 object budget \\
Memory-pruning interval & 10 frames & Period for stale SAM2 memory cleanup \\
Max SAM2 conditioning frames & 4 & Conditioning-frame budget in SAM2 attention \\
\bottomrule
\end{tabular}%
}
\end{table}

\begin{algorithm}[t]
\begin{algorithmpanel}
\caption{Restart-Aware Savvy Semi-Online Inference}
\label{alg:savvy_inference}
\begin{algorithmic}[1]
\Require Video $\{I_t\}_{t=1}^{T}$; video predictor $\Psi$; image segmenter $\Phi$; discovery stride $s$
\Ensure Object masks $\mathcal{Y}=\{Y_t\}_{t=1}^{T}$

\State $\mathcal{O}, \mathcal{B}, \mathcal{A}, \mathcal{Y} \gets \varnothing$
\State $\mathcal{C}_1 \gets \mathrm{HMDProposals}(I_1,\Phi)$
\State $\mathcal{O} \gets \mathrm{Register}(\Psi,\mathcal{C}_1)$

\For{$t=1$ to $T$}
    \State $\widehat{\mathcal{M}}_t \gets \Psi(I_t,\mathcal{O}\cup\mathcal{B})$
    \State Store established masks in $\mathcal{Y}_t$ and provisional masks in $\mathcal{B}$
    \State Update provisional visibility, maximum area, and appearance EMA
    \State $\mathcal{Y}_t \gets \mathrm{SelfConsistencyGate}(\mathcal{Y}_t,\mathcal{A})$
    \State $\mathcal{A} \gets \mathrm{UpdateAcceptedAppearance}(\mathcal{A},\mathcal{Y}_t)$
    \State $\mathcal{Y}_t \gets \mathrm{ArbitrateCompetition}(\mathcal{Y}_t,\mathcal{A})$
    \If{$t=1$}
        \State \textbf{continue} \Comment{bootstrap masks are already established}
    \EndIf
    \If{$\mathrm{MergeDuplicates}(t,\mathcal{O},\mathcal{B},\mathcal{Y})$}
        \State \textbf{continue} \Comment{resume propagation at the next frame}
    \EndIf
    \If{$\mathrm{ResolveCandidates}(t,\mathcal{O},\mathcal{B},\mathcal{Y},\mathrm{finalize}{=}\mathrm{false})$}
        \State \textbf{continue}
    \EndIf
    \State $\Psi \gets \mathrm{PrunePredictorMemory}(\Psi,t)$
    \If{$\mathrm{EnforceCapacity}(t,\mathcal{O},\mathcal{B},\Psi)$}
        \State \textbf{continue}
    \EndIf
    \If{$(t-1) \bmod s = 0$}
        \State $\mathcal{C}_t \gets \mathrm{HMDProposalsAndFilter}(I_t,\mathcal{O},\mathcal{B},\Phi)$
        \State $\mathcal{B} \gets \mathcal{B} \cup \mathrm{InsertTransient}(\Psi,\mathcal{C}_t)$
    \EndIf
\EndFor
\Repeat
    \State $r\gets\mathrm{ResolveCandidates}(T,\mathcal{O},\mathcal{B},\mathcal{Y},\mathrm{finalize}{=}\mathrm{true})$
\Until{no predictor restart is required ($r=\mathrm{false}$)}
\State \Return $\mathcal{Y}$
\end{algorithmic}
\end{algorithmpanel}
\end{algorithm}
\clearpage

\section{Details of OGA Evaluation}
\label{app:oga_details}

\subsection{Overview}
\label{app:oga_overview}

OGA is designed to evaluate open-world video segmentation when prediction granularity does not necessarily match the reference annotation. Conventional VPQ uses thresholded $1{:}1$ tube matches, while original STQ aggregates association over all intersecting prediction--reference tracks without thresholded matching. Both are nonetheless sensitive to the benchmark's object decomposition in different ways. A valid open-world prediction may decompose a reference object into several coherent parts, or may maintain a visually meaningful object hypothesis whose semantic granularity differs from the annotation.

OGA keeps the outer structure of classical video segmentation metrics, but replaces their association construction with granularity-agnostic support. Each reference instance anchors fidelity evaluation. Predictions are assigned through a one-sided support criterion, allowing multiple predictions to support the same reference while assigning each prediction to at most one reference within the scoring scope. OGA then constructs a temporal support chain for each reference, identifies discontinuities, and scores only the dominant coherent fragment. This ownership rule does not constrain the raw intersections used by IP or the framewise relations used by IC.

Unless otherwise stated, all OGA metrics are computed in the class-agnostic setting. Semantic labels are ignored, and evaluation is performed over instance identities. Void or ignore regions are excluded from valid support computation so that prediction spill into unannotated regions does not unfairly penalize a method when the reference annotation is incomplete.

The design principle of OGA is stated as follows
\begin{tealbox}
\textbf{Key principle.}
OGA permits geometrically supported part-level decompositions while penalizing temporal identity replacement through severing and dominant-fragment scoring. Two-sided diagnostics reveal support distributed across persistent parts; such subdivision need not reduce fidelity.
\end{tealbox}

\subsection{Granularity-Agnostic Matching}
\label{app:ga_matching}

Let $\mathcal{G}$ denote the set of reference instances and $\mathcal{P}$ denote the set of predicted instances with positive valid support over an evaluation sequence or window. We use $g\in\mathcal{G}$ for a reference tube and $p\in\mathcal{P}$ for a prediction tube. At frame $t$, their masks are denoted by $g_t$ and $p_t$, respectively. All areas and intersections are computed over the valid reference region; pixels labeled as void or ignore are excluded.

VPQ's thresholded one-to-one matches and STQ's all-overlap association do not explicitly construct reference-anchored, temporally coherent part support. OGA instead uses a one-sided support rule. For each prediction $p$, we measure how much of the prediction is explained by a reference instance $g$:
\begin{equation}
    S(p,g)
    =
    \frac{|p\cap g|}{|p|},
\end{equation}
where $|p\cap g|$ denotes the spatiotemporal intersection between prediction and reference, and $|p|$ denotes the prediction area after excluding void or ignore regions. This asymmetric score measures prediction support: a prediction is considered valid for $g$ if most of the prediction lies inside $g$, even if it covers only part of $g$.

For each prediction, we define its valid reference set as
\begin{equation}
    \mathcal{V}(p)
    =
    \left\{
    g\in\mathcal{G}
    \;\middle|\;
    S(p,g)\geq \tau_{\mathrm{ios}}
    \right\}.
\end{equation}
If $\mathcal{V}(p)=\varnothing$, the prediction remains unmatched. Otherwise, $p$ is assigned to the reference instance with the strongest support:
\begin{equation}
    a(p)
    =
    \arg\max_{g\in\mathcal{V}(p)} S(p,g).
\end{equation}

Tube eligibility uses IoS alone. At equal IoU and IoS thresholds, an IoU-or-IoS test is equivalent because $\operatorname{IoU}(p,g)\leq S(p,g)$; this equivalence need not hold when thresholds differ. The framewise diagnostic relations in Secs.~\ref{app:identity_persistence} and~\ref{app:identity_concentration} do use IoU-or-IoS.

This assignment is one-sided within its sequence or window. Each prediction can support at most one reference instance, but each reference instance may receive support from multiple predictions. The support pool of a reference instance is therefore
\begin{equation}
    \mathcal{P}(g)
    =
    \{p\in\mathcal{P}\mid a(p)=g\}.
\end{equation}
This gives each prediction one GA assignment within the evaluation scope, avoiding duplicate matching credit across references while allowing $n{:}1$ support. Foreground coverage and identity bleeding remain visible in other OGA terms.

For each frame $t$ where $g$ is present, define the active support set
\begin{equation}
    A_t(g)
    =
    \{p\in\mathcal{P}(g)\mid |p_t\cap g_t|>0\},
\end{equation}
and retain the ordered times $t_1<\cdots<t_{K_g}$ for which $A_{t_k}(g)\neq\varnothing$. Their support masses are
\begin{equation}
    M_{t_k}(g)
    =
    \sum_{p\in A_{t_k}(g)}
    |p_{t_k}\cap g_{t_k}|.
\end{equation}
The sequence of non-empty support states
\begin{equation}
    C(g)
    =
    \{(t_k,A_{t_k}(g),M_{t_k}(g))\}_{k=1}^{K_g}
\end{equation}
is the support chain of reference instance $g$. Consecutive chain states therefore mean consecutive \emph{non-empty} support states, not necessarily adjacent video frames. Frames without assigned support are omitted from identity-transition comparisons, but remain part of the full reference area used by GA-STQ and GA-VPQ; missing support is therefore never credited. The chain records both how much support $g$ receives and which prediction identities carry it.

\subsection{GA-Adapted STQ}
\label{app:ga_stq}

Original STQ combines association quality (AQ) with semantic segmentation quality (SQ). In our class-agnostic setting, OGA retains its geometric-mean structure while measuring foreground coverage as geometric quality (GQ):
\begin{equation}
    \mathrm{STQ}
    =
    \sqrt{\mathrm{AQ}\cdot \mathrm{GQ}},
\end{equation}
The association term uses reference-anchored GA support from Sec.~\ref{app:ga_matching} rather than original STQ's aggregation over all intersecting track pairs; GQ replaces semantic SQ because category labels are removed.

Throughout this section, let $\Omega_t$ denote the valid evaluation domain at frame $t$. All frame-level areas are computed over $\Omega_t$ unless otherwise stated. This keeps the metric definition independent of a particular dataset's void or ignore-region policy.

\paragraph{Dominant-fragment support.}
For each reference instance $g$, OGA first constructs its support chain
\begin{equation}
    C(g)=\{(t_k,A_{t_k}(g),M_{t_k}(g))\}_{k=1}^{K_g},
\end{equation}
where $A_t(g)$ is the set of assigned predictions actively supporting $g$ at frame $t$, and $M_t(g)$ is the total prediction--reference overlap mass at that frame. As described in Sec.~\ref{app:dominant_fragment}, this chain is partitioned into temporally coherent fragments, and only the dominant fragment $F^\star(g)$ is used for scoring.

Let $T_g^\star$ denote the set of frames covered by $F^\star(g)$, and let $\mathcal{P}^\star(g)$ denote the set of prediction identities active in this dominant fragment. OGA defines the true-positive support area of $g$ as
\begin{equation}
    \mathrm{TPA}_g^\star
    =
    \sum_{t\in T_g^\star}
    \sum_{p\in \mathcal{P}^\star(g)}
    |p_t\cap g_t|.
\end{equation}
The prediction area credited to the same dominant fragment is
\begin{equation}
    \mathrm{PredArea}_g^\star
    =
    \sum_{t\in T_g^\star}
    \sum_{p\in \mathcal{P}^\star(g)}
    |p_t|.
\end{equation}
Since areas are evaluated over $\Omega_t$, the prediction area above should be read as the area of $p_t$ inside the valid evaluation domain.

Let
\begin{equation}
    B_g
    =
    \sum_{t:g_t\neq\varnothing}
    |g_t|
\end{equation}
be the full reference area of $g$ over the evaluation sequence. The dominant-fragment union is
\begin{equation}
    \mathrm{Union}_g^\star
    =
    B_g+\mathrm{PredArea}_g^\star-\mathrm{TPA}_g^\star,
\end{equation}
and the corresponding virtual IoU is
\begin{equation}
    \mathrm{rawIoU}_g^\star
    =
    \frac{\mathrm{TPA}_g^\star}
    {\mathrm{Union}_g^\star+\varepsilon}.
\end{equation}

Importantly, the union uses the full reference area $B_g$, not only the reference area inside the dominant-fragment frames. This makes the relaxation conservative: a short but pure fragment cannot receive a high score unless it explains a substantial portion of the reference instance over its full temporal extent.

\paragraph{Soft temporal stability.}
Since GA support may involve multiple prediction identities, OGA includes a stability term to penalize changes in the active supporting identity set within the dominant fragment. Let $t_1^\star<\cdots<t_{K_g^\star}^\star$ be the ordered support states in $F^\star(g)$. For each selected state, let
\begin{equation}
    \widetilde{A}_{t_k^\star}(g)
    =
    \{p\in \mathcal{P}^\star(g)\mid |p_{t_k^\star}\cap g_{t_k^\star}|>0\}
\end{equation}
be the active support identity set restricted to the dominant fragment. For consecutive selected support states, OGA computes
\begin{equation}
    J_k(g)
    =
    \frac{
    |\widetilde{A}_{t_k^\star}(g)\cap \widetilde{A}_{t_{k+1}^\star}(g)|
    }{
    |\widetilde{A}_{t_k^\star}(g)\cup \widetilde{A}_{t_{k+1}^\star}(g)|+\varepsilon
    }.
\end{equation}
The soft stability of $g$ is then
\begin{equation}
    \mathrm{Stab}_g^\star
    =
    \frac{1}{K_g^\star-1}
    \sum_{k=1}^{K_g^\star-1}
    J_k(g),
\end{equation}
with $\mathrm{Stab}_g^\star=1$ when $K_g^\star=1$.

\paragraph{Area-efficiency penalty.}
Because GA allows multiple predictions to jointly support one reference instance, OGA also penalizes inefficient support that uses substantially more prediction area than the reference object requires:
\begin{equation}
    \mathrm{AreaPen}_g^\star
    =
    \left(
    \min\left(
    1,
    \frac{B_g}{\mathrm{PredArea}_g^\star+\varepsilon}
    \right)
    \right)^\gamma .
\end{equation}
This term remains near one when the credited support area is comparable to the reference area, and decreases when the assigned support uses excessive area.

\paragraph{GA association quality.}
The final dominant-fragment score for reference instance $g$ is
\begin{equation}
    \mathrm{Score}_g^\star
    =
    \mathrm{rawIoU}_g^\star
    \cdot
    \mathrm{Stab}_g^\star
    \cdot
    \mathrm{AreaPen}_g^\star .
\end{equation}
Its contribution to AQ is weighted by the amount of true-positive support:
\begin{equation}
    Q_g
    =
    \mathrm{TPA}_g^\star
    \cdot
    \mathrm{Score}_g^\star.
\end{equation}
OGA then averages the normalized contribution over reference instances:
\begin{equation}
\ogametric{
\begin{aligned}
    \mathrm{AQ}_{\mathrm{GA}}
    &=
    \frac{1}{|\mathcal{G}|}
    \sum_{g\in\mathcal{G}}
    \frac{Q_g}{B_g+\varepsilon}.
\end{aligned}
}
\end{equation}

\paragraph{Geometric quality.}
The geometric term measures foreground coverage over the chosen valid evaluation domain. Let $P_t(x)$ and $G_t(x)$ denote the prediction and reference label maps at frame $t$, with $0$ denoting background, void, or ignore according to the dataset convention. OGA computes
\begin{equation}
\ogametric{
\begin{aligned}
    \mathrm{GQ}_{\mathrm{GA}}
    &=
    \frac{
    \sum_t
    \left|
    \{x\in\Omega_t\mid P_t(x)\neq 0,\; G_t(x)\neq 0\}
    \right|
    }{
    \sum_t
    \left|
    \{x\in\Omega_t\mid G_t(x)\neq 0\}
    \right|
    +\varepsilon
    }.
\end{aligned}
}
\end{equation}
Thus, GQ measures how much of the valid reference foreground is covered by prediction foreground.

Finally,
\begin{equation}
\ogametric{
\begin{aligned}
    \mathrm{STQ}_{\mathrm{GA}}
    &=
    \sqrt{
    \mathrm{AQ}_{\mathrm{GA}}
    \cdot
    \mathrm{GQ}_{\mathrm{GA}}
    }.
\end{aligned}
}
\end{equation}

\paragraph{Void and ignore-region handling.}
The choice of $\Omega_t$ is dataset-dependent. In datasets with incomplete or partially reconstructed annotations, such as ScanNet, unlabeled regions may still contain visually valid objects or object parts. In this case, we use a void-tolerant domain in which prediction spill into GT-void pixels is excluded from prediction-area and overlap accounting. This prevents a faithful prediction from being penalized merely because the reference annotation is incomplete. In datasets with more complete reconstruction and annotation, such as HM3D, this tolerance can be reduced or removed by choosing $\Omega_t$ to include the full image or the full annotated evaluation region. This design separates the GA matching principle from dataset-specific annotation completeness.

The key difference from original STQ is not the outer aggregation form, but the support unit used to compute AQ. Original STQ aggregates all overlapping prediction--reference track-pair contributions, while OGA evaluates each reference through its dominant coherent support fragment. This preserves an emphasis on temporal association while making coherent part-level support explicit.

\subsection{GA-Adapted VPQ}
\label{app:ga_vpq}

VPQ evaluates panoptic video segmentation by computing PQ over sliding temporal windows and averaging over windows. Classical VPQ uses strict $1{:}1$ matching inside each window: a prediction tube and a reference tube form a true positive only if their IoU exceeds a fixed threshold. OGA keeps the same windowed PQ structure, but replaces the matched-pair IoU with the dominant-fragment score defined in Sec.~\ref{app:ga_stq}.

Let $W=[t,t+k)$ denote a temporal window of length $k$. Within each window, OGA recomputes prediction-to-reference assignment locally using the GA matching rule. This window-local assignment is important because a prediction may be valid support for a reference in one window but not in another, especially under long videos with disappearances, reappearances, or partial observations.

\paragraph{Window-level support construction.}
For each reference instance $g\in\mathcal{G}_W$ appearing in window $W$, OGA collects the set of predictions assigned to it within the window:
\begin{equation}
    \mathcal{P}_W(g)
    =
    \{p\in\mathcal{P}_W\mid a_W(p)=g\},
\end{equation}
where $a_W$ is the window-local GA assignment. As in sequence-level STQ, OGA constructs the support chain of $g$ inside the window, partitions it into sever-induced fragments, and selects the dominant fragment $F_W^\star(g)$ by maximum reference-overlap mass.

Using this dominant fragment, OGA computes a window-level score
\begin{equation}
    \mathrm{Score}_{g}^{W}
    =
    \mathrm{rawIoU}_{g}^{W,\star}
    \cdot
    \mathrm{Stab}_{g}^{W,\star}
    \cdot
    \mathrm{AreaPen}_{g}^{W,\star}.
\end{equation}
The terms are defined analogously to Sec.~\ref{app:ga_stq}, but all areas, supports, and fragments are restricted to the window $W$.

\paragraph{True positives and false negatives.}
A reference instance is counted as a true positive if its dominant-fragment score exceeds the VPQ matching threshold:
\begin{equation}
    g \in \mathrm{TP}_W
    \quad\Longleftrightarrow\quad
    \mathrm{Score}_{g}^{W} > \tau_{\mathrm{vpq}}.
\end{equation}
The number of true positives and false negatives are therefore
\begin{equation}
    \mathrm{TP}_W
    =
    \left|
    \{g\in\mathcal{G}_W\mid \mathrm{Score}_{g}^{W}>\tau_{\mathrm{vpq}}\}
    \right|,
\end{equation}
and
\begin{equation}
    \mathrm{FN}_W
    =
    |\mathcal{G}_W|-\mathrm{TP}_W.
\end{equation}
This mirrors classical VPQ, except that the matching score is no longer a rigid pairwise IoU between one prediction and one reference. Instead, each reference is evaluated through its dominant coherent GA support.

\paragraph{Ordinary and fragment-level false positives.}
Relaxing $1{:}1$ matching changes the meaning of false positives. In classical VPQ, any prediction not matched to a reference is counted as an FP. Under GA matching, however, a prediction may belong to a non-dominant support fragment of some reference. Counting all such predictions as ordinary false positives would double-penalize fragmented support: once through the dominant-fragment selection, and again as unmatched predictions. OGA therefore decomposes false positives into ordinary FP and fragment-level FP:
\begin{equation}
    \mathrm{FP}_W
    =
    \mathrm{FP}_{\mathrm{ord}}^W
    +
    \mathrm{FP}_{\mathrm{frag}}^W.
\end{equation}

The ordinary FP term counts prediction tubes in the window that are not credited by any matched dominant fragment, do not belong to any reference support fragment, and are not dominated by void or ignored regions:
\begin{equation}
    \mathrm{FP}_{\mathrm{ord}}^W
    =
    \left|
    \left\{
    p\in\mathcal{P}_W
    \;\middle|\;
    p\notin\mathcal{P}_{\mathrm{match}}^W,
    \;
    p\notin\mathcal{P}_{\mathrm{frag}}^W,
    \;
    p\notin\mathcal{P}_{\mathrm{void}}^W
    \right\}
    \right|.
\end{equation}
Here $\mathcal{P}_{\mathrm{match}}^W$ denotes predictions credited by matched dominant fragments, $\mathcal{P}_{\mathrm{frag}}^W$ denotes predictions appearing in any sever-induced support fragment for any reference in the window, and $\mathcal{P}_{\mathrm{void}}^W$ denotes predictions whose support lies mostly in void or ignored regions under the chosen evaluation-domain policy.

The fragment-level FP term penalizes temporal fragmentation directly. Let $n_g^W$ be the number of sever-induced fragments in the support chain of reference instance $g$ inside window $W$. OGA defines
\begin{equation}
    \mathrm{FP}_{\mathrm{frag}}^W
    =
    \sum_{g\in\mathcal{G}_W}
    \max(n_g^W-1,0).
\end{equation}
Thus, a reference supported by one coherent fragment incurs no fragment FP, while each additional sever-induced fragment contributes an FP unit. This preserves VPQ's penalty for identity fragmentation without forcing strict $1{:}1$ matching.

\paragraph{Window PQ and VPQ.}
The window-level GA-PQ is then
\begin{equation}
\ogametric{
\begin{aligned}
    \mathrm{PQ}_{\mathrm{GA}}^W
    &=
    \frac{
    \sum_{g\in\mathcal{G}_W:\mathrm{Score}_{g}^{W}>\tau_{\mathrm{vpq}}}
    \mathrm{Score}_{g}^{W}
    }{
    \mathrm{TP}_W
    +
    \frac{1}{2}\mathrm{FP}_W
    +
    \frac{1}{2}\mathrm{FN}_W
    +\varepsilon
    }.
\end{aligned}
}
\end{equation}
For a window length $k$, GA-VPQ is the average over all sliding windows of that length:
\begin{equation}
\ogametric{
\begin{aligned}
    \mathrm{VPQ}_{k}^{\mathrm{GA}}
    &=
    \frac{1}{|\mathcal{W}_k|}
    \sum_{W\in\mathcal{W}_k}
    \mathrm{PQ}_{\mathrm{GA}}^W.
\end{aligned}
}
\end{equation}
We also report $\mathrm{VPQ}_{\infty}$ by taking the full evaluated sequence as a single window.

Compared with classical VPQ, the denominator retains the same TP/FP/FN structure, but the support unit changes. True positives are reference instances explained by dominant coherent support fragments; ordinary false positives remain unsupported predictions; and fragment-level false positives explicitly penalize excess severed support. This makes VPQ robust to granularity mismatch while preserving sensitivity to identity breakage and over-fragmentation.

\subsection{Worked OGA Example}
\label{app:oga_worked_example}

Consider one reference object of area $A$ in each of four frames. Predictions are non-overlapping label maps with exact foreground coverage, no other references or predictions, and no void-only IDs. Use $\tau_{\mathrm{ios}}=\tau_{\mathrm{iou}}=\tau_{\mathrm{sever}}=\tau_{\mathrm{vpq}}=0.5$, $\gamma=1$, dominant-fragment scoring, and soft support-set stability. All masks match framewise, so $D=1$; GQ, IP$_P$, and IC$_P$ are also one in every case.

\paragraph{One persistent identity.}
An exact prediction with the same ID in all four frames has one support fragment, with $B_g=C_g=D_g=4A$, stability one, and no excess-area penalty. Thus $q_g=1$, GA-AQ $=1$, GA-STQ $=1$, and GA-VPQ$_\infty=1$. One prediction carries all reference overlap, so IP$_G=\mathrm{IC}_G=1$.

\paragraph{Two persistent equal halves.}
Two disjoint predictions of area $A/2$ cover the reference in every frame and retain their respective IDs. The active set is always $\{a,b\}$, giving one fragment with the same $B_g,C_g,D_g$ and stability as above. All fidelity scores remain one. Each ID carries half the reference volume, so IP$_G=1/2$; two IDs ever support the reference, so IC$_G=1/2$. Both prediction-side diagnostics remain one because each half supports only this reference. OGA fidelity therefore permits this persistent subdivision, while its diagnostics report distributed identity support.

\paragraph{Temporal replacement: $a,a,a,b$.}
Each frame has an exact whole-object mask, but the ID changes at the fourth frame. The transition from $a$ to $b$ has $\rho=0$, splitting the support chain into masses $3A$ and $A$. The dominant fragment retains the first three frames: $B_g=4A$, $C_g=D_g=3A$, with stability and area penalty both one. Hence
\begin{equation}
q_g=\frac{3A}{4A+3A-3A}=\frac34,\qquad
\mathrm{AQ}_{\mathrm{GA}}=\frac{3A}{4A}\frac34=\frac9{16},\qquad
\mathrm{STQ}_{\mathrm{GA}}=\sqrt{\frac9{16}}=\frac34.
\end{equation}
The dominant fragment passes the strict $q_g>0.5$ VPQ gate. The extra fragment adds one FP, with one TP and no FN, yielding
\begin{equation}
\mathrm{VPQ}_{\infty}^{\mathrm{GA}}=\frac{3/4}{1+\tfrac12\cdot1}=\frac12.
\end{equation}
The dominant prediction covers three quarters of the reference volume, so IP$_G=3/4$, while the two ever-supporting IDs give IC$_G=1/2$. Prediction-side purity and concentration stay at one despite the identity replacement.

These examples distinguish persistent spatial subdivision from temporal identity replacement. The empirical spatial-fragmentation test (Sec.~\ref{app:spatial_fragmentation}) extends this analysis to child tracks whose assignment and temporal support may change.

\subsection{Dominant Fragment Selection}
\label{app:dominant_fragment}

Under GA matching, a reference instance may be supported by multiple prediction identities over time. This is necessary for granularity-agnostic evaluation, but it introduces a new question: which portion of the support should be credited as temporally coherent? If all assigned predictions were credited indiscriminately, fragmented or flickering predictions could be over-rewarded. OGA therefore converts the support pool of each reference instance into a support chain, severs the chain at identity discontinuities, and scores only the dominant coherent fragment.

For a reference instance $g$, let
\begin{equation}
    C(g)=\{(t_k,A_{t_k}(g),M_{t_k}(g))\}_{k=1}^{K_g}
\end{equation}
denote its non-empty support chain, where $A_{t_k}(g)$ is the set of prediction identities supporting $g$ at support state $t_k$, and
\begin{equation}
    M_{t_k}(g)
    =
    \sum_{p\in A_{t_k}(g)}
    |p_{t_k}\cap g_{t_k}|
\end{equation}
is the total support mass at that state. The chain includes only frames where $g$ is present and receives non-empty assigned support.

\paragraph{Persistent-support ratio.}
To determine whether support remains temporally coherent across adjacent support states, OGA measures how much support is carried by the same prediction identities before and after the transition. For two consecutive states $t_k$ and $t_{k+1}$, define
\begin{equation}
    \rho_g^{(k)}
    =
    \frac{
    \sum\limits_{p\in A_{t_k}(g)\cap A_{t_{k+1}}(g)}
    \left(
    |p_{t_k}\cap g_{t_k}|
    +
    |p_{t_{k+1}}\cap g_{t_{k+1}}|
    \right)
    }{
    \sum\limits_{p\in A_{t_k}(g)\cup A_{t_{k+1}}(g)}
    \left(
    |p_{t_k}\cap g_{t_k}|
    +
    |p_{t_{k+1}}\cap g_{t_{k+1}}|
    \right)
    +\varepsilon
    }.
\end{equation}
The numerator measures support mass carried by persistent identities appearing on both sides of the transition. The denominator measures the total support mass active across the transition. Thus, $\rho_g^{(k)}$ is high when the same prediction identities continue to support $g$, and low when support is largely replaced by different identities.

A sever is declared when the persistent-support ratio is at or below a threshold,
\begin{equation}
    \rho_g^{(k)} \leq \tau_{\mathrm{sever}}.
\end{equation}
Sever points partition the support chain into temporally coherent fragments:
\begin{equation}
    C(g)
    \rightarrow
    \{F_g^{(1)},F_g^{(2)},\ldots,F_g^{(n_g)}\}.
\end{equation}

\paragraph{Fragment mass.}
Each fragment is scored by the amount of reference-overlap mass it carries. For a fragment $F_g^{(r)}$, define
\begin{equation}
    W_g^{(r)}
    =
    \sum_{(t,A_t(g),M_t(g))\in F_g^{(r)}}
    M_t(g).
\end{equation}
The dominant fragment is selected as
\begin{equation}
    F^\star(g)
    =
    \arg\max_{F_g^{(r)}} W_g^{(r)}.
\end{equation}
In implementation, ties are broken by fragment length, favoring the longer support fragment when two fragments carry the same overlap mass.

The selected fragment determines both the credited prediction identities and the credited temporal span:
\begin{equation}
    T_g^\star
    =
    \{t\mid (t,A_t(g),M_t(g))\in F^\star(g)\},
\end{equation}
and
\begin{equation}
    \mathcal{P}^\star(g)
    =
    \bigcup_{(t,A_t(g),M_t(g))\in F^\star(g)}
    A_t(g).
\end{equation}
Only $T_g^\star$ and $\mathcal{P}^\star(g)$ are used to compute the dominant-fragment score in GA-STQ and GA-VPQ.

\paragraph{Conservative effect.}
Dominant fragment selection is intentionally conservative. A reference instance may receive support from many predictions across the evaluated sequence, but only the most coherent high-mass fragment is credited. Support outside the dominant fragment is not used to increase the GT score. In VPQ, extra sever-induced fragments are additionally penalized through fragment-level false positives. Therefore, GA matching relaxes the granularity constraint, but dominant-fragment selection prevents the relaxation from becoming a free aggregation of disconnected prediction fragments.

\subsection{Dominant Fragment as a Practical Relaxation of Coherent Support Selection}
\label{app:mwcs}

We formalize the support selection problem underlying the dominant-fragment rule. Dominant-fragment selection provides a tractable, conservative surrogate for selecting the strongest temporally coherent support core among all predictions assigned to a reference instance.

\paragraph{Setup.}
Fix a reference instance $g$ over a video. Let
\begin{equation}
    \mathcal{P}(g)=\{p_1,\ldots,p_N\}
\end{equation}
denote the set of prediction identities assigned to $g$ under GA matching. For each prediction $p\in\mathcal{P}(g)$ and frame $t$, define the framewise support mass
\begin{equation}
    a_{p,t}
    =
    |p_t\cap g_t|.
\end{equation}
For a selected subset $S\subseteq\mathcal{P}(g)$, define its active support at frame $t$ as
\begin{equation}
    A_t(S)
    =
    \{p\in S\mid a_{p,t}>0\},
\end{equation}
and its total support mass as
\begin{equation}
    m_t(S)
    =
    \sum_{p\in S} a_{p,t}.
\end{equation}

\paragraph{Temporal coherence.}
The purpose of dominant-fragment selection is to prevent support from being optimistically pooled across temporally disconnected identity fragments. For a selected subset $S$ and adjacent frames $t,t+1$, define the persistent-support ratio
\begin{equation}
    \rho_t(S)
    =
    \frac{
    \sum\limits_{p\in A_t(S)\cap A_{t+1}(S)}
    \left(a_{p,t}+a_{p,t+1}\right)
    }{
    \sum\limits_{p\in A_t(S)\cup A_{t+1}(S)}
    \left(a_{p,t}+a_{p,t+1}\right)
    +\varepsilon
    }.
\end{equation}
A transition is coherent if
\begin{equation}
    \rho_t(S) > \tau_{\mathrm{sever}}.
\end{equation}
Thus, a support subset is sever-free over an interval $[u,v]$ if every adjacent transition within the interval satisfies the coherence condition whenever support is present. This ideal formulation requires support at every video frame in the interval; the implemented surrogate below instead compares consecutive non-empty support states, retaining unsupported reference area in its scoring denominator.

\paragraph{Ideal coherent support-core search.}
An ideal evaluator would select the subset of predictions and temporal interval that maximizes coherent support for $g$:
\begin{equation}
\label{eq:ideal_core_search}
\begin{aligned}
\max_{S\subseteq\mathcal{P}(g),\;[u,v]}
&\quad
\sum_{t=u}^{v} m_t(S) \\
\mathrm{s.t.}
&\quad
m_t(S)>0,
\qquad \forall t\in[u,v],\\
&\quad
\rho_t(S)>\tau_{\mathrm{sever}},
\qquad \forall t=u,\ldots,v-1 .
\end{aligned}
\end{equation}
This formulation searches for the strongest temporally coherent explanation of the reference object. A more score-faithful variant could replace the numerator objective with a virtual-IoU or area-efficiency objective, but Eq.~\eqref{eq:ideal_core_search} already captures the core difficulty: the evaluator must jointly choose a prediction subset and a coherent temporal support interval.

\paragraph{Graph view and search space.}
A graph view organizes the candidate support units in Eq.~\eqref{eq:ideal_core_search}. Its nodes represent prediction-time support units,
\begin{equation}
    v_{p,t}
    \quad\Longleftrightarrow\quad
    a_{p,t}>0,
\end{equation}
with node weight $a_{p,t}$. Same-identity links connect successive support units of a prediction. Coherence is evaluated jointly for the selected subset and interval through $\rho_t(S)$, whose numerator and denominator both depend on that subset.

This view is conceptually related to maximum-weight connected subgraph selection~\cite{mwcs}, which seeks a connected node subset with maximal total weight. Here, exhaustive search over $N$ prediction identities and $T$ frames would consider $(2^N-1)T(T+1)/2$ subset--interval pairs before feasibility filtering. OGA avoids this enumeration for each reference instance and VPQ window through the dominant-fragment surrogate below.

\paragraph{Dominant-fragment surrogate.}
OGA therefore adopts a deterministic surrogate. Instead of searching over all subsets $S\subseteq\mathcal{P}(g)$, it first builds the support chain induced by all assigned predictions:
\begin{equation}
    C(g)=\{(t,A_t(g),M_t(g))\}_{t:A_t(g)\neq\varnothing}.
\end{equation}
It then computes the persistent-support ratio between adjacent support states, splits the chain at sever points, and obtains a set of sever-delimited fragments
\begin{equation}
    \mathcal{F}(g)
    =
    \{F_g^{(1)},\ldots,F_g^{(n_g)}\}.
\end{equation}
Each fragment is scored by its total reference-overlap mass:
\begin{equation}
    W_g^{(r)}
    =
    \sum_{(t,A_t(g),M_t(g))\in F_g^{(r)}}
    M_t(g).
\end{equation}
The credited support is the dominant fragment
\begin{equation}
    F^\star(g)
    =
    \arg\max_{F_g^{(r)}\in\mathcal{F}(g)}
    W_g^{(r)}.
\end{equation}

\paragraph{Why this is conservative.}
The dominant-fragment rule relaxes spatial granularity but tightens temporal credit assignment. Multiple predictions may jointly support a reference instance, but disconnected support fragments are not freely pooled. Once sever points appear, only the highest-mass coherent fragment contributes to the GT score; support outside this fragment is not credited. In GA-VPQ, additional sever-induced fragments are further penalized through fragment-level false positives. Together, dominant-fragment scoring and fragment-level false positives reward coherent reference support and penalize temporal fragmentation and identity replacement.

\paragraph{Practical motivation.}
The ideal coherent-core formulation jointly searches prediction subsets and temporal intervals. Dominant-fragment selection replaces this search with deterministic chain construction, severing, and maximum-mass selection, yielding a reproducible evaluation rule that credits the strongest sever-delimited support fragment.

\subsection{Structural Diagnostics: Persistence and Concentration}
\label{app:structural_diagnostics}

GA matching relaxes the rigid $1{:}1$ constraint by allowing multiple predictions to support the same reference instance. This makes the primary metrics fairer under granularity mismatch, but it also exposes structural failure modes that strict matching would otherwise collapse into a single score. In particular, once $n{:}1$ support is allowed, we need to distinguish between support that is persistent and compact, and support that is obtained through many fragmented or redundant identities.

OGA therefore reports two complementary structural diagnostics: identity persistence (IP) and identity concentration (IC). IP is a \emph{mass-based} diagnostic: it asks whether most overlap mass is carried by one dominant identity. IC is a \emph{graph-based} diagnostic: it asks how many distinct identities ever participate in valid support relations. These two views are related but not interchangeable.

\paragraph{Valid-domain areas and ID averaging.}
All diagnostic overlap and area accounting excludes void/ignore pixels. Diagnostic prediction IDs are collected before valid-domain clipping: an ID occurring only in void contributes a zero entry to the prediction-side IP and IC means. In the diagnostic formulas below, $\mathcal{P}$ denotes this collected ID set, with zero assigned when valid area or support is absent. GA fidelity instead forms tubes only from IDs with positive valid support. Thus the metrics share valid-domain pixel accounting, but need not share the same prediction-ID averaging universe. For example, adding a void-only ID to an otherwise perfect one-ID prediction leaves fidelity unchanged but reduces IP$_P$ and IC$_P$ from one to $1/2$.

Both diagnostics are defined on two axes. The prediction axis asks whether each predicted identity remains attached to a compact set of reference objects. The reference axis asks whether each reference object is explained by a compact set of prediction identities. Low prediction-axis scores can reflect identity bleeding across references or unmatched predictions. Low reference-axis scores can reflect fragmentation across prediction IDs or missing support.

We therefore interpret IP and IC jointly. High IP but low IC can occur when one identity dominates the overlap mass but many auxiliary fragments also appear. High IC but low IP can occur when only a few identities are involved, but no single identity carries the reference consistently. Together, the two diagnostics provide a structural view of how a method organizes prediction identities beyond aggregate overlap scores.

\subsection{Identity Persistence}
\label{app:identity_persistence}

Identity persistence (IP) measures whether the support mass of an object is dominated by a single identity over time. While GA-STQ and GA-VPQ provide primary fidelity scores, they do not by themselves reveal whether a method obtains support through stable identities or through many short-lived fragments. IP provides a mass-based diagnostic for this question.

Let $M_{p,g}$ denote the sequence-level overlap mass between prediction $p$ and reference instance $g$:
\begin{equation}
    M_{p,g}
    =
    \sum_t |p_t\cap g_t|.
\end{equation}
Let
\begin{equation}
    A_p=\sum_t |p_t|,
    \qquad
    B_g=\sum_t |g_t|
\end{equation}
denote the total prediction and reference areas over the evaluation sequence, computed over the chosen evaluation domain $\Omega_t$.

IP includes a framewise discovery factor. Let $\mathcal{P}_t$ contain prediction IDs with positive valid area after GT-void clipping, and let $\mathcal{M}_t$ be the distinct prediction IDs that match at least one reference in frame $t$:
\begin{equation}
\begin{aligned}
\operatorname{match}_t(p,g)&\iff
\mathrm{IoU}(p_t,g_t)\ge\tau_{\mathrm{iou}}\ \text{or}\
\frac{|p_t\cap g_t|}{|p_t|}\ge\tau_{\mathrm{ios}},\\
\mathcal{M}_t&=\{p\in\mathcal{P}_t:\exists g\in\mathcal{G}_t,\ \operatorname{match}_t(p,g)\},\\
D_t&=\min\!\left(\frac{|\mathcal{M}_t|}{\max(|\mathcal{G}_t|,1)},1\right),
\qquad D=\frac{1}{T}\sum_{t=1}^{T}D_t.
\end{aligned}
\end{equation}
The numerator counts \emph{matched prediction IDs}, not distinct GT objects; the cap keeps $D_t\leq1$. The threshold test controls $D$, but $M_{p,g}$, $A_p$, and $B_g$ accumulate all valid pixel overlaps and areas without filtering by IoS.

\paragraph{Prediction-axis persistence.}
Prediction-axis persistence measures whether each prediction identity primarily supports one reference instance:
\begin{equation}
\ogametric{
\begin{aligned}
    \mathrm{IP}_P(p)
    &=
    \frac{\max_{g\in\mathcal{G}} M_{p,g}}
    {A_p+\varepsilon}.
\end{aligned}
}
\end{equation}
A high $\mathrm{IP}_P(p)$ means that prediction $p$ is mostly attached to a single reference object. A low value indicates identity bleeding: the same prediction identity spreads its mass across multiple reference objects.

The sequence-level prediction-axis persistence is
\begin{equation}
\ogametric{
\begin{aligned}
    \mathrm{IP}_P
    &=
    D\frac{1}{|\mathcal{P}|}
    \sum_{p\in\mathcal{P}}
    \mathrm{IP}_P(p).
\end{aligned}
}
\end{equation}

\paragraph{Reference-axis persistence.}
Reference-axis persistence measures whether each reference instance is mainly carried by one prediction identity:
\begin{equation}
\ogametric{
\begin{aligned}
    \mathrm{IP}_G(g)
    &=
    \frac{\max_{p\in\mathcal{P}} M_{p,g}}
    {B_g+\varepsilon}.
\end{aligned}
}
\end{equation}
A high $\mathrm{IP}_G(g)$ means that the reference object is consistently explained by one dominant prediction. A low value indicates identity fragmentation: the reference object is split across multiple prediction IDs over time.

The sequence-level reference-axis persistence is
\begin{equation}
\ogametric{
\begin{aligned}
    \mathrm{IP}_G
    &=
    D\frac{1}{|\mathcal{G}|}
    \sum_{g\in\mathcal{G}}
    \mathrm{IP}_G(g).
\end{aligned}
}
\end{equation}

The two axes should be interpreted together. $\mathrm{IP}_P$ captures whether prediction identities remain pure with respect to reference objects, while $\mathrm{IP}_G$ captures whether reference objects are carried by persistent prediction identities. Both are scaled by the same $D$: changing $\tau_{\mathrm{ios}}$ can affect IP through framewise discovery, but not through the underlying global-overlap ratios. If $D$ is already saturated at $1$, changing that threshold leaves IP unchanged. A method can have high $\mathrm{IP}_P$ but low $\mathrm{IP}_G$ when it produces many pure but short-lived fragments; conversely, low $\mathrm{IP}_P$ indicates that prediction identities bleed across multiple reference objects.

\subsection{Identity Concentration}
\label{app:identity_concentration}

Identity concentration (IC) measures whether the prediction--reference support structure is compact. Unlike IP, which is based on overlap mass, IC is based on the number of distinct identities that ever participate in valid support relations. It therefore captures structural fragmentation and bleeding even when the overlap mass is dominated by one identity.

IC uses a frame-induced \emph{ever-support} graph, distinct from the tube- or window-level assignment $a(p)$. For any frame with $|p_t|>0$, define
\begin{equation}
    S_t(p,g)=\frac{|p_t\cap g_t|}{|p_t|}.
\end{equation}
Let $p\sim g$ denote a pre-assignment validity edge:
\begin{equation}
    p\sim g
    \quad\Longleftrightarrow\quad
    \exists t:\
    \operatorname{IoU}(p_t,g_t)\geq\tau_{\mathrm{iou}}
    \ \text{or}\
    S_t(p,g)\geq\tau_{\mathrm{ios}}.
\end{equation}
We define the set of reference instances ever supported by prediction $p$ as
\begin{equation}
    \mathcal{G}_{\mathrm{ever}}(p)
    =
    \{g\in\mathcal{G}\mid p\sim g\},
\end{equation}
and the set of predictions ever supporting reference instance $g$ as
\begin{equation}
    \mathcal{P}_{\mathrm{ever}}(g)
    =
    \{p\in\mathcal{P}\mid p\sim g\}.
\end{equation}

\paragraph{Prediction-axis concentration.}
Prediction-axis concentration measures whether a predicted identity remains structurally attached to a small number of reference objects:
\begin{equation}
\ogametric{
\begin{aligned}
    \mathrm{IC}_P(p)
    &=
    \begin{cases}
    \frac{1}{|\mathcal{G}_{\mathrm{ever}}(p)|}, 
    & |\mathcal{G}_{\mathrm{ever}}(p)|>0,\\[3pt]
    0, & \text{otherwise}.
    \end{cases}
\end{aligned}
}
\end{equation}
A high $\mathrm{IC}_P(p)$ means that prediction $p$ is structurally concentrated on one reference object. A low value indicates identity bleeding, where the same prediction identity becomes valid support for multiple reference objects over time.

The dataset-level prediction-axis concentration is
\begin{equation}
\ogametric{
\begin{aligned}
    \mathrm{IC}_P
    &=
    \frac{1}{|\mathcal{P}|}
    \sum_{p\in\mathcal{P}}
    \mathrm{IC}_P(p).
\end{aligned}
}
\end{equation}

\paragraph{Reference-axis concentration.}
Reference-axis concentration measures whether a reference object is structurally supported by a compact set of prediction identities:
\begin{equation}
\ogametric{
\begin{aligned}
    \mathrm{IC}_G(g)
    &=
    \begin{cases}
    \frac{1}{|\mathcal{P}_{\mathrm{ever}}(g)|}, 
    & |\mathcal{P}_{\mathrm{ever}}(g)|>0,\\[3pt]
    0, & \text{otherwise}.
    \end{cases}
\end{aligned}
}
\end{equation}
A high $\mathrm{IC}_G(g)$ means that the reference object is supported by only one or a few prediction identities. A low value indicates identity fragmentation: many prediction IDs become valid support for the same reference object.

The dataset-level reference-axis concentration is
\begin{equation}
\ogametric{
\begin{aligned}
    \mathrm{IC}_G
    &=
    \frac{1}{|\mathcal{G}|}
    \sum_{g\in\mathcal{G}}
    \mathrm{IC}_G(g).
\end{aligned}
}
\end{equation}

\paragraph{Relation to IP.}
IC and IP are complementary. IP measures whether most support mass is dominated by one identity, while IC measures how many identities participate in the support graph at all. For example, a method may achieve high $\mathrm{IP}_G$ if one prediction carries most of a reference object's overlap mass, but still have low $\mathrm{IC}_G$ if many short-lived auxiliary predictions also touch the same reference. Conversely, a method may have reasonable concentration but low persistence if a small number of identities alternate support without one identity dominating the overlap mass.

Thus, IC should not be interpreted as an overlap score. It is a structural compactness diagnostic for the prediction--reference support graph. Together with IP, it reveals whether a method maintains a clean object identity structure or obtains its primary scores through fragmented, redundant, or bleeding identities.

\subsection{Prediction--Reference Support Matrix}
\label{app:support_matrix}

The prediction--reference support matrix provides a direct visualization of the structural relations summarized by IP and IC. While IP and IC reduce identity structure to scalar diagnostics, the support matrix exposes the full many-to-many pattern between prediction identities and reference instances. It is therefore useful for interpreting whether a method produces compact object identities, fragmented support, or identity bleeding.

Let $\mathcal{P}=\{p_1,\ldots,p_N\}$ denote prediction identities and $\mathcal{G}=\{g_1,\ldots,g_M\}$ denote reference instances. We define a weighted support matrix
\begin{equation}
    W \in \mathbb{R}^{N\times M},
    \qquad
    W_{i,j}
    =
    \frac{M_{p_i,g_j}}{A_{p_i}+\varepsilon},
\end{equation}
where
\begin{equation}
    M_{p_i,g_j}
    =
    \sum_t |p_{i,t}\cap g_{j,t}|
\end{equation}
is the sequence-level overlap mass and $A_{p_i}=\sum_t |p_{i,t}|$ is the total area of prediction $p_i$. Thus, $W_{i,j}$ measures the fraction of prediction $p_i$ explained by reference object $g_j$. A binary support matrix can also be obtained by thresholding the relaxed matching relation:
\begin{equation}
    E_{i,j}
    =
    \mathbb{1}[p_i\sim g_j],
\end{equation}
where $p_i\sim g_j$ denotes that prediction $p_i$ and reference $g_j$ form a valid relaxed match at least once.

The rows and columns of this matrix correspond to the two diagnostic axes. Row dispersion indicates prediction-side bleeding: if one prediction row has nonzero support over many references, the same predicted identity is being shared across multiple objects. Column dispersion indicates reference-side fragmentation: if one reference column receives support from many prediction rows, the reference object is split across many predicted identities. IC summarizes this binary structure:
\begin{equation}
    \mathrm{IC}_P(p_i)
    =
    \frac{1}{\sum_j E_{i,j}},
    \qquad
    \mathrm{IC}_G(g_j)
    =
    \frac{1}{\sum_i E_{i,j}},
\end{equation}
with score $0$ when the corresponding row or column has no valid support.

This matrix is analogous in spirit to a detection confusion matrix, but it is defined over instance support rather than semantic classes. A clean prediction structure should be sparse and concentrated: each prediction row should mainly support one reference, and each reference column should be explained by a compact set of prediction identities. Different failure modes appear as different matrix patterns. Over-fragmentation creates tall columns with many active prediction rows. Identity bleeding creates wide rows with support across many reference columns. Missed objects appear as empty or weak reference columns, while spurious predictions appear as empty or weak prediction rows.

This view is especially useful under GA evaluation. Since GA intentionally permits multiple predictions to support one reference object, a high primary score alone does not reveal whether the support came from a compact object set or from many auxiliary fragments. The support matrix makes this structure explicit. IP measures whether the mass in each row or column is dominated by one entry, while IC measures how many entries are active at all. Together, the scalar diagnostics and the matrix visualization provide a more complete picture of identity organization under granularity-agnostic matching.

\paragraph{Visualization protocol.}
The definition above specifies the support matrix itself, but the readability of the visualization also depends on the ordering of prediction and reference identities. 
In Appendix~\ref{app:support_matrix_visualization}, we describe the plotting protocol used in our figures, including the dominant-reference sorting rule, the two normalizations by $A_p$ and $B_g$, and representative observations for Savvy, DEVA+SAM, and EntitySAM.
\subsection{Additional Temporal Diagnostics}
\label{app:temporal_diagnostics}

In addition to IP and IC, OGA includes optional temporal diagnostics that characterize the structure of support over time. These diagnostics are not used as primary fidelity scores. Instead, they provide additional views of whether a reference object is supported by a temporally connected and recurring set of prediction identities. We report two such diagnostics: connected support coverage, denoted as Cluster, and compact support-pattern coverage, denoted as Patt$_k$.

\paragraph{Connected support coverage.}
For each reference instance $g$, consider the prediction identities that ever provide valid support to $g$. We build a support graph whose nodes are these prediction identities. Two prediction identities are connected if they co-occur as active support for $g$ in at least one frame, or if they appear in adjacent support states, corresponding to an end-to-end handoff. The connected components of this graph represent temporally linked support groups.

Let $\mathcal{C}_g=\{C_g^{(1)},\ldots,C_g^{(r_g)}\}$ denote the connected components of the support graph for $g$. Each component covers a subset of frames in which at least one prediction identity in that component supports $g$. We define the coverage of a component as
\begin{equation}
    \operatorname{cov}(C_g^{(r)})
    =
    \left|
    \bigcup_{p\in C_g^{(r)}} T_{p,g}
    \right|,
\end{equation}
where $T_{p,g}$ is the set of frames where prediction $p$ supports reference instance $g$. The Cluster score for $g$ is the frame coverage of the largest connected support component:
\begin{equation}
    \operatorname{Cluster}(g)
    =
    \frac{
    \max_{C\in\mathcal{C}_g}
    \operatorname{cov}(C)
    }{
    |T_g|+\varepsilon
    },
\end{equation}
where $T_g$ is the set of frames in which $g$ is present. The dataset-level score is
\begin{equation}
    \operatorname{Cluster}
    =
    \frac{1}{|\mathcal{G}|}
    \sum_{g\in\mathcal{G}}
    \operatorname{Cluster}(g).
\end{equation}

A high Cluster score indicates that most of the reference object's temporal extent is covered by one connected support component. A low score suggests that support for the same reference object is broken into temporally disconnected groups.

\paragraph{Compact support-pattern coverage.}
Cluster measures connectedness, but it does not directly measure whether the support can be explained by a small recurring set of prediction identities. Patt$_k$ addresses this by asking how much of a reference object's temporal extent can be covered by a compact pattern of $k$ prediction identities.

For a reference instance $g$, let $T_{p,g}$ again denote the set of frames where prediction $p$ supports $g$. For a fixed pattern size $k$, we define
\begin{equation}
    \operatorname{Patt}_k(g)
    =
    \max_{\{p_1,\ldots,p_k\}\subseteq \mathcal{P}_{\mathrm{ever}}(g)}
    \frac{
    \left|
    T_{p_1,g}
    \cap
    \cdots
    \cap
    T_{p_k,g}
    \right|
    }{
    |T_g|+\varepsilon
    }.
\end{equation}
This measures the best recurring $k$-identity support pattern for $g$. For example, Patt$_1$ is high when a single prediction identity repeatedly supports the reference object, while Patt$_2$ can capture a stable two-part decomposition that co-occurs over time.

The dataset-level Patt$_k$ score is
\begin{equation}
    \operatorname{Patt}_k
    =
    \frac{1}{|\mathcal{G}|}
    \sum_{g\in\mathcal{G}}
    \operatorname{Patt}_k(g).
\end{equation}
\paragraph{Complementary temporal axes.}
Cluster and Patt$_k$ capture complementary notions of temporal structure. 
Cluster measures the span of the best connected support chain: identities may change over time, but they remain linked through co-occurrence or adjacent handoff. 
Patt$_k$ measures the span of the best co-existing support core: a fixed set of $k$ prediction identities must support the reference object on the same frames. 
Thus, Cluster is tolerant to temporally connected handoffs, while Patt$_k$ is stricter and emphasizes recurring simultaneous support.

\begin{table*}[t]
\centering
\small
\setlength{\tabcolsep}{6pt}
\caption{
\textbf{Standard OGA evaluation settings.}
These settings are shared across datasets and methods. Dataset-specific ignore labels and sampling policies appear in Sec.~\ref{app:dataset_protocol}.
}
\label{tab:oga_hyperparams}
\resizebox{\linewidth}{!}{
\begin{tabular}{lll}
\toprule
\textbf{Component} & \textbf{Parameter / Setting} & \textbf{Value} \\
\midrule

\multirow{5}{*}{GA matching}
& Prediction-support threshold $\tau_{\mathrm{ios}}$ & 0.5 \\
& Prediction ownership & Each prediction assigned to at most one GT \\
& Reference-side support & Multiple predictions may support one GT \\
& Assignment score & $\mathrm{IoS}(p\!\to\!g)=|p\cap g|/|p|$ \\
& Assignment scope for STQ & Sequence-level \\
\midrule

\multirow{5}{*}{Dominant fragment}
& Sever threshold $\tau_{\mathrm{sever}}$ & 0.5 \\
& Sever criterion & $\rho_g^{(k)} \leq \tau_{\mathrm{sever}}$ \\
& Fragment selection score & Total GT-overlap mass \\
& Tie-breaker & Longer fragment \\
& Dominant-fragment scoring & Enabled \\
\midrule

\multirow{4}{*}{GA-STQ}
& Soft stability term & Enabled \\
& Area-penalty exponent $\gamma$ & 1.0 \\
& AQ normalization & Mean over GT instances \\
& GQ definition & Foreground coverage over valid evaluation domain $\Omega_t$ \\
\midrule

\multirow{5}{*}{GA-VPQ}
& VPQ matching threshold $\tau_{\mathrm{vpq}}$ & 0.5 \\
& Window lengths $k$ & 0, 5, 15, 25, 35, $\infty$ \\
& $k=0$ interpretation & Single-frame PQ \\
& $k=\infty$ interpretation & Full evaluated-sequence VPQ \\
& Fragment-level FP & $\sum_g \max(n_g^W-1,0)$ \\
\midrule

\multirow{5}{*}{IC diagnostics}
& Frame-level relaxed match IoU threshold & 0.5 \\
& Frame-level relaxed match IoS threshold & 0.5 \\
& Void-tolerant matching & Enabled when dataset policy uses void tolerance \\
& Redundancy factor for IC$_G$ & Disabled \\
& Weighted IC$_P$ & Disabled \\
\midrule

\multirow{4}{*}{Temporal diagnostics}
& Co-occurrence similarity threshold & 0.5 \\
& Maximum pattern size & 3 \\
& Cluster graph edge & Co-occurrence or adjacent handoff \\
& Patt$_k$ aggregation & Best co-existing $k$-identity support core \\
\midrule

\multirow{3}{*}{Void / ignore policy}
& VIPSeg ignore label & -1 in in-memory evaluation maps \\
& ScanNet ignore label & 0 \\
& HM3D ignore label & 0 \\
\bottomrule
\end{tabular}
}
\end{table*}

\subsection{Evaluation Hyperparameters}
\label{app:oga_hyperparameters}

Table~\ref{tab:oga_hyperparams} summarizes the standard evaluation configuration, shared across methods. The prediction-support threshold $\tau_{\mathrm{ios}}$ controls tube admissibility, and the sever threshold $\tau_{\mathrm{sever}}$ controls splitting between consecutive support states. Both are $0.5$ in the standard configuration.

For GA-STQ, dominant-fragment scoring, soft support-set stability, and the area-efficiency penalty are enabled. For GA-VPQ, the same dominant-fragment score is used as the window-level matching score, with a VPQ threshold of $0.5$. We report VPQ over windows $k\in\{0,5,15,25,35,\infty\}$, where $k=0$ denotes single-frame PQ and $k=\infty$ denotes VPQ over the full evaluated sequence.

For structural diagnostics, IP uses global valid-pixel overlaps scaled by the framewise discovery factor $D$ defined in Sec.~\ref{app:identity_persistence}; the IoS threshold affects IP through $D$, not by filtering overlap mass. IC is computed from the unweighted ever-support graph: IC$_P$ is the inverse number of GTs ever matched by a prediction, and IC$_G$ is the inverse number of predictions ever matched to a GT. We disable optional redundancy weighting and weighted IC$_P$ in the reported results to keep the diagnostic structural and interpretable. Temporal diagnostics use connected support components for Cluster and best co-existing support cores up to $k=3$ for Patt$_k$.

\clearpage
\subsection{Recommended Reporting Protocol}
\label{app:reporting_protocol}

For class-agnostic long-horizon OVS, we recommend the complementary reporting core below. These scores can be distributed across main-paper and appendix tables. Evaluate the same predictions under conventional and OGA protocols so that their comparison isolates the evaluation rule.

\begin{center}
\renewcommand{\arraystretch}{1.18}
\begin{tabular}{@{}p{0.21\linewidth}p{0.35\linewidth}p{0.38\linewidth}@{}}
\toprule
\textbf{Reporting block} & \textbf{Recommended scores} & \textbf{Interpretation} \\
\midrule
Spatial and temporal scales & Conventional and OGA VPQ$_0$, VPQ$_\infty$ & Single-frame quality and identity maintenance over the full evaluated sequence. \\
Association and spatial support & Conventional and OGA STQ, with their association and spatial components & Separate association from spatial support; identify the definition of each spatial term. \\
Identity organization & IP$_P$, IP$_G$, IC$_P$, IC$_G$ & Prediction-side purity/concentration and the distribution of reference-side support. \\
\bottomrule
\end{tabular}
\end{center}

When interpreting granularity sensitivity, also report VPQ's SQ/RQ decomposition to separate credited-match quality from matching counts. Name the STQ spatial term explicitly: in our VIPSeg comparison, conventional IoU and OGA foreground-coverage GQ measure different quantities. Finite-window VPQ profiles and the optional Cluster and Patt$_k$ diagnostics provide more detailed temporal analysis.

State the window units, frame sampling, evaluated duration, any prefix truncation, valid-domain and void policies, evaluation thresholds, and aggregation rule. VPQ$_\infty$ covers all retained evaluated frames, which may be a prefix of the source video. Appendix~\ref{app:frame_sampling} specifies our sampling implementation.

Read fidelity and structural diagnostics jointly. IP combines dominant-overlap purity or coverage with the discovery factor; IC measures identity concentration. Valid persistent parts can retain perfect OGA fidelity with lower reference-side IP/IC; the diagnostics describe support organization rather than additional fidelity targets.

\clearpage

\section{Dataset and Evaluation Protocol}
\label{app:dataset_protocol}

This section summarizes the dataset-specific preprocessing and evaluation protocols used in our experiments. 
All datasets are evaluated in a class-agnostic open-world setting: semantic category labels are removed, and methods are evaluated only by their ability to discover, segment, and maintain object identities over time. For ScanNet and HM3D (HM3D-Sem v0.2), all videos are sampled from their validation splits.
Unless otherwise stated, all reported OGA metrics use the same core evaluation functions described in Sec.~\ref{app:oga_details}.

Our benchmark suite is designed to cover two common modes of open-world video segmentation. 
VIPSeg represents the standard video panoptic segmentation regime, with dynamic objects, cluttered scenes, and panoptic instance annotations with semantic category labels. 
It is therefore useful for studying how conventional video segmentation metrics behave after semantic labels are removed but annotation granularity remains fixed. 
ScanNet and HM3D represent the long-horizon scene-centric regime, where videos contain dynamic ego-motion, repeated object re-observation, and loop-closure-like revisits in indoor environments. 
Together, these datasets cover both common panoptic video segmentation and realistic long-horizon open-world object maintenance.

Long-driving panoptic datasets such as Waymo Open Dataset and Cityscapes-VPS provide an important complementary direction, but we leave them to future work; the present suite already spans the two OVS modes central to this paper.

\subsection{VIPSeg}
\label{app:vipseg}

VIPSeg is used as a controlled testbed for granularity-agnostic evaluation under the standard video panoptic segmentation setting. Its videos contain moving objects, cluttered scenes, and annotation-defined object instances, making it a useful benchmark for exposing the limitation of rigid $1{:}1$ matching. In particular, VIPSeg allows us to isolate an important point: class-agnostic evaluation alone does not remove annotation-granularity bias. Even when semantic labels are ignored, a method can still be favored or penalized depending on whether its predicted object decomposition matches the benchmark annotation.

For our class-agnostic protocol, all VIPSeg categories are collapsed into a single foreground entity category. The RGB panoptic annotations and predictions are converted into plain instance maps, where each segment is represented only by its identity. Ignored pixels are represented as $-1$ in the in-memory evaluation maps, rather than in the panoptic PNG encoding.

Savvy predictions are exported in the standard VIPSeg panoptic submission format. Savvy produces a set of object masks per frame. These masks are flattened into a panoptic map using area-priority rasterization: larger masks are written first and smaller masks later, so visible small regions are not swallowed by overlapping large masks. The resulting panoptic PNGs and prediction JSON are then converted back into class-agnostic instance maps for OGA evaluation.

We report GA-adapted STQ, AQ, GQ, VPQ$_k$, VPQ$_\infty$, IP, IC, and temporal diagnostics. For VPQ, we use window lengths $k\in\{0,5,15,25,35,\infty\}$, where $k=0$ denotes the single-frame setting and $k=\infty$ denotes the full-video window.

\subsection{ScanNet}
\label{app:scannet}

ScanNet is used to represent long-horizon indoor exploration under real camera motion. Compared with VIPSeg, ScanNet sequences are longer and contain stronger viewpoint changes, repeated object re-observation, and loop-closure-like revisits. These properties make ScanNet a natural testbed for evaluating persistent object discovery and long-range identity maintenance.

ScanNet also introduces an important annotation challenge: because the reference labels are derived from reconstructed 3D scenes, some visually valid regions may be unlabeled or incomplete due to reconstruction coverage. For this reason, ScanNet is evaluated with a void-tolerant policy: prediction spill into GT-void regions is excluded from prediction-area and overlap accounting. This prevents a method from being penalized for segmenting visually valid regions that are absent from the reconstructed annotation.

The ScanNet protocol therefore emphasizes two aspects of OVS that are not fully captured by short panoptic videos: whether a method can maintain a stable object set over long temporal spans, and whether it can re-associate objects after significant viewpoint change or absence.

\subsection{HM3D}
\label{app:hm3d}

HM3D is used as a long-horizon indoor benchmark for evaluating open-world object maintenance under realistic embodied navigation. Unlike short panoptic videos, HM3D trajectories are designed to include extended ego-motion, repeated object re-observation, viewpoint change, and room-to-room transitions. This makes the dataset complementary to VIPSeg and ScanNet: VIPSeg tests standard video panoptic behavior under dynamic clutter, ScanNet tests real reconstructed indoor videos, and HM3D provides controlled long-horizon indoor trajectories with semantic and instance renderings.

\paragraph{HM3D-106 trajectory-composition benchmark.}
Among the 100 HM3D validation scenes available for trajectory collection, 36 provide the geometry, navigation meshes, and semantic annotations required for rendering and region-aware selection; the remaining 64 lack the required semantic annotations. Habitat-Sim exposes 569 semantic regions across the eligible scenes; this counts available regions, not regions visited by the trajectories. Reserving every third scene in sorted order yields 24 scenes for Composition~1 and 12 separate scenes for Composition~2.

The benchmark contains \textbf{72 Composition~1 within-region multi-loop scans}, \textbf{two 12-scan Composition~2 adjacent-region batches}, and \textbf{10 additional scans}. Composition~1 selects three distinct regions per scene, preferring broad region groups. Each scan connects two or three navigable loop centers within one region. Composition~2 generates one adjacent-region round trip per scene in each batch. The two batches use different region pairs; one second-batch scan shares a region ID with the first batch.

HM3D-106 contains \textbf{293,331 frames across 106 scans}. All scans provide JPEG RGB, 16-bit PNG semantic and instance labels, camera poses, and an instance-to-category lookup; depth is omitted. Each scan also has a 30 FPS RGB preview video; this is the preview encoding rate, not the evaluation sampling rate.

\paragraph{Trajectory design.}
The collection protocol for the 96 composition scans combines yaw sweeps, smoothed navigable connectors, and moderate pitch variation. Composition~1 uses a base density of 440 frames per $360^\circ$, yaw extents sampled in $[270^\circ,450^\circ]$, and two or three loop centers separated by 2--8\,m. Pitch amplitude is sampled independently per loop in $[6^\circ,12^\circ]$, with two sinusoidal cycles and a random phase. Connectors follow rounded Habitat-Sim shortest paths, with 0.01\,m spatial spacing, 0.005\,m spacing near junctions, and a $0.9^\circ$ per-frame yaw cap for connectors and transitions.

Composition~2 travels from region A to B and back to A, preferring different inferred region groups connected by a 2--20\,m navigable path. The endpoints have distinct region IDs, but need not have distinct category labels: scenes exposing only unknown categories can fall back to two different \texttt{other} regions. Initial and waypoint sweeps span $[90^\circ,180^\circ]$ and $[180^\circ,270^\circ]$, followed by a final $360^\circ$ loop at A. In-place sweeps use twice the base density, with smoothstep yaw progress when view counts are derived from yaw extent; connector/transition yaw is capped at $0.45^\circ$ per frame. These trajectories provide revisits with changes in scale, visibility, and viewpoint.

\paragraph{Evaluation role.}
HM3D-106 is used to test whether OVS methods can maintain a growing object set over long indoor trajectories. The benchmark stresses object discovery, re-association after absence, and identity stability under loop-closure-like revisits. Since both semantic and instance renderings are available, HM3D also allows us to analyze the effect of reference annotation granularity. The comparison across all 106 trajectories uses instance-level reference maps and the prefix-sampling protocol in Sec.~\ref{app:frame_sampling}; results are reported in Sec.~\ref{app:full_hm3d_results}. The explicit semantic-versus-instance robustness analysis uses a separate 10-scan subset in Sec.~\ref{app:hm3d_granularity_robustness}.

\subsection{Class-Agnostic Conversion}
\label{app:class_agnostic_conversion}

All evaluations remove semantic category labels and retain only instance identities. For VIPSeg, this is done by collapsing all panoptic categories into a single foreground entity category before constructing instance maps. For ScanNet and HM3D, reference maps are treated directly as class-agnostic object or instance identities. In all cases, OGA evaluates spatial support and temporal identity structure, not semantic classification.

This conversion is essential for OVS because the object set is not assumed to follow a predefined category ontology. However, it is not sufficient by itself. A class-agnostic metric can still impose a rigid annotation decomposition through $1{:}1$ matching. OGA therefore combines class-agnostic conversion with granularity-agnostic support construction, allowing valid predictions to be evaluated even when their semantic decomposition differs from the reference annotation.

\subsection{Void and Ignore-Region Handling}
\label{app:void_handling}

OGA separates the core matching rule from the dataset-specific valid evaluation domain $\Omega_t$. This is important because different datasets have different annotation completeness and ignore-label conventions.

For VIPSeg, ignored pixels are represented as $-1$ in the in-memory class-agnostic maps used by the evaluator. For ScanNet and HM3D, label $0$ is treated as void or background. In datasets with incomplete or partially reconstructed annotations, such as ScanNet, we use a void-tolerant evaluation domain: prediction pixels falling into GT-void regions are excluded from prediction-area and overlap computation. This prevents faithful predictions from being penalized merely because the reference annotation is incomplete.

For datasets with more complete annotations, such as HM3D, this tolerance can be reduced or removed by choosing a broader evaluation domain. In this sense, void handling is a dataset policy, not a defining component of GA matching. The GA principle remains the same: predictions are assigned to references by support, and reference objects are scored through coherent dominant fragments.

\subsection{Frame Sampling and Evaluation Windows}
\label{app:frame_sampling}

Long-horizon scene videos can contain hundreds or thousands of frames. To keep evaluation tractable while preserving temporal structure, ScanNet and HM3D use adaptive frame sampling. Let $T$ be the number of GT frames in a sequence. The sampling interval is
\begin{equation}
    \Delta =
    \begin{cases}
    1, & T\leq 300,\\
    2, & 300<T\leq 1000,\\
    3, & 1000<T\leq 1500,\\
    3, & T>1500.
    \end{cases}
\end{equation}
For every sequence, conventional and OGA evaluation sample the first $\min(T,1500)$ source frames at interval $\Delta$. This covers the complete sequence when $T\leq1500$ and an initial prefix otherwise.

For VPQ, we evaluate multiple temporal scales using sliding windows over the retained sampled frames. Unless otherwise stated, we report $k\in\{0,5,15,25,35,\infty\}$, where $k=0$ corresponds to single-frame PQ and $k=\infty$ evaluates all retained sampled frames as one window. Thus, VPQ$_\infty$ measures identity maintenance across the complete evaluated sequence or prefix.

\section{Analysis of Granularity Mismatch}
\label{app:granularity_mismatch}

This section provides additional analysis of the granularity mismatch problem discussed in the main paper. The goal is to clarify why class-agnostic evaluation alone is insufficient for open-world video segmentation. Even when semantic labels are removed, standard metrics still assume that predicted instances and reference instances should follow the same object decomposition. This assumption can incorrectly penalize valid open-world predictions that segment meaningful parts, sub-objects, or alternative object groupings.

OGA addresses this issue by treating the annotation as a reference decomposition rather than the only valid decomposition. Under GA matching, multiple predictions may jointly support one reference instance, as long as each prediction remains owned by at most one reference. This allows the evaluation to distinguish granularity mismatch from true segmentation or tracking failure.

\subsection{Qualitative Failure of Rigid 1:1 Matching on VIPSeg}
\label{app:vipseg_qualitative_matching}

VIPSeg provides a useful testbed for visualizing granularity mismatch because it contains standard video panoptic annotations with annotation-defined object instances. In such benchmarks, a predicted mask may be visually meaningful and temporally stable but still fail under rigid $1{:}1$ matching if it corresponds to a part-level decomposition of an annotated object.

\begin{figure}[t]
    \centering
\includegraphics[width=\textwidth]{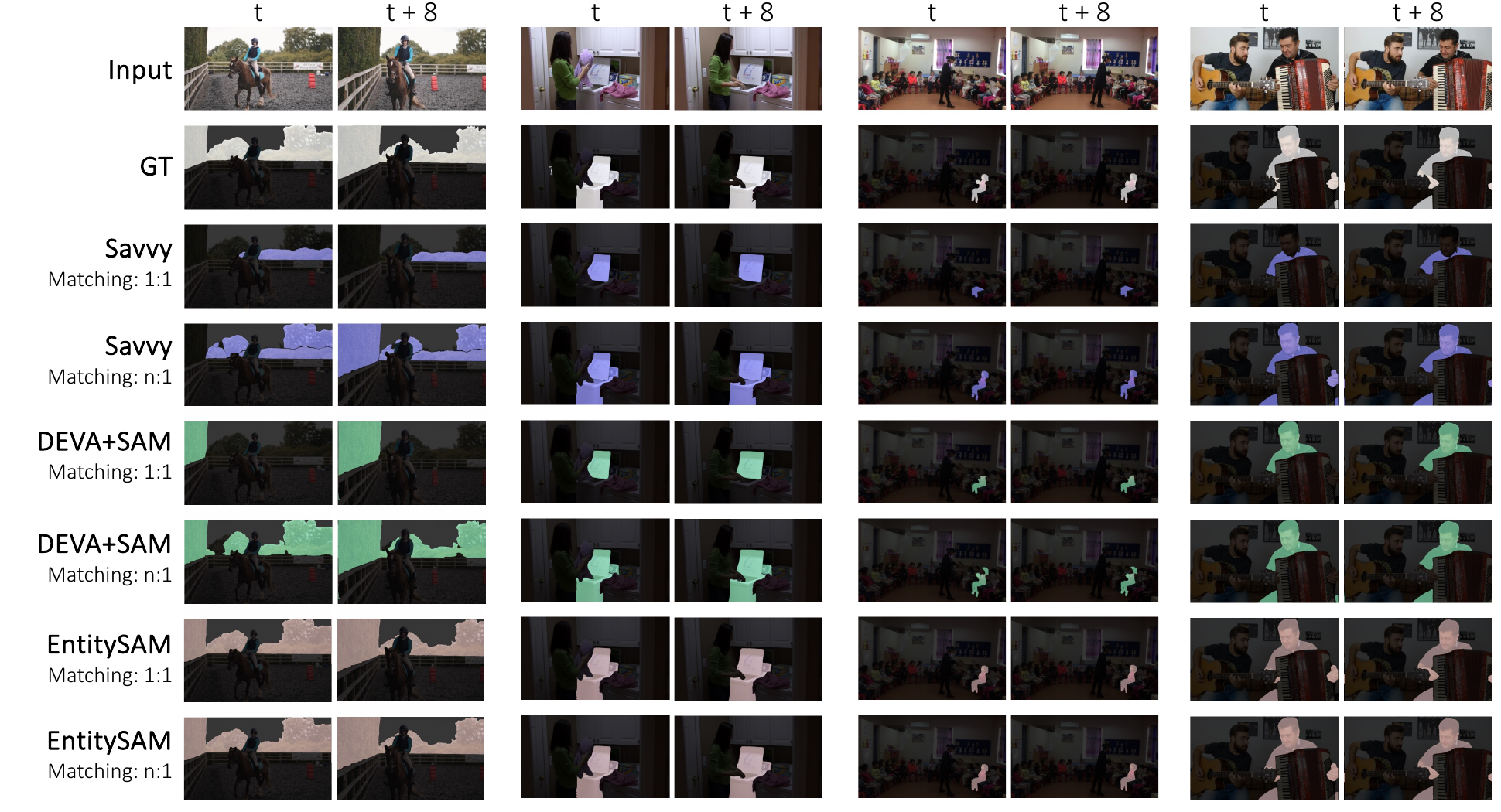}
    
\caption{
\textbf{Rigid $1{:}1$ matching underestimates valid open-world support.}
For each VIPSeg example, we show two frames $(t,t+8)$, the GT reference object, and the support credited to each method under rigid $1{:}1$ matching versus GA $n{:}1$ matching.
Rows labeled $1{:}1$ and $n{:}1$ visualize the support credited by the evaluation rule, not different model predictions.
Under $1{:}1$ matching, only one predicted identity can explain the reference object, so valid part-level predictions are discarded from credited support and may be counted as false positives or fragmentation errors.
Under GA matching, multiple reference-contained predictions can jointly support the same GT instance while each prediction remains assigned to at most one GT.
This exposes the central limitation of class-agnostic but rigid evaluation: even without semantic labels, valid open-world support can be unfairly penalized when its granularity differs from the annotation.
}
\label{fig:vipseg_granularity_failure}
\end{figure}

Figure~\ref{fig:vipseg_granularity_failure} illustrates this failure mode. A reference object may be decomposed by an open-world method into several coherent prediction identities. Under strict $1{:}1$ matching, at most one prediction can be credited to the reference object. The remaining predictions are treated as unmatched false positives or identity fragments, even if they are spatially contained in the reference object and correspond to semantically meaningful parts. This produces an artificially low score for methods whose granularity differs from the annotation.

Under GA matching, these predictions are instead interpreted as a support set for the same reference instance. This does not mean that all fragments are freely credited: the support must satisfy the prediction-side ownership rule, and temporal discontinuities are still handled by dominant-fragment selection. Thus, GA matching relaxes the granularity assumption while preserving identity and temporal constraints.

\subsection{Why Class-Agnostic Evaluation Alone Is Insufficient}
\label{app:class_agnostic_insufficient}

A natural first step for evaluating open-world segmentation is to remove semantic labels and evaluate only instance masks. This is necessary, but not sufficient. Class-agnostic evaluation removes the category-classification requirement, but it does not remove the assumption that predictions and annotations must share the same instance granularity.

This distinction is important. In a category-aware setting, a method can fail because it assigns the wrong semantic label. In a class-agnostic setting, this failure mode is removed. However, a method can still be penalized if it predicts a valid decomposition that differs from the benchmark's annotation convention. For example, an open-world method may segment a bicycle into wheels, frame, and handlebar, while the annotation treats the entire bicycle as one instance. These are not necessarily failures of open-world perception; they are failures to match a fixed reference granularity.

Rigid $1{:}1$ matching conflates these cases with true errors. If a reference object is supported by several predicted parts, the best-matching part may receive partial credit, while the remaining valid parts become false positives. In this case, the metric penalizes a meaningful part-level decomposition simply because it violates annotation-aligned granularity.

It is important, however, to distinguish the two directions of granularity mismatch. The case where multiple predictions support one reference instance corresponds to an $n{:}1$ relation. This often arises when an open-world method predicts meaningful parts of an annotated object, and is the direction that OGA explicitly relaxes. The converse case, where one prediction overlaps multiple reference instances, corresponds to a $1{:}n$ relation. This is less desirable as a relaxation because it may indicate over-grouping or identity bleeding: a single predicted identity is being shared across multiple annotated objects.

For this reason, OGA keeps the reference side as the anchor. Each prediction may be assigned to at most one reference instance, while each reference instance may receive support from multiple predictions. This asymmetric design avoids uncontrolled many-to-many credit assignment. If both directions were relaxed simultaneously, a large ambiguous prediction could be partially credited to many references, and many predictions could also jointly explain each reference. The evaluation graph would become difficult to interpret, and the metric would no longer provide a controlled diagnostic of object identity structure.

OGA credits sufficiently contained $n{:}1$ part support while assigning each prediction to at most one reference within each fidelity-scoring scope. Dominant-fragment selection and fragment-level VPQ penalties assess temporal continuity, while the two-sided diagnostics reveal identity bleeding and fragmentation. Together, these components distinguish annotation-granularity mismatch from errors in spatial support and identity maintenance.

This distinction explains the VIPSeg results in the main paper. Methods without benchmark-aligned granularity recover substantially under GA evaluation, because part-level or alternative decompositions are no longer automatically discarded. Methods with annotation-aligned granularity benefit less, because their predictions already conform more closely to the rigid $1{:}1$ assumption. The resulting comparison is more informative: it separates benchmark granularity alignment from genuine open-world identity maintenance.

\section{Additional Experimental Results}
\label{app:additional_results}

\begin{table*}[t]
\centering
\caption{\textbf{Rigid $1{:}1$ versus granularity-agnostic evaluation on VIPSeg.} Methods without benchmark-aligned granularity recover the most when valid $n{:}1$ support is credited. Parentheses show the absolute change from rigid evaluation.}
\label{tab:vipseg}
\resizebox{\textwidth}{!}{
\begin{tabular}{lccccccc}
\toprule
\textbf{Method} & \textbf{Eval.} & \textbf{VPQ$_{\infty}$} & \textbf{SQ} & \textbf{RQ} & \textbf{STQ} & \textbf{AQ} & \textbf{IoU/GQ} \\
\midrule
\multirow{2}{*}{EntitySAM~\cite{entitysam}}
& $1{:}1$ & 54.68 & 84.68 & 64.57 & 43.67 & 42.92 & 44.42 \\
& GA & \textbf{59.23} \textcolor{posdelta}{(+4.55)} & 84.50 \textcolor{negdelta}{(-0.18)} & \textbf{69.52} \textcolor{posdelta}{(+4.95)} & 65.13 \textcolor{posdelta}{(+21.46)} & 49.70 \textcolor{posdelta}{(+6.78)} & 88.43 \textcolor{posdelta}{(+44.01)} \\
\midrule
\multirow{2}{*}{SAM2~\cite{sam2}}
& $1{:}1$ & 24.92 & 81.85 & 30.45 & 37.23 & 37.52 & 36.94 \\
& GA & 51.76 \textcolor{posdelta}{(+26.84)} & \textbf{85.40} \textcolor{posdelta}{(+3.55)} & 60.09 \textcolor{posdelta}{(+29.64)} & 60.91 \textcolor{posdelta}{(+23.68)} & 51.21 \textcolor{posdelta}{(+13.69)} & 74.62 \textcolor{posdelta}{(+37.68)} \\
\midrule
\multirow{2}{*}{DEVA+SAM~\cite{deva}}
& $1{:}1$ & 28.52 & 83.44 & 34.18 & 44.85 & 46.08 & 43.66 \\
& GA & 53.05 \textcolor{posdelta}{(+24.53)} & 82.18 \textcolor{negdelta}{(-1.26)} & 63.04 \textcolor{posdelta}{(+28.86)} & 68.24 \textcolor{posdelta}{(+23.39)} & 54.91 \textcolor{posdelta}{(+8.83)} & 86.59 \textcolor{posdelta}{(+42.93)} \\
\midrule
\multirow{2}{*}{\textbf{Savvy (Ours)}}
& $1{:}1$ & 26.15 & 83.38 & 31.35 & 46.29 & 48.42 & 44.26 \\
& GA & 55.35 \textcolor{posdelta}{\textbf{(+29.20)}} & 82.50 \textcolor{negdelta}{(-0.88)} & 66.43 \textcolor{posdelta}{\textbf{(+35.08)}} & \textbf{72.09} \textcolor{posdelta}{\textbf{(+25.80)}} & \textbf{59.01} \textcolor{posdelta}{(+10.59)} & \textbf{89.53} \textcolor{posdelta}{\textbf{(+45.27)}} \\
\bottomrule
\end{tabular}
}
\end{table*}

\subsection{Full VIPSeg Results}
\label{app:full_vipseg_results}

Table~\ref{tab:vipseg} compares conventional and GA scores on VIPSeg. VIPSeg is useful here as a controlled video-panoptic testbed whose annotation-defined object granularity exposes the metric effect cleanly; it is not our primary long-horizon benchmark. VPQ$_\infty$ rises for all four methods under GA, while some other metric terms decrease, so class-agnostic evaluation alone does not eliminate sensitivity to a differently granular decomposition.

The size of the recovery depends strongly on benchmark alignment. Savvy improves from 26.15 to 55.35 in VPQ$_\infty$, SAM2 from 24.92 to 51.76, and DEVA+SAM from 28.52 to 53.05. EntitySAM, whose coarser predictions more closely follow the benchmark decomposition, improves only from 54.68 to 59.23. This comparison motivates the asymmetric $n{:}1$ support rule: several contained predictions may explain one reference object, while each prediction receives one GA assignment per evaluation window.

EntitySAM's conventional VPQ advantage is concentrated in one-to-one recognition: its SQ is close to Savvy's (84.68 versus 83.38), whereas its RQ is substantially higher (64.57 versus 31.35). EntitySAM trains its entity modules on COCO panoptic masks with category labels removed~\cite{entitysam}. The observed matching advantage is consistent with this supervision supplying a partition prior compatible with VIPSeg annotations. RQ reflects matching counts, including missed and spurious tracks; AQ provides the complementary association view.

Under OGA, the RQ gap narrows to 69.52 versus 66.43. Savvy achieves higher AQ under both conventional evaluation (48.42 versus 42.92) and OGA (59.01 versus 49.70), and the strongest OGA STQ, whereas EntitySAM retains the highest VPQ$_\infty$. SQ changes little and can decrease slightly. Together, the breakdown distinguishes matching to the annotation partition from temporal association quality.

\subsection{Full ScanNet Results}
\label{app:full_scannet_results}

The conventional ScanNet scores in Table~\ref{tab:scannet_rigid_full} and the OGA scores in Table~\ref{tab:scannet_hm3d_fidelity} use the same 142 validation scenes and macro-average scene scores. The full conventional window profile enables direct comparison across temporal scales.

\begin{table}[H]
\centering
\setlength{\tabcolsep}{7pt}
\caption{\textbf{Conventional fidelity results on all 142 ScanNet validation scenes.} VPQ uses $1{:}1$ tube matches; STQ uses all-overlap association. Scores are percentages, macro-averaged over scenes; no GA reassociation is applied.}
\label{tab:scannet_rigid_full}
\resizebox{\linewidth}{!}{%
\begin{tabular}{l|ccccccccc}
\toprule
\textbf{Method} & \textbf{VPQ$_0$} & \textbf{VPQ$_5$} & \textbf{VPQ$_{15}$} & \textbf{VPQ$_{25}$} & \textbf{VPQ$_{35}$} & \textbf{VPQ$_\infty$} & \textbf{STQ} & \textbf{AQ} & \textbf{GQ} \\
\midrule
EntitySAM & 33.30 & 31.56 & 29.22 & 27.46 & 25.89 & 8.92 & 22.35 & 8.75 & 61.61 \\
DEVA+SAM & 33.23 & 31.62 & 29.97 & 29.05 & 28.28 & 11.52 & 46.66 & 23.32 & \textbf{95.87} \\
Savvy (Ours) & \textbf{38.26} & \textbf{35.82} & \textbf{33.20} & \textbf{31.71} & \textbf{30.63} & \textbf{18.44} & \textbf{53.84} & \textbf{32.29} & 91.11 \\
\bottomrule
\end{tabular}}
\end{table}

Table~\ref{tab:scannet_hm3d_fidelity} reports the full OGA fidelity results on ScanNet. ScanNet represents the long-horizon indoor exploration setting, where strong ego-motion, repeated re-observation, and incomplete reconstruction make persistent object maintenance substantially more difficult than in short panoptic videos. Across all VPQ windows, Savvy achieves the strongest performance. In particular, Savvy improves VPQ$_{\infty}$ from 18.53 to 28.63 over DEVA+SAM and from 10.82 to 28.63 over EntitySAM. Savvy also achieves the best STQ and AQ, indicating stronger long-horizon association quality under GA evaluation.

DEVA+SAM obtains the highest GQ on ScanNet, which suggests strong foreground coverage over the valid region. However, its lower AQ and VPQ$_{\infty}$ indicate that this coverage does not translate into equally strong long-range identity maintenance. EntitySAM has lower AQ and VPQ$_{\infty}$, consistent with the sparser discovered object set and coarser predictions observed in the qualitative comparisons.

Table~\ref{tab:scannet_hm3d_diagnostics} provides the corresponding structural diagnostics. Prediction-axis diagnostics measure whether predicted identities remain pure, while reference-axis diagnostics measure whether GT objects remain whole. Both axes are necessary: high IP$_P$ or IC$_P$ alone can be misleading, because a method may produce many short-lived or spatially fragmented predictions that each remain pure with respect to one GT while still splitting that GT across many identities. Savvy achieves the best scores on both prediction-side and reference-side diagnostics, indicating that its identities are not only relatively pure on the prediction axis, but also provide more persistent and compact support for reference objects. Its gains on IP$_G$, IC$_G$, Cluster, and Patt$_k$ show that the improved prediction-side purity is accompanied by stronger reference-side persistence and temporal organization.

The size-stratified results in Table~\ref{tab:scannet_stratified} further show that Savvy's advantage is not limited to large objects. The diagnostics should again be read jointly across axes: prediction-side purity alone does not guarantee good object maintenance unless it is accompanied by reference-side persistence and concentration. Savvy obtains the best structural diagnostics on medium and large objects across all reported measures, and also improves most small-object diagnostics. The only exception is IP$_G$ for small objects, where DEVA+SAM is slightly higher. However, Savvy remains stronger on small-object IP$_P$, IC$_P$, IC$_G$, Cluster, and Patt$_k$, suggesting cleaner and more temporally organized support even in the difficult small-object regime.

\begin{table*}[t]
\centering
\setlength{\tabcolsep}{10pt} 
\caption{\textbf{Detailed OGA fidelity results on ScanNet.} Primary OVS fidelity metrics and varied VPQ windows. Scores are percentages.}\vspace{-5pt}
\label{tab:scannet_hm3d_fidelity}
\resizebox{\linewidth}{!}{
\begin{tabular}{@{}c@{\hspace{11pt}}@{}l | ccccccccc }
\toprule
& & \multicolumn{9}{c}{\textbf{Fidelity \& VPQ Windows $\uparrow$}} \\
\cmidrule(lr){3-11} 
& \textbf{Method} & \textbf{VPQ$_{0}$} & \textbf{VPQ$_{5}$} & \textbf{VPQ$_{15}$} & \textbf{VPQ$_{25}$} & \textbf{VPQ$_{35}$} & \textbf{VPQ$_{\infty}$} & \textbf{STQ} & \textbf{AQ} & \textbf{GQ} \\

\midrule
& EntitySAM & 32.31 & 30.54 & 28.45 & 26.85 & 25.48 & 10.82 & 17.65 & 6.27 & 61.61 \\
& DEVA+SAM & 43.58 & 40.86 & 39.12 & 38.05 & 37.16 & 18.53 & 43.15 & 20.35 & \textbf{95.87} \\
\multirow{-3}{*}{\rotatebox[origin=c]{90}{\small\textbf{ScanNet}}} 
& \cellcolor{ac!30}Savvy (Ours) 
& \cellcolor{ac!30}\textbf{58.48} 
& \cellcolor{ac!30}\textbf{54.16} 
& \cellcolor{ac!30}\textbf{51.58} 
& \cellcolor{ac!30}\textbf{50.12} 
& \cellcolor{ac!30}\textbf{48.98} 
& \cellcolor{ac!30}\textbf{28.63} 
& \cellcolor{ac!30}\textbf{53.93} 
& \cellcolor{ac!30}\textbf{32.59} 
& \cellcolor{ac!30}91.11 \\

\bottomrule

\end{tabular}
}
\end{table*}

\begin{table*}[t]
\centering
\setlength{\tabcolsep}{12pt} 
\caption{\textbf{Detailed OGA structural diagnostics on ScanNet.} Structural diagnostics for identity persistence (\textbf{IP}), identity concentration (\textbf{IC}), and temporal patterns. Scores are percentages.}\vspace{-5pt}
\label{tab:scannet_hm3d_diagnostics}
\resizebox{\linewidth}{!}{
\begin{tabular}{@{}c@{\hspace{20pt}}@{}l | CCCC | CCCC }
\toprule
& & \multicolumn{4}{c|}{\textbf{Persistence \& Concentration $\uparrow$}} & \multicolumn{4}{c}{\textbf{Clustering \& Patterns $\uparrow$}}\\
\cmidrule(lr){3-6} \cmidrule(lr){7-10} 
& \textbf{Method} & \textbf{IP$_P$} & \textbf{IP$_G$} & \textbf{IC$_P$} & \textbf{IC$_G$} & \textbf{Clust.} & \textbf{Patt$_1$} & \textbf{Patt$_2$} & \textbf{Patt$_3$}\\

\midrule
& EntitySAM & 24.93 & 17.31 & 9.02 & 19.19 & 26.65 & 25.91 & 1.88 & 0.18 \\
& DEVA+SAM & 63.21 & 50.38 & 54.82 & 25.90 & 43.09 & 40.06 & 12.82 & 5.63 \\
\multirow{-3}{*}{\rotatebox[origin=c]{90}{\small\textbf{ScanNet}}} 
& \cellcolor{ac!30}Savvy (Ours) 
& \cellcolor{ac!30}\textbf{75.44} 
& \cellcolor{ac!30}\textbf{57.92} 
& \cellcolor{ac!30}\textbf{62.68} 
& \cellcolor{ac!30}\textbf{32.03} 
& \cellcolor{ac!30}\textbf{65.71} 
& \cellcolor{ac!30}\textbf{58.67} 
& \cellcolor{ac!30}\textbf{20.58} 
& \cellcolor{ac!30}\textbf{9.50} \\

\bottomrule
\end{tabular}
}
\end{table*}

\begin{table*}[t]
\centering
\setlength{\tabcolsep}{12pt} 
\caption{\textbf{Size-Stratified OGA Structural Diagnostics on ScanNet.} Performance broken down by object size: Small (\textbf{S}), Medium (\textbf{M}), and Large (\textbf{L}).}\vspace{-5pt}
\label{tab:scannet_stratified}
\resizebox{\linewidth}{!}{
\begin{tabular}{l c | CCCC | CCCC }
\toprule
& & \multicolumn{4}{c|}{\textbf{Persistence \& Concentration $\uparrow$}} & \multicolumn{4}{c}{\textbf{Clustering \& Patterns}}\\
\cmidrule(lr){3-6} \cmidrule(lr){7-10} 
\textbf{Method} & \textbf{Size} & \textbf{IP$_{p}$} & \textbf{IP$_{g}$} & \textbf{IC$_{p}$} & \textbf{IC$_{g}$} & \textbf{Clust.} & \textbf{Patt$_{1}$} & \textbf{Patt$_{2}$} & \textbf{Patt$_{3}$}\\

\midrule
& S & 0.75 & 10.90 & 0.26 & 6.42 & 1.66 & 1.65 & 0.00 & 0.00 \\
EntitySAM & M & 7.82 & 11.74 & 1.80 & 13.60 & 8.56 & 8.29 & 0.48 & 0.01 \\
& L & 25.67 & 20.21 & 9.97 & 21.85 & 37.15 & 35.97 & 2.84 & 0.31 \\

\midrule
& S & 58.88 & \textbf{54.37} & 54.08 & 16.17 & 7.22 & 7.07 & 0.59 & 0.02 \\
DEVA+SAM & M & 66.59 & 50.40 & 59.25 & 34.21 & 29.09 & 27.83 & 5.69 & 1.66 \\
& L & 65.81 & 49.92 & 53.76 & 22.10 & 52.45 & 48.09 & 17.30 & 8.14 \\

\midrule
\cellcolor{ac!30} & \cellcolor{ac!30}S & \cellcolor{ac!30}\textbf{69.66} & \cellcolor{ac!30}48.03 & \cellcolor{ac!30}\textbf{62.03} & \cellcolor{ac!30}\textbf{24.07} & \cellcolor{ac!30}\textbf{12.55} & \cellcolor{ac!30}\textbf{11.44} & \cellcolor{ac!30}\textbf{2.25} & \cellcolor{ac!30}\textbf{0.42} \\
\cellcolor{ac!30}\multirow{-2}{*}{Savvy (Ours)} & \cellcolor{ac!30}M & \cellcolor{ac!30}\textbf{78.30} & \cellcolor{ac!30}\textbf{55.12} & \cellcolor{ac!30}\textbf{66.34} & \cellcolor{ac!30}\textbf{44.50} & \cellcolor{ac!30}\textbf{51.37} & \cellcolor{ac!30}\textbf{48.01} & \cellcolor{ac!30}\textbf{10.43} & \cellcolor{ac!30}\textbf{3.59} \\
\cellcolor{ac!30} & \cellcolor{ac!30}L & \cellcolor{ac!30}\textbf{81.44} & \cellcolor{ac!30}\textbf{59.79} & \cellcolor{ac!30}\textbf{62.83} & \cellcolor{ac!30}\textbf{25.66} & \cellcolor{ac!30}\textbf{76.12} & \cellcolor{ac!30}\textbf{66.57} & \cellcolor{ac!30}\textbf{26.79} & \cellcolor{ac!30}\textbf{13.15} \\
\bottomrule

\end{tabular}
}
\end{table*}

\begin{table*}[t]
\centering
\setlength{\tabcolsep}{12pt} 
\caption{\textbf{Size-Stratified OGA Structural Diagnostics on HM3D.} Performance broken down by object size: Small (\textbf{S}), Medium (\textbf{M}), and Large (\textbf{L}). Scores are percentages.}
\vspace{-5pt}
\label{tab:hm3d_stratified}
\resizebox{\linewidth}{!}{
\begin{tabular}{l c | CCCC | CCCC }
\toprule
& & \multicolumn{4}{c|}{\textbf{Persistence \& Concentration $\uparrow$}} & \multicolumn{4}{c}{\textbf{Clustering \& Patterns $\uparrow$}}\\
\cmidrule(lr){3-6} \cmidrule(lr){7-10} 
\textbf{Method} & \textbf{Size} & \textbf{IP$_P$} & \textbf{IP$_G$} & \textbf{IC$_P$} & \textbf{IC$_G$} & \textbf{Clust.} & \textbf{Patt$_1$} & \textbf{Patt$_2$} & \textbf{Patt$_3$}\\

\midrule
& S & 0.05 & 5.43 & 0.08 & 1.94 & 0.28 & 0.28 & 0.00 & 0.00 \\
EntitySAM & M & 3.60 & 4.57 & 4.46 & 20.18 & 4.42 & 4.37 & 0.12 & 0.00 \\
& L & 8.30 & 9.16 & 15.48 & \textbf{54.86} & 22.67 & 22.41 & 0.57 & 0.05 \\

\midrule
& S & 35.54 & 26.29 & 79.86 & 13.09 & 3.33 & 3.32 & 0.07 & 0.01 \\
DEVA+SAM & M & 36.61 & 25.72 & \textbf{78.55} & 47.32 & 23.50 & 23.23 & 2.67 & 0.59 \\
& L & 34.18 & 26.01 & \textbf{67.04} & 33.28 & 38.59 & 36.93 & 9.86 & 3.46 \\

\midrule
\cellcolor{ac!30} & \cellcolor{ac!30}S 
& \cellcolor{ac!30}\textbf{53.25} 
& \cellcolor{ac!30}\textbf{33.40} 
& \cellcolor{ac!30}\textbf{82.95} 
& \cellcolor{ac!30}\textbf{16.08} 
& \cellcolor{ac!30}\textbf{7.77} 
& \cellcolor{ac!30}\textbf{7.69} 
& \cellcolor{ac!30}\textbf{0.25} 
& \cellcolor{ac!30}\textbf{0.02} \\
\cellcolor{ac!30}\multirow{-2}{*}{Savvy (Ours)} 
& \cellcolor{ac!30}M 
& \cellcolor{ac!30}\textbf{53.74} 
& \cellcolor{ac!30}\textbf{40.63} 
& \cellcolor{ac!30}77.58 
& \cellcolor{ac!30}\textbf{58.46} 
& \cellcolor{ac!30}\textbf{44.35} 
& \cellcolor{ac!30}\textbf{42.70} 
& \cellcolor{ac!30}\textbf{4.92} 
& \cellcolor{ac!30}\textbf{1.05} \\
\cellcolor{ac!30} & \cellcolor{ac!30}L 
& \cellcolor{ac!30}\textbf{51.97} 
& \cellcolor{ac!30}\textbf{42.03} 
& \cellcolor{ac!30}63.36 
& \cellcolor{ac!30}37.72 
& \cellcolor{ac!30}\textbf{66.62} 
& \cellcolor{ac!30}\textbf{58.09} 
& \cellcolor{ac!30}\textbf{14.58} 
& \cellcolor{ac!30}\textbf{5.71} \\
\bottomrule

\end{tabular}
}
\end{table*}

\subsection{Full HM3D Results}
\label{app:full_hm3d_results}

Table~\ref{tab:hm3d106_full} gives the complete conventional and OGA fidelity results, macro-averaged over the same 106 HM3D trajectories using instance-level reference maps for every method and protocol. Under conventional evaluation, Savvy leads all VPQ windows, STQ, and AQ. In particular, its VPQ$_\infty$ is 17.31 versus 6.58 for DEVA+SAM and 1.35 for EntitySAM. This ranking is already present before OGA. DEVA+SAM has the highest GQ (94.81 versus Savvy's 91.26), but that geometry advantage does not translate to stronger temporal association.

Under OGA, Savvy still leads all VPQ windows, STQ, and AQ, reaching VPQ$_\infty$ of 19.59 versus 8.47 for DEVA+SAM and 1.46 for EntitySAM. The rigid-to-OGA VPQ$_\infty$ changes are +2.28, +1.89, and +0.11 points, respectively. They are notably smaller than on ScanNet, and the change is not uniformly positive across metrics: STQ and AQ fall for all three methods, and EntitySAM's VPQ$_0$ falls from 17.38 to 14.97. These paired scores expose how reference-anchored support and temporal coherence affect each fidelity component.

Table~\ref{tab:hm3d106_diagnostics} reports the OGA structural diagnostics over these same 106 trajectories; unlike the fidelity metrics, IP and IC have no conventional counterpart. Savvy leads both persistence axes (IP$_P$ 53.15 and IP$_G$ 37.07) and reference-side concentration (IC$_G$ 32.16). DEVA+SAM narrowly leads prediction-side concentration (IC$_P$ 74.56 versus 74.46), so prediction-side concentration alone does not capture Savvy's stronger reference-side organization. These full-106 diagnostics are distinct from the 10-scan semantic-versus-instance analysis in Sec.~\ref{app:hm3d_granularity_robustness}. ScanNet structural diagnostics are reported separately in Table~\ref{tab:scannet_hm3d_diagnostics}.

\begin{table*}[t]
\centering
\setlength{\tabcolsep}{6pt}
\caption{\textbf{Complete instance-level HM3D-106 fidelity results.} Conventional (VPQ: $1{:}1$) and OGA (asymmetric $n{:}1$ support) scores are macro-averaged over the same 106 trajectories and shown on a 0--100 scale. Both protocols score the same predictions. Full-106 OGA structural diagnostics appear in Table~\ref{tab:hm3d106_diagnostics}.}
\label{tab:hm3d106_full}
\resizebox{\linewidth}{!}{%
\begin{tabular}{ll|ccccccccc}
\toprule
\textbf{Method} & \textbf{Protocol} & \textbf{VPQ$_0$} & \textbf{VPQ$_5$} & \textbf{VPQ$_{15}$} & \textbf{VPQ$_{25}$} & \textbf{VPQ$_{35}$} & \textbf{VPQ$_\infty$} & \textbf{STQ} & \textbf{AQ} & \textbf{GQ} \\
\midrule
EntitySAM & Conventional & 17.38 & 15.54 & 12.68 & 10.72 & 9.22 & 1.35 & 10.50 & 1.80 & 64.37 \\
& OGA $n{:}1$ & 14.97 & 13.32 & 10.92 & 9.30 & 8.10 & 1.46 & 5.97 & 0.74 & 64.37 \\
\midrule
DEVA+SAM & Conventional & 29.97 & 27.77 & 25.26 & 23.77 & 22.70 & 6.58 & 31.34 & 10.58 & \textbf{94.81} \\
& OGA $n{:}1$ & 30.41 & 28.03 & 25.68 & 24.26 & 23.26 & 8.47 & 25.26 & 6.94 & \textbf{94.81} \\
\midrule
Savvy (Ours) & Conventional & \textbf{42.25} & \textbf{38.89} & \textbf{35.39} & \textbf{33.59} & \textbf{32.41} & \textbf{17.31} & \textbf{40.35} & \textbf{18.12} & 91.26 \\
& OGA $n{:}1$ & \textbf{46.77} & \textbf{42.97} & \textbf{39.48} & \textbf{37.64} & \textbf{36.41} & \textbf{19.59} & \textbf{35.69} & \textbf{14.28} & 91.26 \\
\bottomrule
\end{tabular}}
\end{table*}

\begin{table*}[t]
\centering
\setlength{\tabcolsep}{10pt}
\caption{\textbf{Full instance-level HM3D-106 OGA structural diagnostics.} Per-trajectory n:1 scores are macro-averaged over the same 106 trajectories as Table~\ref{tab:hm3d106_full} and scaled to 0--100. IP/IC measure persistence/concentration on the prediction (P) and reference (G) axes. Clust. and Patt$_k$ summarize temporal support structure. These diagnostics have no standard rigid $1{:}1$ counterpart.}
\label{tab:hm3d106_diagnostics}
\resizebox{\linewidth}{!}{%
\begin{tabular}{l|cccc|cccc}
\toprule
& \multicolumn{4}{c|}{\textbf{Persistence and concentration $\uparrow$}} & \multicolumn{4}{c}{\textbf{Temporal structure $\uparrow$}} \\
\cmidrule(lr){2-5} \cmidrule(lr){6-9}
\textbf{Method} & \textbf{IP$_P$} & \textbf{IP$_G$} & \textbf{IC$_P$} & \textbf{IC$_G$} & \textbf{Clust.} & \textbf{Patt$_1$} & \textbf{Patt$_2$} & \textbf{Patt$_3$} \\
\midrule
EntitySAM & 8.36 & 5.80 & 14.46 & 15.06 & 4.74 & 4.69 & 0.12 & 0.01 \\
DEVA+SAM & 35.55 & 26.11 & \textbf{74.56} & 26.57 & 14.59 & 14.27 & 2.27 & 0.67 \\
Savvy (Ours) & \textbf{53.15} & \textbf{37.07} & 74.46 & \textbf{32.16} & \textbf{27.32} & \textbf{25.59} & \textbf{3.75} & \textbf{1.16} \\
\bottomrule
\end{tabular}}
\end{table*}

\subsection{Robustness to HM3D Annotation Granularity}
\label{app:hm3d_granularity_robustness}

We additionally evaluate how OGA results change under different HM3D reference granularities. This analysis is a diagnostic robustness check distinct from the full-106 comparison. Semantic-level annotations collapse multiple physical instances into coarser reference regions, thereby weakening the identity-maintenance requirement that is central to OVS. As a result, semantic-level evaluation can benefit methods with coarser predictions or weaker instance-level object management, since merging or reusing identities across objects of the same semantic type is less strongly penalized. Instance-level references therefore provide a stricter test of object identity maintenance, while the paired semantic-level results expose sensitivity to annotation granularity.

This granularity comparison is evaluated on a controlled 10-scan subset selected from our 106-scan HM3D benchmark. The subset is chosen to reduce confounding from incomplete sequence metadata or severe method failure, so that the comparison more directly reflects the effect of reference annotation granularity.

Table~\ref{tab:hm3d_semantic_instance} compares results under semantic and instance reference decompositions. Changing the reference granularity affects all methods, confirming that OVS evaluation is sensitive to how the reference object set is defined. The shift to instance-level GT reduces most fidelity and persistence scores, but the magnitude differs sharply across methods. EntitySAM drops most severely, with VPQ$_0$ falling from 36.65 to 18.71, VPQ$_\infty$ from 14.65 to 2.90, and STQ from 24.04 to 10.28, indicating that its coarse partitions benefit substantially from semantic-level reference collapse. DEVA+SAM also drops under instance GT, especially in IP$_P$ and IP$_G$. Savvy is affected as well, but remains substantially stronger than both baselines on VPQ$_0$, VPQ$_\infty$, STQ, and AQ. Interestingly, IC$_G$ increases for all methods under instance GT. Finer reference objects can reduce the number of prediction IDs touching each individual GT, increasing concentration, while imposing stricter requirements on persistence and object-level fidelity.

Despite this change in reference decomposition, Savvy remains strongest on the primary OVS fidelity metrics under both settings. This suggests that Savvy's advantage is not tied to one particular HM3D annotation convention. The comparison also illustrates why instance-level evaluation is more appropriate for OVS: it better exposes whether a method can maintain distinct object identities rather than merely cover broad semantic regions. Semantic-level results are therefore useful for diagnosing annotation-granularity sensitivity, but they should not replace instance-level evaluation when the goal is open-world object maintenance.

\begin{table*}[t]
\centering
\small
\setlength{\tabcolsep}{8pt}
\caption{
\textbf{HM3D robustness under semantic versus instance annotation granularity.}
Scores are percentages.
Changing the reference granularity affects all methods, but Savvy remains strongest on the primary OVS fidelity metrics under both semantic and instance annotations.
}
\label{tab:hm3d_semantic_instance}
\resizebox{\textwidth}{!}{
\begin{tabular}{l l | CCCCC | CCCC}
\toprule
\textbf{Method} & \textbf{GT granularity}
& \textbf{VPQ$_0$} & \textbf{VPQ$_\infty$} & \textbf{STQ} & \textbf{AQ} & \textbf{GQ}
& \textbf{IP$_P$} & \textbf{IP$_G$} & \textbf{IC$_P$} & \textbf{IC$_G$} \\
\midrule

\rowcolor{ducklingyellow!30}
Savvy & Semantic
& 65.98 & 22.12 & 48.93 & 25.91 & 93.01
& 89.71 & 60.80 & 87.28 & 34.55 \\
\rowcolor{softcyan!30}
Savvy & Instance
& 55.52 & 28.21 & 44.50 & 21.53 & 93.01
& 67.52 & 53.95 & 80.84 & 39.28 \\

\rowcolor{ducklingyellow!30}
EntitySAM & Semantic
& 36.65 & 14.65 & 24.04 & 7.95 & 75.55
& 27.95 & 15.77 & 19.04 & 10.89 \\
\rowcolor{softcyan!30}
EntitySAM & Instance
& 18.71 & 2.90 & 10.28 & 1.52 & 75.55
& 10.16 & 8.63 & 27.76 & 18.99 \\

\rowcolor{ducklingyellow!30}
DEVA+SAM & Semantic
& 44.86 & 11.21 & 32.76 & 11.35 & 96.24
& 66.41 & 43.31 & 64.44 & 20.20 \\
\rowcolor{softcyan!30}
DEVA+SAM & Instance
& 32.66 & 9.36 & 26.15 & 7.26 & 96.24
& 33.12 & 25.70 & 78.08 & 29.56 \\

\bottomrule
\end{tabular}}
\end{table*}

\subsection{Additional Ablation Studies}
\label{app:additional_ablations}

The main paper reports module-level ablations on the 30-scene ScanNet minieval subset, including the HMD instantiation of MD, deferred admission, and track consolidation. Here, we provide additional sensitivity studies for system-level parameters and OGA metric thresholds. These experiments characterize performance across a range of system and evaluation settings.

The threshold sweeps use the 30-scene ScanNet minieval subset and the standard ScanNet hyperparameters, varying the threshold under study while retaining the other settings.

\paragraph{Savvy parameter sensitivity.}
Table~\ref{tab:savvy_param_ablations} studies four implementation parameters: discovery stride $s$, transient-buffer window $B$, competition margin $\delta$, and appearance self-consistency threshold $\tau_{\rm app}$. These parameters control different tradeoffs in the semi-online pipeline. The discovery stride determines how frequently HMD is invoked for new-object discovery; the transient window controls how long a newly discovered candidate must accumulate evidence before promotion; the competition margin controls how decisive the appearance-score gap must be before one established track suppresses another in overlap arbitration; and the appearance threshold controls how strict Savvy is when accepting long-gap reappearances.

For discovery stride, we evaluate $s\in\{1,3,5\}$. A smaller stride invokes HMD more frequently, which slightly improves short-window fidelity, STQ, AQ, and GQ. However, the default $s=3$ achieves the best VPQ$_\infty$, IP$_P$, and IC$_P$, while $s=5$ gives only a small further improvement on IC$_G$. This indicates that more frequent discovery is not always better: dense discovery can improve immediate coverage, but can also introduce more proposal pressure for deferred admission and consolidation. The default stride provides a good balance between discovery responsiveness and long-horizon identity stability.

For the transient window, we evaluate $B\in\{15,30,60\}$. The results are highly stable across this range. A shorter window ($B=15$) slightly improves VPQ$_0$, IP$_P$, and IC$_P$, suggesting faster admission and more responsive short-term discovery. A longer window ($B=60$) slightly improves VPQ$_\infty$, GQ, and IP$_G$, suggesting more conservative admission and stronger long-range reference-side support. The default $B=30$ remains close to the best across all metrics while avoiding the additional runtime cost of $B=60$, which is substantially slower in practice.

For competition arbitration, we vary the competition margin $\delta\in\{0,0.05,0.10,0.15,0.20\}$. The zero-margin setting is clearly worse on VPQ$_\infty$, STQ, and AQ, indicating that suppressing tracks based on negligible appearance-score differences can harm long-range consistency. Once a nonzero margin is used, performance becomes stable. The default $\delta=0.10$ is close to the best setting and obtains the strongest VPQ$_\infty$ among the tested margins, while larger margins yield only marginal changes. This supports the use of a conservative margin in competition arbitration: the system should suppress a competing track only when the evidence is sufficiently separated.

For appearance self-consistency, we vary $\tau_{\rm app}\in\{-\infty,0.0,0.1,0.2,0.4\}$, where $\tau_{\rm app}=-\infty$ disables the reappearance gate. Disabling the gate substantially lowers VPQ$_\infty$, STQ, AQ, IP$_G$, and IC$_G$, showing that long-gap appearance verification is important for preventing incorrect identity revival. In contrast, thresholds from $0.0$ to $0.2$ produce nearly identical results, and even $\tau_{\rm app}=0.4$ remains broadly stable with a small drop in AQ. This suggests that the benefit comes from having a reappearance consistency gate, rather than from a finely tuned threshold.

The module-level ablation in Table~\ref{tab:savvy_ablation} removes appearance self-consistency and competition arbitration together, whereas Table~\ref{tab:savvy_param_ablations} varies one consolidation subroutine while keeping the other active. These experiments characterize complementary aspects of the pipeline: module contributions and sensitivity to individual parameter changes. Without appearance gating, a reactivated track can still enter competition; with $\delta=0$, any appearance-score difference can trigger suppression. The observed drops support retaining both safeguards with conservative thresholds.

Overall, Table~\ref{tab:savvy_param_ablations} shows that Savvy is robust across reasonable parameter choices. The default configuration performs consistently near the top while balancing short-window coverage, long-window identity maintenance, and computational cost. The larger drops occur mainly when a mechanism is effectively disabled or made overly aggressive, such as using no appearance gate or zero competition margin.

\begin{table*}[t]
\centering
\small
\setlength{\tabcolsep}{8pt}
\caption{
\textbf{Savvy system-parameter sensitivity.}
We vary discovery stride, transient-buffer window, competition margin, and appearance self-consistency threshold.
Bold indicates the default configuration; shaded cells indicate the best score within each ablation group.
Savvy remains stable across reasonable parameter choices, while degenerate settings such as disabling appearance gating or using zero competition margin reduce long-window consistency.
}
\label{tab:savvy_param_ablations}
\resizebox{\linewidth}{!}{
\begin{tabular}{l c | ccccc | cccc}
\toprule
\textbf{Ablation} & \textbf{Value} 
& \textbf{VPQ$_0$} & \textbf{VPQ$_\infty$} & \textbf{STQ} & \textbf{AQ} & \textbf{GQ}
& \textbf{IP$_P$} & \textbf{IP$_G$} & \textbf{IC$_P$} & \textbf{IC$_G$} \\
\midrule
Disc. stride $s$ & 1 & \cellcolor{ac!15}59.15 & 30.42 & \cellcolor{ac!15}56.61 & \cellcolor{ac!15}34.95 & \cellcolor{ac!15}92.83 & 72.63 & \cellcolor{ac!15}58.74 & 60.27 & 31.02 \\
Disc. stride $s$ & \textbf{3} & 58.53 & \cellcolor{ac!15}31.30 & 55.97 & 34.77 & 91.59 & \cellcolor{ac!15}72.95 & 58.02 & \cellcolor{ac!15}61.16 & 31.61 \\
Disc. stride $s$ & 5 & 57.82 & 31.24 & 55.53 & 34.47 & 90.93 & 71.70 & 57.57 & 60.05 & \cellcolor{ac!15}32.02 \\
\midrule
Trans. win. $B$ & 15 & \cellcolor{ac!15}58.81 & 30.15 & 55.73 & 34.49 & 91.56 & \cellcolor{ac!15}73.07 & 57.63 & \cellcolor{ac!15}61.49 & 31.32 \\
Trans. win. $B$ & \textbf{30} & 58.53 & 31.30 & \cellcolor{ac!15}55.97 & \cellcolor{ac!15}34.77 & 91.59 & 72.95 & 58.02 & 61.16 & \cellcolor{ac!15}31.61 \\
Trans. win. $B$ & 60 & 58.38 & \cellcolor{ac!15}31.74 & 55.95 & 34.75 & \cellcolor{ac!15}91.66 & 73.02 & \cellcolor{ac!15}58.46 & 60.92 & 31.43 \\
\midrule
Comp. margin $\delta$ & 0    & \cellcolor{ac!15}58.56 & 26.13 & 54.78 & 33.42 & \cellcolor{ac!15}91.62 & 72.68 & 57.64 & 60.54 & 31.24 \\
Comp. margin $\delta$ & 0.05 & 58.54 & 30.65 & 55.78 & 34.54 & 91.60 & 72.91 & \cellcolor{ac!15}58.02 & 60.95 & 31.38 \\
Comp. margin $\delta$ & \textbf{0.10} & 58.53 & \cellcolor{ac!15}31.30 & 55.97 & 34.77 & 91.59 & 72.95 & 58.02 & 61.16 & 31.61 \\
Comp. margin $\delta$ & 0.15 & 58.55 & 31.20 & 55.95 & 34.74 & 91.58 & \cellcolor{ac!15}72.96 & 57.95 & \cellcolor{ac!15}61.18 & 31.61 \\
Comp. margin $\delta$ & 0.2 & 58.55 & 31.27 & \cellcolor{ac!15}55.99 & \cellcolor{ac!15}34.78 & 91.57 & 72.92 & 57.96 & 61.18 & \cellcolor{ac!15}31.62 \\
\midrule
Appr. threshold $\tau_{\rm app}$ & $-\infty$ & \cellcolor{ac!15}58.55 & 27.20 & 54.44 & 33.27 & \cellcolor{ac!15}91.61 & 72.71 & 57.18 & 60.38 & 31.09 \\
Appr. threshold $\tau_{\rm app}$ & 0.0 & 58.53 & 31.28 & 55.72 & 34.68 & 91.59 & 72.89 & 57.97 & 61.16 & 31.42 \\
Appr. threshold $\tau_{\rm app}$ & \textbf{0.1} & 58.53 & \cellcolor{ac!15}31.30 & \cellcolor{ac!15}55.97 & \cellcolor{ac!15}34.77 & 91.59 & \cellcolor{ac!15}72.95 & \cellcolor{ac!15}58.02 & 61.16 & \cellcolor{ac!15}31.61 \\
Appr. threshold $\tau_{\rm app}$ & 0.2 & 58.54 & 31.30 & 55.97 & 34.77 & 91.58 & 72.95 & 58.01 & 61.17 & 31.51 \\
Appr. threshold $\tau_{\rm app}$ & 0.4 & 58.53 & 31.29 & 55.66 & 34.03 & 91.57 & 72.94 & 58.01 & \cellcolor{ac!15}61.18 & 31.51 \\
\bottomrule
\end{tabular}
}
\begin{minipage}{0.98\linewidth}
\footnotesize
\emph{Note:} $\tau_{\rm app}=-\infty$ disables appearance self-consistency gating.
\end{minipage}
\vspace{-10pt}
\end{table*}

\paragraph{OGA metric sensitivity.}
Table~\ref{tab:oga_threshold_sensitivity} reports sweeps of prediction-support threshold $\tau_{\mathrm{ios}}$ and sever threshold $\tau_{\mathrm{sever}}$ over $\{0.3,0.5,0.7\}$ under this minieval setup. Primary fidelity is nearly unchanged across the IoS rows.

For framewise diagnostics, stricter IoS thresholds can remove weaker support relations under the IoU-or-IoS rule, affecting IC through the ever-support graph. They can also change the discovery factor $D$, and hence IP, while leaving the raw best-track intersection ratios unchanged. If $D=1$ remains saturated, IP is unchanged.

Severing changes how support chains split into dominant fragments. It leaves single-frame VPQ$_0$ and GQ unchanged, while potentially affecting VPQ$_\infty$, STQ, and AQ. Higher thresholds sever more identity transitions. Savvy's VPQ$_\infty$ decreases from 32.90 at $\tau_{\mathrm{sever}}=0.3$ to 31.30 at $0.5$ and 27.86 at $0.7$. Some IP/IC entries also differ across sever rows, although their definitions do not depend on the sever threshold. These differences cannot be attributed to severing alone.

EntitySAM's fidelity is nearly invariant across the sever rows. Savvy achieves the highest VPQ, STQ, and AQ across these comparisons.

\begin{table*}[t]
\centering
\small
\setlength{\tabcolsep}{6pt}
\caption{\small
\textbf{OGA metric-threshold sensitivity on the 30-scene ScanNet minieval subset.}
Runs use standard ScanNet settings while varying the indicated prediction-support threshold $\tau_{\mathrm{ios}}$ or sever threshold $\tau_{\mathrm{sever}}$.
Rows with $\tau=0.5$ correspond to the default configuration, and shaded cells mark deviation from the default.
}
\label{tab:oga_threshold_sensitivity}
\resizebox{\linewidth}{!}{
\begin{tabular}{l l c | ccccc | cccc}
\toprule
\textbf{Method} & \textbf{Threshold} & \textbf{Value} 
& \textbf{VPQ$_0$} & \textbf{VPQ$_\infty$} & \textbf{STQ} & \textbf{AQ} & \textbf{GQ}
& \textbf{IP$_P$} & \textbf{IP$_G$} & \textbf{IC$_P$} & \textbf{IC$_G$} \\
\midrule
EntitySAM 
& $\tau_{\mathrm{ios}}$ & 0.3 & 33.53 & 13.09 & 20.77 & 8.83 & 62.92 & \cellcolor{negdelta!22}27.29 & \cellcolor{negdelta!22}19.57 & \cellcolor{negdelta!22}9.50 & \cellcolor{negdelta!22}21.40 \\
& $\tau_{\mathrm{ios}}$ & 0.5 & 33.53 & 13.09 & 20.77 & 8.83 & 62.92 & 26.24 & 18.82 & 11.44 & 20.29 \\
& $\tau_{\mathrm{ios}}$ & 0.7 & 33.53 & 13.09 & 20.77 & 8.83 & 62.92 & \cellcolor{negdelta!22}24.88 & \cellcolor{negdelta!22}17.89 & \cellcolor{negdelta!22}11.71 & \cellcolor{negdelta!22}19.41 \\
\cmidrule{2-12}
& $\tau_{\mathrm{sever}}$ & 0.3 & 33.53 & 13.09 & 20.77 & 8.83 & 62.92 & 26.24 & 18.82 & 11.44 & 20.29 \\
& $\tau_{\mathrm{sever}}$ & 0.5 & 33.53 & 13.09 & 20.77 & 8.83 & 62.92 & 26.24 & 18.82 & 11.44 & 20.29 \\
& $\tau_{\mathrm{sever}}$ & 0.7 & 33.53 & 13.09 & 20.77 & 8.83 & 62.92 & 26.24 & 18.82 & 11.44 & 20.29 \\
\midrule
DEVA+SAM 
& $\tau_{\mathrm{ios}}$ & 0.3 & 44.71 & 20.50 & 44.46 & 21.70 & 96.37 & 63.19 & 51.52 & 49.98 & 25.67 \\
& $\tau_{\mathrm{ios}}$ & 0.5 & 44.71 & 20.50 & 44.46 & 21.70 & 96.37 & 63.19 & 51.52 & 49.98 & 25.67 \\
& $\tau_{\mathrm{ios}}$ & 0.7 & 44.71 & \cellcolor{negdelta!22}21.16 & \cellcolor{negdelta!22}44.51 & \cellcolor{negdelta!22}21.74 & 96.37 & \cellcolor{negdelta!22}60.71 & \cellcolor{negdelta!22}49.41 & \cellcolor{negdelta!22}52.97 & \cellcolor{negdelta!22}25.57 \\
\cmidrule{2-12}
& $\tau_{\mathrm{sever}}$ & 0.3 & 44.71 & \cellcolor{negdelta!22}21.16 & \cellcolor{negdelta!22}44.51 & \cellcolor{negdelta!22}21.74 & 96.37 & \cellcolor{negdelta!22}60.71 & \cellcolor{negdelta!22}49.41 & \cellcolor{negdelta!22}52.97 & \cellcolor{negdelta!22}25.57 \\
& $\tau_{\mathrm{sever}}$ & 0.5 & 44.71 & 20.50 & 44.46 & 21.70 & 96.37 & 63.19 & 51.52 & 49.98 & 25.67 \\
& $\tau_{\mathrm{sever}}$ & 0.7 & 44.71 & \cellcolor{negdelta!22}19.40 & \cellcolor{negdelta!22}44.11 & \cellcolor{negdelta!22}21.29 & 96.37 & \cellcolor{negdelta!22}60.71 & \cellcolor{negdelta!22}49.41 & \cellcolor{negdelta!22}52.97 & \cellcolor{negdelta!22}25.57 \\
\midrule
Savvy 
& $\tau_{\mathrm{ios}}$ & 0.3 & 58.53 & 31.30 & 55.52 & 34.77 & 91.59 & \cellcolor{negdelta!22}73.40 & \cellcolor{negdelta!22}58.36 & \cellcolor{negdelta!22}56.09 & \cellcolor{negdelta!22}28.70 \\
& $\tau_{\mathrm{ios}}$ & 0.5 & 58.53 & 31.30 & 55.52 & 34.77 & 91.59 & 72.95 & 58.02 & 61.16 & 31.61 \\
& $\tau_{\mathrm{ios}}$ & 0.7 & 58.53 & 31.30 & 55.52 & 34.77 & 91.59 & \cellcolor{negdelta!22}71.11 & \cellcolor{negdelta!22}56.64 & \cellcolor{negdelta!22}63.30 & \cellcolor{negdelta!22}33.23 \\
\cmidrule{2-12}
& $\tau_{\mathrm{sever}}$ & 0.3 & 58.53 & \cellcolor{negdelta!22}32.90 & \cellcolor{negdelta!22}55.92 & \cellcolor{negdelta!22}35.25 & 91.59 & 72.95 & 58.02 & 61.16 & \cellcolor{negdelta!22}31.51 \\
& $\tau_{\mathrm{sever}}$ & 0.5 & 58.53 & 31.30 & 55.52 & 34.77 & 91.59 & 72.95 & 58.02 & 61.16 & 31.61 \\
& $\tau_{\mathrm{sever}}$ & 0.7 & 58.53 & \cellcolor{negdelta!22}27.86 & \cellcolor{negdelta!22}54.60 & \cellcolor{negdelta!22}33.71 & 91.59 & 72.95 & 58.02 & 61.16 & \cellcolor{negdelta!22}31.51 \\
\bottomrule
\end{tabular}
}
\vspace{-10pt}
\end{table*}

\section{Synthetic Stress-Test Protocol and Results}
\label{app:stress_tests}
\vspace{-5pt}
The main paper reports synthetic stress tests for OGA under controlled perturbations. Here we provide the implementation-level definitions of these perturbations and the complete quantitative results. Each perturbation isolates a structural failure mode to test the selective response of OGA diagnostics. The perturbations are implemented directly on prediction label maps before evaluation, while the GT remains unchanged.

\textbf{Temporal dropout.}
Temporal dropout removes prediction support intermittently. For each predicted identity present in a frame, its pixels are independently replaced by the ignore label with probability $p_{\mathrm{drop}}$. This preserves the remaining identity labels but creates missing temporal support. As shown in Table~\ref{tab:oga_stress_tests}, dropout strongly reduces VPQ$_\infty$, STQ, AQ, and GQ as severity increases. This is expected: the perturbation removes visible support, so both foreground coverage and long-window association degrade. IP$_G$ also decreases because each reference object receives less persistent support from the prediction set.

\textbf{Macro sever.}
Macro sever simulates low-frequency permanent identity replacement. For each eligible track whose lifespan exceeds a minimum length, the track is split into large temporal chunks by scheduling one or more cut points. After each cut point, the original identity is remapped to a fresh identity for subsequent frames. This preserves spatial masks but breaks long-range identity continuity. In Table~\ref{tab:oga_stress_tests}, VPQ$_0$ and GQ remain unchanged under sever because the per-frame masks are preserved. In contrast, VPQ$_\infty$, STQ, AQ, IP$_G$, and IC$_G$ degrade sharply, showing that OGA detects the identity fragmentation even when spatial segmentation is unchanged.

\textbf{Sparse flicker.}
Sparse flicker simulates high-frequency identity pollution. For each eligible track, a small fraction of its frames are selected subject to a minimum temporal gap, and the object's ID is replaced by a fresh hallucinated ID only on those isolated frames. This preserves spatial masks while injecting short-lived identity fragments. The results show an even stronger separation between spatial fidelity and identity structure: VPQ$_0$ and GQ remain fixed, but VPQ$_\infty$, STQ, AQ, and IC$_G$ collapse rapidly. Notably, IP$_P$ and IC$_P$ increase under flicker and sever. This does not mean the prediction structure improves. Rather, the newly created short-lived IDs are often locally pure with respect to one reference object, so prediction-axis purity/concentration can appear high while the reference side degrades. The tandem behavior of IP$_P$/IC$_P$ with IP$_G$/IC$_G$ is therefore essential: high prediction-axis scores alone can indicate many pure fragments, whereas the GT-axis scores expose that each reference object is being split across many identities.

\textbf{Spatial clutter.}
Spatial clutter is implemented by dynamically dilating each predicted instance mask using area-dependent kernel sizes and iteration counts. In the aggressive settings, the dilated mask may overwrite other labeled regions or compete through majority voting among overlapping dilations. This perturbation targets spatial overreach and inter-object interference. Unlike sever or flicker, clutter changes the spatial masks themselves. Accordingly, Table~\ref{tab:oga_stress_tests} shows broad degradation across VPQ$_0$, VPQ$_\infty$, STQ, AQ, IP, and IC. Interestingly, GQ increases with clutter because more valid foreground pixels are covered by prediction foreground. This confirms that GQ alone is not sufficient: excessive spatial expansion can improve foreground coverage while harming identity structure and region fidelity.

\textbf{Void-only expansion.}
Void-only expansion applies the same dynamic dilation mechanism while restricting new pixels to ignored regions of the prediction map, preserving existing prediction labels. Valid-domain accounting excludes added pixels in GT-void regions, while extension into annotated foreground can increase GQ. VPQ$_0$, VPQ$_\infty$, and IC remain nearly stable as GQ increases, showing that this perturbation changes foreground coverage more strongly than identity structure.

\begin{table*}[t]
\centering
\small
\setlength{\tabcolsep}{6pt}
\caption{\small
\textbf{Stress test of OGA metrics across synthetic failure modes.}
We evaluate OGA under controlled perturbations at increasing severity levels.
The metrics respond selectively: dropout mainly removes support, sever and flicker preserve spatial masks but break identity continuity, clutter degrades spatial and identity structure broadly, and void-only expansion has limited negative effect under void-tolerant evaluation.
}
\label{tab:oga_stress_tests}
\resizebox{\linewidth}{!}{
\begin{tabular}{l c | CCCCC | CCCC}
\toprule
\textbf{Failure Mode} & \textbf{Severity} 
& \textbf{VPQ$_0$} & \textbf{VPQ$_\infty$} & \textbf{STQ} & \textbf{AQ} & \textbf{GQ}
& \textbf{IP$_P$} & \textbf{IP$_G$} & \textbf{IC$_P$} & \textbf{IC$_G$} \\
\midrule
Clutter 
& 1 & \cellcolor{ac!20}58.18 & \cellcolor{ac!20}30.96 & \cellcolor{ac!20}57.11 & \cellcolor{ac!20}35.36 & \cellcolor{ac!5}95.25 & \cellcolor{ac!20}72.34 & \cellcolor{ac!20}59.56 & \cellcolor{ac!20}60.62 & \cellcolor{ac!20}31.17 \\
& 2 & \cellcolor{ac!18}55.93 & \cellcolor{ac!19}29.05 & \cellcolor{ac!19}55.98 & \cellcolor{ac!18}33.49 & \cellcolor{ac!12}96.53 & \cellcolor{ac!18}71.34 & \cellcolor{ac!19}59.14 & \cellcolor{ac!19}59.93 & \cellcolor{ac!18}30.61 \\
& 4 & \cellcolor{ac!12}49.20 & \cellcolor{ac!14}23.25 & \cellcolor{ac!14}51.84 & \cellcolor{ac!13}28.40 & \cellcolor{ac!18}97.75 & \cellcolor{ac!14}68.52 & \cellcolor{ac!15}56.31 & \cellcolor{ac!14}57.58 & \cellcolor{ac!14}29.00 \\
& 8 & \cellcolor{ac!5}41.01 & \cellcolor{ac!5}11.65 & \cellcolor{ac!5}43.56 & \cellcolor{ac!5}20.63 & \cellcolor{ac!20}98.12 & \cellcolor{ac!5}62.88 & \cellcolor{ac!5}50.65 & \cellcolor{ac!5}52.88 & \cellcolor{ac!5}25.78 \\
\midrule
Dropout 
& 0.05 & \cellcolor{ac!20}56.32 & \cellcolor{ac!20}10.03 & \cellcolor{ac!20}44.26 & \cellcolor{ac!20}23.27 & \cellcolor{ac!20}87.01 & \cellcolor{ac!20}72.07 & \cellcolor{ac!20}54.52 & \cellcolor{ac!20}60.62 & \cellcolor{ac!20}31.30 \\
& 0.10 & \cellcolor{ac!15}54.13 & \cellcolor{ac!12}6.26 & \cellcolor{ac!14}39.20 & \cellcolor{ac!14}19.28 & \cellcolor{ac!15}82.81 & \cellcolor{ac!16}71.11 & \cellcolor{ac!16}51.25 & \cellcolor{ac!17}60.22 & \cellcolor{ac!16}30.97 \\
& 0.15 & \cellcolor{ac!11}51.81 & \cellcolor{ac!8}4.26 & \cellcolor{ac!10}35.11 & \cellcolor{ac!9}16.42 & \cellcolor{ac!10}78.23 & \cellcolor{ac!11}69.88 & \cellcolor{ac!10}47.41 & \cellcolor{ac!10}59.22 & \cellcolor{ac!8}30.22 \\
& 0.20 & \cellcolor{ac!5}49.16 & \cellcolor{ac!5}2.94 & \cellcolor{ac!5}30.99 & \cellcolor{ac!5}13.73 & \cellcolor{ac!5}73.26 & \cellcolor{ac!5}68.50 & \cellcolor{ac!5}43.54 & \cellcolor{ac!5}58.48 & \cellcolor{ac!5}29.91 \\
\midrule
Flicker 
& 0.025 & \cellcolor{ac!5}58.53 & \cellcolor{ac!20}2.72 & \cellcolor{ac!20}36.10 & \cellcolor{ac!20}14.54 & \cellcolor{ac!5}91.59 & \cellcolor{ac!5}74.65 & \cellcolor{ac!20}56.58 & \cellcolor{ac!5}75.22 & \cellcolor{ac!20}16.78 \\
& 0.050 & \cellcolor{ac!5}58.53 & \cellcolor{ac!9}0.71 & \cellcolor{ac!13}28.54 & \cellcolor{ac!12}9.13 & \cellcolor{ac!5}91.59 & \cellcolor{ac!12}75.16 & \cellcolor{ac!15}55.20 & \cellcolor{ac!13}78.28 & \cellcolor{ac!12}12.15 \\
& 0.075 & \cellcolor{ac!5}58.53 & \cellcolor{ac!7}0.38 & \cellcolor{ac!9}23.74 & \cellcolor{ac!9}6.77 & \cellcolor{ac!5}91.59 & \cellcolor{ac!16}75.42 & \cellcolor{ac!10}53.79 & \cellcolor{ac!16}79.39 & \cellcolor{ac!8}10.28 \\
& 0.100 & \cellcolor{ac!5}58.53 & \cellcolor{ac!5}0.05 & \cellcolor{ac!5}18.94 & \cellcolor{ac!5}4.05 & \cellcolor{ac!5}91.59 & \cellcolor{ac!20}75.68 & \cellcolor{ac!5}52.37 & \cellcolor{ac!20}80.91 & \cellcolor{ac!5}8.41 \\
\midrule
Sever 
& 1 & \cellcolor{ac!5}58.53 & \cellcolor{ac!20}13.11 & \cellcolor{ac!20}44.94 & \cellcolor{ac!20}22.54 & \cellcolor{ac!5}91.59 & \cellcolor{ac!5}73.84 & \cellcolor{ac!20}44.44 & \cellcolor{ac!5}68.93 & \cellcolor{ac!20}21.84 \\
& 2 & \cellcolor{ac!5}58.53 & \cellcolor{ac!11}7.03 & \cellcolor{ac!13}40.29 & \cellcolor{ac!14}17.97 & \cellcolor{ac!5}91.59 & \cellcolor{ac!13}74.33 & \cellcolor{ac!11}38.77 & \cellcolor{ac!11}72.08 & \cellcolor{ac!11}18.57 \\
& 3 & \cellcolor{ac!5}58.53 & \cellcolor{ac!8}5.09 & \cellcolor{ac!9}37.36 & \cellcolor{ac!11}15.53 & \cellcolor{ac!5}91.59 & \cellcolor{ac!18}74.59 & \cellcolor{ac!6}35.47 & \cellcolor{ac!15}74.31 & \cellcolor{ac!6}16.81 \\
& 4 & \cellcolor{ac!5}58.53 & \cellcolor{ac!5}3.15 & \cellcolor{ac!5}34.43 & \cellcolor{ac!5}11.29 & \cellcolor{ac!5}91.59 & \cellcolor{ac!20}74.72 & \cellcolor{ac!5}34.54 & \cellcolor{ac!20}76.62 & \cellcolor{ac!5}16.56 \\
\midrule
Void 
& 1 & \cellcolor{ac!20}58.64 & \cellcolor{ac!20}31.57 & \cellcolor{ac!5}57.57 & \cellcolor{ac!5}35.94 & \cellcolor{ac!5}95.25 & \cellcolor{ac!20}72.21 & \cellcolor{ac!5}59.89 & \cellcolor{ac!20}61.07 & \cellcolor{ac!20}31.51 \\
& 2 & \cellcolor{ac!18}58.46 & \cellcolor{ac!20}31.56 & \cellcolor{ac!10}58.15 & \cellcolor{ac!20}36.21 & \cellcolor{ac!11}96.53 & \cellcolor{ac!16}71.86 & \cellcolor{ac!9}60.70 & \cellcolor{ac!7}60.89 & \cellcolor{ac!19}31.50 \\
& 4 & \cellcolor{ac!13}57.99 & \cellcolor{ac!15}31.37 & \cellcolor{ac!14}58.49 & \cellcolor{ac!19}36.20 & \cellcolor{ac!16}97.75 & \cellcolor{ac!13}71.49 & \cellcolor{ac!13}61.42 & \cellcolor{ac!6}60.88 & \cellcolor{ac!15}31.41 \\
& 8 & \cellcolor{ac!5}57.28 & \cellcolor{ac!5}30.99 & \cellcolor{ac!20}59.17 & \cellcolor{ac!15}36.12 & \cellcolor{ac!20}98.53 & \cellcolor{ac!5}70.75 & \cellcolor{ac!20}62.86 & \cellcolor{ac!5}60.86 & \cellcolor{ac!5}31.23 \\
\bottomrule
\end{tabular}
}
\vspace{-10pt}
\end{table*}

\subsection{Simultaneous Spatial Fragmentation}
\label{app:spatial_fragmentation}
We additionally isolate simultaneous spatial fragmentation using Savvy predictions on the 30-scene ScanNet minieval split, without additional inference. Each predicted track is partitioned in every frame into $k\in\{2,4,8\}$ approximately equal-area stripes. The split orientation is fixed once per track using its video-level major image-space axis, and each stripe ordinal retains a persistent identity across frames. Stripe~0 keeps the original ID; the remaining stripes receive globally unique IDs. The foreground union is unchanged. Thus, this perturbation introduces concurrent spatial child identities rather than replacing an identity at temporal cut points. The $k=1$ baseline uses the unperturbed predictions and their reported scores.

\begin{table}[H]
\centering
\small
\setlength{\tabcolsep}{5pt}
\caption{\textbf{Persistent spatial fragmentation on ScanNet minieval.} Macro-averaged OGA scores ($\times100$) after splitting each prediction into $k$ persistent spatial stripes. Foreground support is identical across conditions.}
\label{tab:spatial_fragmentation}
\begin{tabular}{c r r r r r r r}
\toprule
$k$ & GQ & STQ & AQ & VPQ$_0$ & VPQ$_\infty$ & IP$_G$ & IC$_G$ \\
\midrule
1 & 91.59 & 55.52 & 34.77 & 58.53 & 31.30 & 58.02 & 31.51 \\
2 & 91.59 & 55.59 & 34.93 & 60.67 & 28.85 & 34.94 & 17.19 \\
4 & 91.59 & 55.84 & 35.26 & 62.25 & 26.49 & 19.66 & 8.63 \\
8 & 91.59 & 55.34 & 34.72 & 62.63 & 22.18 & 10.65 & 5.00 \\
\bottomrule
\end{tabular}
\end{table}

Table~\ref{tab:spatial_fragmentation} shows complementary responses to this perturbation. GQ is invariant, while STQ and AQ remain nearly unchanged. Short-window VPQ can increase because smaller child masks obtain favorable local IoS associations: VPQ$_0$ rises from 58.53 to 62.63. Reference-side IP$_G$ and IC$_G$ decrease from 58.02 and 31.51 to 10.65 and 5.00, respectively, exposing the distribution of each reference object's support across more identities. VPQ$_\infty$ also decreases, from 31.30 to 22.18. For this stripe construction, splitting can change child-tube IoS assignments, sever points, and fragment-level false positives even though the foreground union is unchanged.

Overall, the stress tests validate the diagnostic design of OGA. Spatial perturbations, missing-support perturbations, and identity-only perturbations produce different metric signatures. In particular, sever and flicker demonstrate why IP and IC must be read on both axes: prediction-side scores may remain high or even increase when many short-lived fragments are locally pure, while reference-side IP and IC reveal fragmentation of the underlying GT object. This selective behavior is the intended outcome of OGA: the metric suite does not collapse all failures into one score, but separates foreground coverage, long-horizon association, prediction-side purity, reference-side fragmentation, and void-region tolerance.

\begin{figure}[t]
    \centering
    \includegraphics[width=\linewidth]{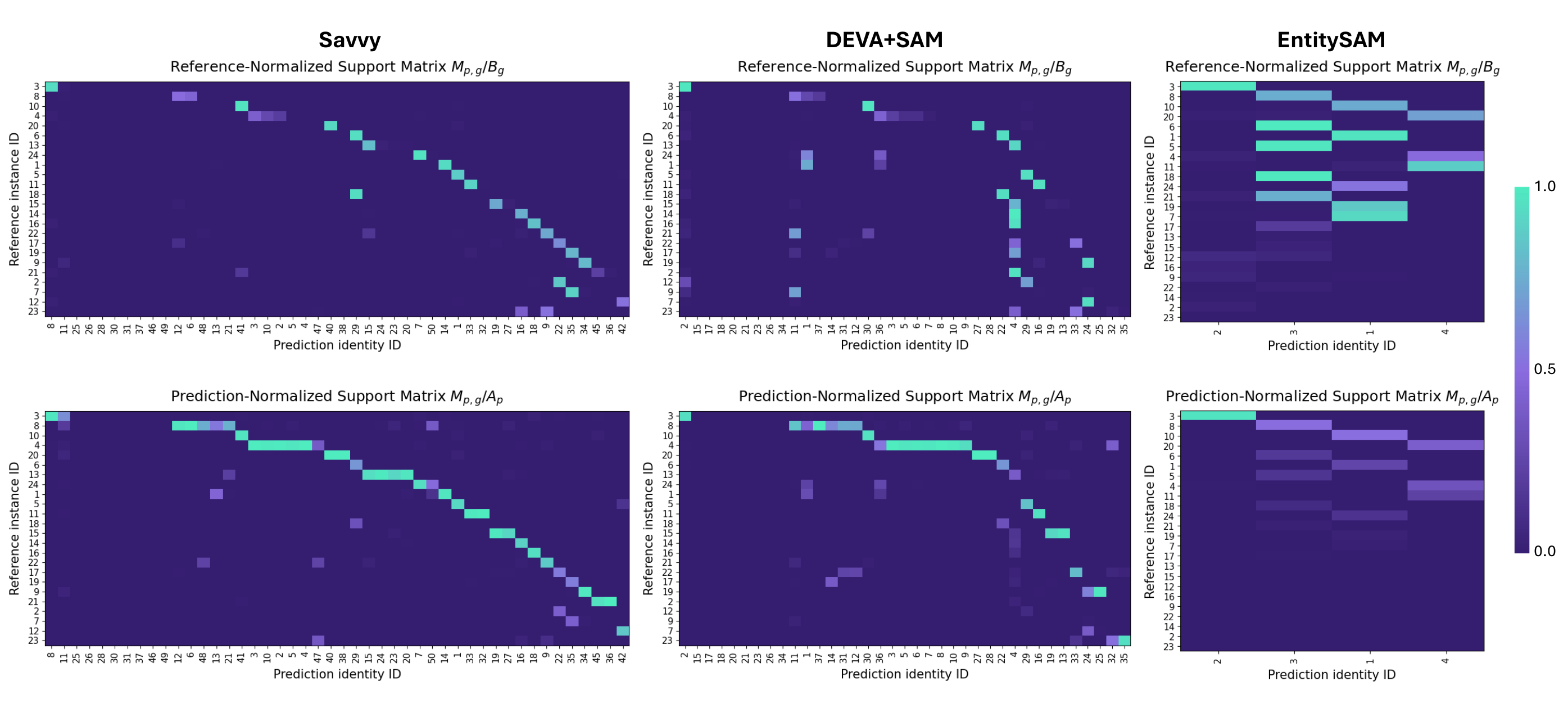}
    \caption{\small
\textbf{Prediction--reference support matrices for long-horizon object identity structure.}
We visualize sequence-level support between prediction identities and reference instances for Savvy, DEVA+SAM, and EntitySAM on ScanNet  \texttt{scene0019\_00}. The top row normalizes overlap mass by reference area, $M_{p,g}/B_g$, showing how each reference instance is covered by prediction identities. 
The bottom row normalizes by prediction area, $M_{p,g}/A_p$, showing how each prediction identity distributes its support over reference instances. 
Savvy produces a sparse, concentrated structure, indicating compact object identities with limited bleeding or fragmentation. 
DEVA+SAM shows a noisier pattern with more scattered off-diagonal support. 
EntitySAM exhibits a small number of broad prediction identities spanning many references, revealing under-discovery and identity merging despite apparently high local support. 
}
\label{fig:support_matrices}
\end{figure}

\section{OGA Visualization Protocol}
\label{app:oga_visualization_protocol}

Beyond scalar OGA metrics, we use two complementary visual diagnostics to inspect the prediction--reference support structure induced by each method. 
The support matrix provides a direct, instance-level view of how prediction identities overlap with reference objects within a sequence, making fragmentation, merging, and off-diagonal leakage visually explicit. 
The identity-concentration typology then summarizes the same structural behavior across scenes by placing each sequence in the $\mathrm{IC}_P$--$\mathrm{IC}_G$ space. 
Together, these visualizations help distinguish whether a method obtains strong scores through a compact and stable object set, or through fragmented, redundant, or entangled support patterns.

\subsection{Support Matrices}
\label{app:support_matrix_visualization}

We visualize the prediction--reference support matrix at the sequence level. 
For each video, we first accumulate the overlap mass between every prediction identity $p_i$ and every reference instance $g_j$:
\begin{equation}
    M_{i,j}
    =
    \sum_t |p_{i,t} \cap g_{j,t}|.
\end{equation}
Prediction pixels falling into void or unlabeled reference regions are ignored, consistent with the void-tolerant evaluation protocol used in our metrics. 
\begin{samepage}
We then visualize two normalized versions of this overlap matrix:
\begin{equation}
    W^{P}_{i,j}
    =
    \frac{M_{i,j}}{A_{p_i}+\varepsilon},
    \qquad
    A_{p_i}=\sum_t |p_{i,t}|,
\end{equation}
and
\begin{equation}
    W^{G}_{i,j}
    =
    \frac{M_{i,j}}{B_{g_j}+\varepsilon},
    \qquad
    B_{g_j}=\sum_t |g_{j,t}|.
\end{equation}
\end{samepage}
The prediction-normalized matrix $W^{P}$ measures the purity of each prediction identity: a clean prediction should concentrate its mass on a single reference instance. 
The reference-normalized matrix $W^{G}$ measures how each reference object is covered by predictions: a compact object representation should receive support from a small number of prediction identities. 
Thus, $W^{P}$ is more directly related to prediction-side identity bleeding, while $W^{G}$ is more directly related to reference-side coverage and fragmentation.

To make the matrix visually interpretable, we sort rows and columns according to the dominant reference assignment. 
Specifically, we first sort reference columns by their total overlap mass,
\begin{equation}
    s_j = \sum_i M_{i,j},
\end{equation}
in descending order. 
This places the most visible or most strongly supported reference instances earlier in the matrix. 
Given this ordered set of reference columns, each prediction identity $p_i$ is assigned a dominant reference index
\begin{equation}
    d(i)
    =
    \arg\max_j M_{i,j},
\end{equation}
where the index $j$ is taken after the reference-column ordering. 
Prediction rows are then sorted lexicographically by their dominant reference $d(i)$ and, within the same dominant reference group, by their dominant overlap mass
\begin{equation}
    m_i = \max_j M_{i,j},
\end{equation}
in descending order. 
This produces a block-diagonal layout when predictions are well organized: predictions explaining the same reference object are placed close together, and references with strong support appear as compact local structures.

This sorting is used only for visualization and does not affect the evaluation metrics. 
Its purpose is to expose structural patterns that are otherwise difficult to see under arbitrary identity numbering. 
A sparse near-diagonal pattern indicates compact and well-separated object identities. 
Tall columns indicate reference-side fragmentation, where one reference instance is split across multiple prediction identities. 
Wide rows indicate prediction-side bleeding, where one prediction identity supports multiple reference objects. 
Weak isolated off-diagonal entries usually correspond to minor overlap leakage, whereas dense off-diagonal bands or repeated horizontal structures indicate systematic identity merging.

In Fig.~\ref{fig:support_matrices}, we use the transposed reference--prediction layout, where rows correspond to reference instances and columns correspond to prediction identities.
\paragraph{Observed structures.}
Savvy produces a sparse and concentrated support structure in both normalizations, with most reference instances explained by compact prediction identities and limited off-diagonal support. 
DEVA+SAM preserves a partially diagonal structure but exhibits more scattered off-diagonal activations, suggesting noisier association and occasional identity leakage. 
EntitySAM shows a qualitatively different failure mode: only a small number of prediction identities support many reference instances, producing broad support shared across reference instances. 
This pattern indicates under-discovery and identity merging, where coarse prediction identities are shared across multiple objects rather than forming a persistent object-level set.

\subsection{Identity-Concentration Typology}
\label{app:ic_typology}

\begin{figure}[t]
    \centering\includegraphics[width=0.7\linewidth]{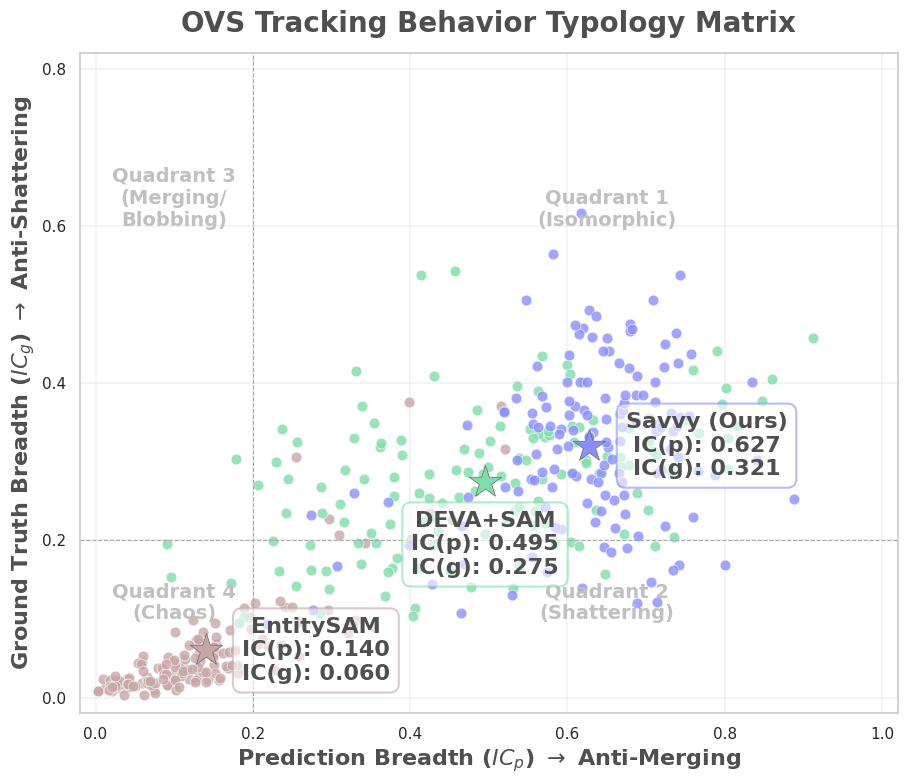}
    \caption{\small
\textbf{Identity-concentration typology on ScanNet.}
We visualize scene-level identity structure by plotting prediction-axis concentration ($\mathrm{IC}_P$) against reference-axis concentration ($\mathrm{IC}_G$). 
High $\mathrm{IC}_P$ indicates that prediction identities remain attached to a small number of reference objects, while high $\mathrm{IC}_G$ indicates that reference objects are supported by a compact set of prediction identities, spatiotemporally. 
Thus, the upper-right region represents compact, near-isomorphic support with limited prediction-side bleeding and limited reference-side fragmentation. 
Savvy consistently lies in this compact-support regime. 
DEVA+SAM shifts toward lower $\mathrm{IC}_G$, indicating stronger reference-side fragmentation. 
EntitySAM clusters in the low-concentration regime, consistent with under-discovery and broad prediction identities that induce entangled many-to-many support.
}
    \label{fig:ic_typology}
\end{figure}

The prediction--reference support matrix in Appendix~\ref{app:support_matrix_visualization} provides a direct visualization of the support structure for individual scenes. 
We further summarize this structure across scenes using an identity-concentration typology based on $\mathrm{IC}_P$ and $\mathrm{IC}_G$. 
This view is complementary to the support matrix: instead of showing the full many-to-many relation for one sequence, it places each evaluated scene into a two-dimensional behavior space.

The horizontal axis, $\mathrm{IC}_P$, measures prediction-axis concentration. 
A high $\mathrm{IC}_P$ indicates that each prediction identity is structurally attached to only a small number of reference instances, while a low value indicates prediction-side bleeding or merging, where the same prediction identity becomes valid support for multiple reference objects. 
The vertical axis, $\mathrm{IC}_G$, measures reference-axis concentration. 
A high $\mathrm{IC}_G$ indicates that each reference object is supported by a compact set of prediction identities, while a low value indicates reference-side fragmentation, where many prediction IDs participate in supporting the same object.

This yields a simple qualitative typology. 
The upper-right region corresponds to compact, near-isomorphic support: prediction identities remain object-specific, and reference objects are not split across many predictions. 
The lower-right region indicates reference-side fragmentation or shattering: prediction identities may remain relatively pure, but individual reference objects are supported by many IDs over time. 
The upper-left region indicates prediction-side bleeding or merging: reference objects may be compactly covered, but prediction identities are shared across multiple objects. 
The lower-left region corresponds to entangled many-to-many support, where both prediction-side and reference-side concentration are weak.

As shown in Fig.~\ref{fig:ic_typology}, Savvy occupies the high-concentration regime more consistently than the baselines. 
This indicates structurally compact support: predictions tend to remain attached to individual objects, and references tend to be explained by fewer prediction identities. 
DEVA+SAM shifts toward lower $\mathrm{IC}_G$, suggesting stronger reference-side fragmentation despite moderate prediction-side concentration. 
EntitySAM clusters in the low-concentration regime, consistent with under-discovery and coarse prediction identities that create entangled many-to-many support. 
Together with the support-matrix visualization, this analysis connects Savvy's fidelity and persistence gains to a more compact organization of object identities.

\clearpage
\section{Failure Cases and Limitations}
\label{app:failure_analysis}

Figures~\ref{fig:scannet_qualitative_1}--\ref{fig:scannet_qualitative_4} show long ScanNet sequences, while Figures~\ref{fig:hm3d_qualitative_home}--\ref{fig:hm3d_qualitative_livingroom} show HM3D indoor home, bedroom, and living-room sequences. Each detailed figure visualizes a single sequence over time, split into two vertical blocks from left to right. We compare the input video, GT, Savvy, DEVA+SAM, and EntitySAM.

Across both datasets, Savvy maintains a more detailed and temporally consistent object decomposition. It preserves separate object identities for furniture, fixtures, surfaces, and small scene objects as the camera moves through the environment and later revisits previously observed regions. DEVA+SAM often provides broad foreground coverage, but its masks frequently show local blobbing, boundary bleeding, and inconsistent splitting of nearby structures. EntitySAM produces coarser partitions with limited new-object discovery; existing identities often drift onto newly observed regions or absorb distinct objects into large scene-level masks.

These qualitative trends are consistent with the OGA diagnostics. High foreground coverage alone does not imply good open-world video segmentation. A method must also maintain a stable and expanding object set, avoid redundant or bleeding identities, and re-associate objects after disappearance and viewpoint change.

\subsection{Failure Cases}
\label{app:failure_cases}
\begin{figure*}[th!]
    \centering
    \includegraphics[width=\linewidth]{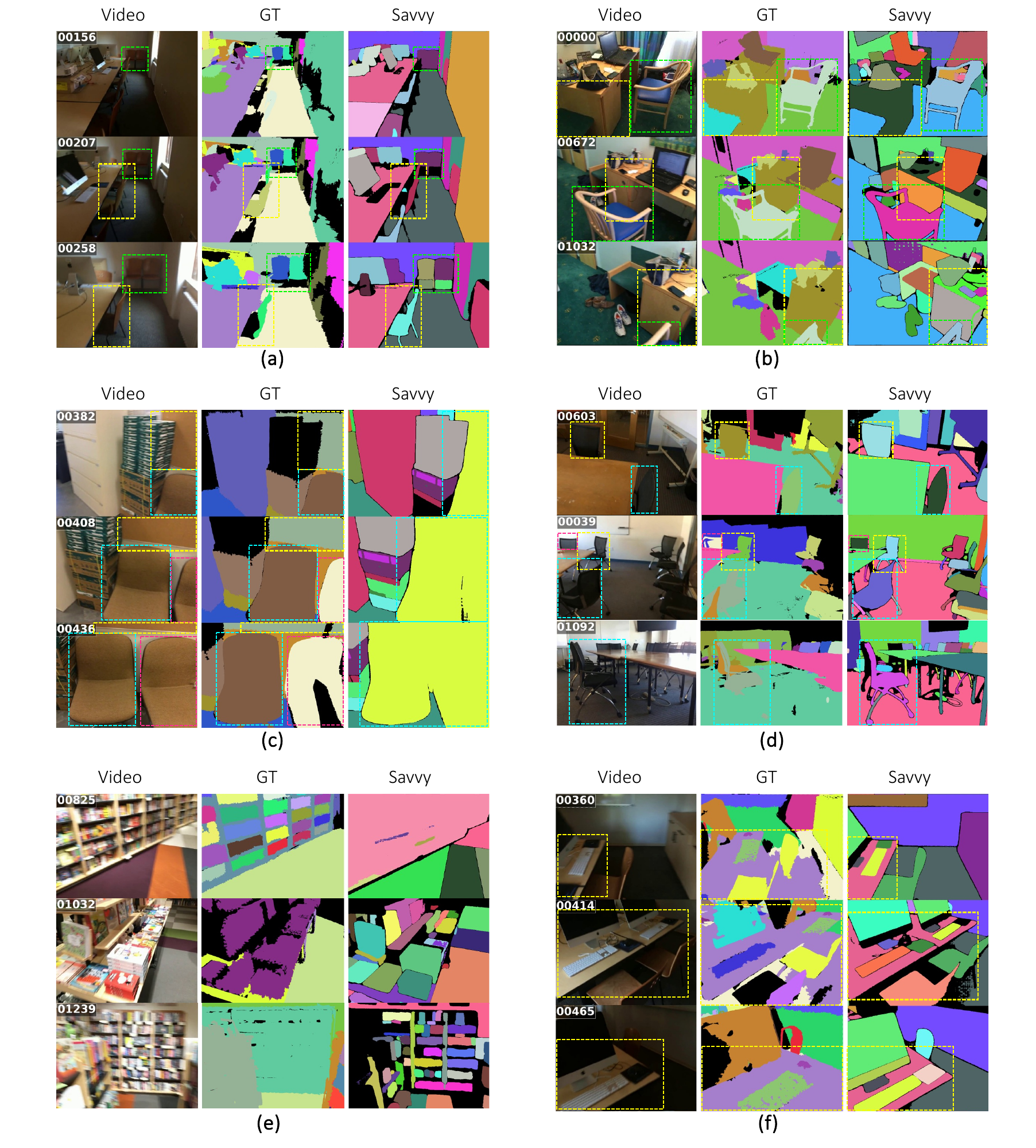}
    \caption{\small
\textbf{Representative failure cases and evaluation limitations of Savvy.}
Each panel shows input frames, GT, and Savvy predictions; dashed boxes mark the relevant regions.
The examples cover:
(a) granularity inconsistency under changing scale, where a chair-like object near the far end of an elongated room is represented at different granularities as the camera approaches;
(b) re-association failure after reappearance, where a previously observed object is segmented again but not linked back to its earlier identity;
(c) spatial identity reuse / blobbing, where visually similar nearby brown objects entering the view are merged into a shared predicted identity;
(d) identity confusion among repeated similar chairs, including both shared IDs across different GT chairs and split IDs for the same GT chair;
(e) over-clutter with motion blur, difficult lighting/reflections, and annotation-granularity inconsistency, where dense bookstore-like shelves induce unstable coarse-to-fine predictions and the GT itself changes granularity across views;
and (f) reference-side annotation limitations, where visually plausible object boundaries are not consistently reflected in the GT.
}
\label{fig:savvy_failure_cases}
\end{figure*}

Savvy reduces many common long-horizon OVS failures through deferred admission, appearance self-consistency, competition arbitration, and part-whole absorption. However, the setting remains challenging because the system must simultaneously decide \emph{what} to segment, \emph{at what granularity}, and \emph{whether a newly observed region should inherit an old identity}. Figure~\ref{fig:savvy_failure_cases} summarizes six representative failure modes. They fall into two broad groups: system-side failures, such as unstable granularity, missed re-association, local identity blobbing, and confusion among repeated similar objects; and evaluation/reference-side limitations, where dense clutter, motion blur, incomplete labels, or inconsistent annotation granularity make the notion of a single correct object decomposition ambiguous.

In Fig.~\ref{fig:savvy_failure_cases}, each panel shows input frames, GT, and Savvy predictions, with dashed boxes marking the relevant regions. 
The examples highlight recurring challenges in long-horizon open-world segmentation.

In Fig.~\ref{fig:savvy_failure_cases}(a), the camera moves through an elongated room and approaches a chair-like object near the far end. 
As the object becomes larger and clearer in view, Savvy changes its decomposition of the highlighted region, alternating between a larger local structure and a finer object-level mask. 
This illustrates granularity inconsistency under changing scale and viewpoint: the object becomes easier to segment, but the maintained representation is not always stable across the approach.

Figure~\ref{fig:savvy_failure_cases}(b) shows a re-association failure after reappearance. 
The highlighted chair-like object is visible, disappears or undergoes a large viewpoint change, and later re-enters the scene. 
Savvy segments the object reasonably in the later frames, but it may not recover the original identity. 
This is the failure mode targeted by appearance self-consistency and track consolidation, but large viewpoint change and occlusion can still break long-gap identity linkage.

Figure~\ref{fig:savvy_failure_cases}(c) shows spatial identity reuse/blobbing within a frame. 
This is different from long-range temporal ID reuse across unrelated object categories, which we observe rarely. 
Here, several visually similar brown objects enter the view together and are spatially close; Savvy can blob them into a shared predicted identity. 
The failure is therefore better understood as local spatial merging under appearance ambiguity, rather than as a frequent temporal reuse of one ID on different object types.

Figure~\ref{fig:savvy_failure_cases}(d) shows identity confusion among repeated similar chairs. 
The highlighted regions contain multiple office chairs with similar appearance, shape, and material. 
Compared with the GT, Savvy exhibits both directions of identity inconsistency: in the upper frames, different GT chairs can receive the same predicted identity, while in the middle and lower frames, the same GT chair can be split into different predicted identities as the viewpoint changes. 
This reflects the difficulty of maintaining instance-level identities among repeated same-category objects under ego-motion, occlusion, and changing scale.

Figure~\ref{fig:savvy_failure_cases}(e) shows a compound failure in a bookstore-like scene. 
The input frames contain dense shelves, many small visually separable items, and motion blur. 
Savvy alternates between coarse shelf-level regions and fine book/item-level predictions, producing granularity inconsistency. 
At the same time, the GT annotation itself changes granularity across frames: some views are annotated more like shelf/display regions, while others expose finer structures. 
This case is therefore both a system-side challenge and a reference-side annotation challenge. 
Additional scene-level factors, including uneven lighting, reflections from glossy surfaces, mirror-like structures, and occasional overexposure, further make stable object discovery and consistent granularity selection difficult. 

Finally, Fig.~\ref{fig:savvy_failure_cases}(f) illustrates reference-side annotation limitations. 
The highlighted table/desk regions contain visually meaningful boundaries and objects that are not always reflected consistently in the GT. 
Savvy may produce plausible object boundaries that disagree with incomplete or coarse annotations. 
OGA mitigates these disagreements through void handling and $n{:}1$ support; any remaining ambiguity stems from the available reference rather than necessarily from a tracking error.

These examples suggest several directions for future work: stronger long-gap re-identification, more stable granularity selection, explicit reasoning over repeated similar objects, and adaptive suppression of overly fine discovery in highly cluttered regions.

\subsection{Limitations}
\label{app:limitations}

Savvy combines persistent discovery with identity consolidation over long videos. The following discussion identifies remaining system and computational challenges, clarifies OGA's evaluation scope, and outlines complementary benchmark regimes.

\paragraph{System limitations.}
Savvy's explicit object set supports persistent discovery and re-association; maintaining stable granularity and identity across changing views remains challenging. First, object granularity is not always stable. The same physical region may be represented as a whole object in one view and as several parts in another, especially when the camera moves from a far view to a close view and the visual evidence becomes more detailed. HMD and part-whole absorption reduce this instability, but they do not fully solve the problem of choosing a single persistent granularity in open-world scenes.

Second, long-gap re-association remains difficult. When an object disappears for many frames and later reappears under a different scale, viewpoint, lighting condition, or occlusion pattern, appearance self-consistency may reject the reappearance, or the system may create a new identity instead of linking it back to the old one. Conversely, in rare cases, a weak old identity can be revived on an incorrect nearby object. This is especially challenging for repeated same-category objects, such as many similar chairs in a conference room, where local appearance cues alone are often insufficient for reliable identity recovery.

Third, Savvy can still produce local identity blobbing. Although we rarely observe long-range temporal reuse of the same identity across clearly different object types, spatially adjacent objects with similar appearance can be merged into one predicted identity within a frame. This occurs most often when multiple similar objects enter the view together, when object boundaries are weak, or when the propagated masks have already become uncertain.

Fourth, very cluttered scenes remain challenging. In scenes such as bookshelves, stores, or dense tabletop regions, the image may contain hundreds of visually separable items. Savvy may discover these at a finer granularity than the reference annotation, leading to excessive object identities and unstable fine-scale fragments. Suppressing all such fine predictions would lose potentially valid open-world objects, while admitting all of them can produce an overly dense object set. This exposes a deeper unresolved problem: open-world segmentation needs mechanisms for adaptive granularity selection, not merely better mask propagation.

\paragraph{Computational limitations.}
Savvy's computation supports repeated propagation of an active object set, image-level discovery, transient buffering, appearance-memory updates, and track consolidation. Memory pruning and active-track control make long videos feasible, but the pipeline remains sensitive to the number of active objects and the frequency of discovery. Larger transient buffers and more frequent HMD invocation can improve some metrics, but they also increase runtime. These costs motivate more efficient discovery scheduling and active-state management.

\paragraph{Evaluation scope.}
OGA is designed to relax annotation granularity while preserving diagnostic control. Within each GA scoring scope, each prediction is assigned to at most one reference instance, while each reference may receive support from multiple predictions. This asymmetric assignment permits part-to-whole support without duplicating a prediction's fidelity credit across references; it does not constrain IP's raw intersections or IC's framewise support relations. OGA therefore relaxes granularity in the controlled $n{:}1$ direction rather than treating arbitrary $1{:}n$ over-grouping as equally valid. Two-sided diagnostics separately reveal prediction-side identity bleeding.

\paragraph{Benchmark-scope limitations.}
Our benchmark suite covers standard video panoptic segmentation through VIPSeg and indoor scene-centric segmentation through ScanNet and HM3D. Quantitative ScanNet and HM3D results characterize object maintenance over adaptively sampled prefixes of up to 1,500 source frames. Longer evaluated spans and outdoor videos with moving cameras and dynamic traffic, such as KITTI- or Waymo-style settings, provide complementary directions for future work.

\paragraph{Future directions.}
The limitations above suggest several concrete directions. First, stronger long-gap re-identification is needed, potentially using more view-invariant object descriptors, geometric memory, or explicit scene-level localization reasoning. Second, open-world segmentation needs better granularity stabilization: the system should decide when to preserve part-level identities, when to absorb them into whole objects, and when to suppress overly fine discovery in dense clutter. Third, repeated similar objects require stronger relational reasoning beyond local appearance, such as spatial layout, object permanence, and scene-context constraints. Finally, evaluation itself should continue to evolve toward protocols that represent multiple valid object decompositions and long-horizon identity structure without collapsing all errors into a single overlap score.

\clearpage
\section{Out-of-Domain Stress Test on Egocentric Activity Video}
\label{app:ood_epickitchens}

\paragraph{Setup.}
Our primary long-horizon benchmarks, ScanNet and HM3D, emphasize static indoor scenes with camera motion, repeated re-observation, and changing viewpoint. To probe whether Savvy's object-lifecycle mechanisms remain informative outside this regime, we run an out-of-domain stress test on egocentric activity video from EPIC-KITCHENS~\citep{epickitchens}. This setting adds hand-object manipulation, deforming hands, rapid head motion, motion blur, partial near-field crops, and frequent short-cycle revisits. Savvy is applied with frozen hyperparameters, using the same indoor configuration summarized in Table~\ref{tab:savvy_hyperparams}, with no per-domain retuning. The clips are sampled at $10$ FPS and downsampled to $5$ FPS for model input; the qualitative figures retain the $10$ FPS clip indices and timestamps. We characterize transfer through qualitative tracking examples and annotation-free diagnostics; the clips lack exhaustive long-horizon instance annotations for this setting.

\begin{figure*}[t]
  \centering
  \includegraphics[width=0.98\linewidth]{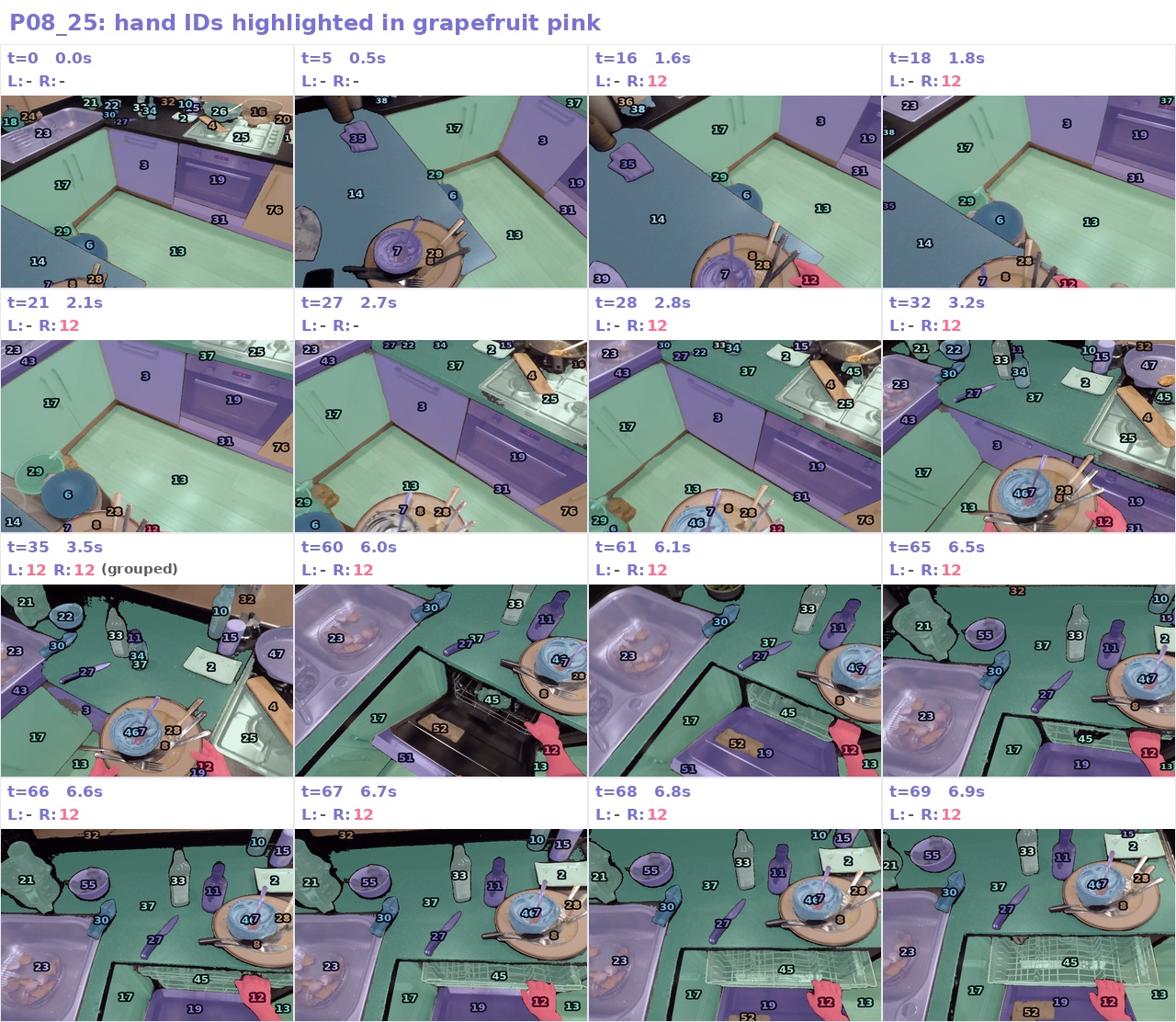}
  \caption{\small
  \textbf{Out-of-domain stress test on EPIC-KITCHENS: relatively clean appearance.}
  In \texttt{P08\_25}, the right hand retains identity~\texttt{12} from $t{=}16$ to $t{=}69$ despite camera motion and background change, as long as the hand remains relatively rigid and in focus.
  The \texttt{L}/\texttt{R} labels are Savvy identities read off by manual inspection, not ground truth labels.
  The short grouping event at $t{=}35$ illustrates the nearby-instance confusion analyzed in Fig.~\ref{fig:ood_epickitchens_failure}.}
  \label{fig:ood_epickitchens_success}
\end{figure*}

\begin{figure*}[t]
  \centering
  \includegraphics[width=0.98\linewidth]{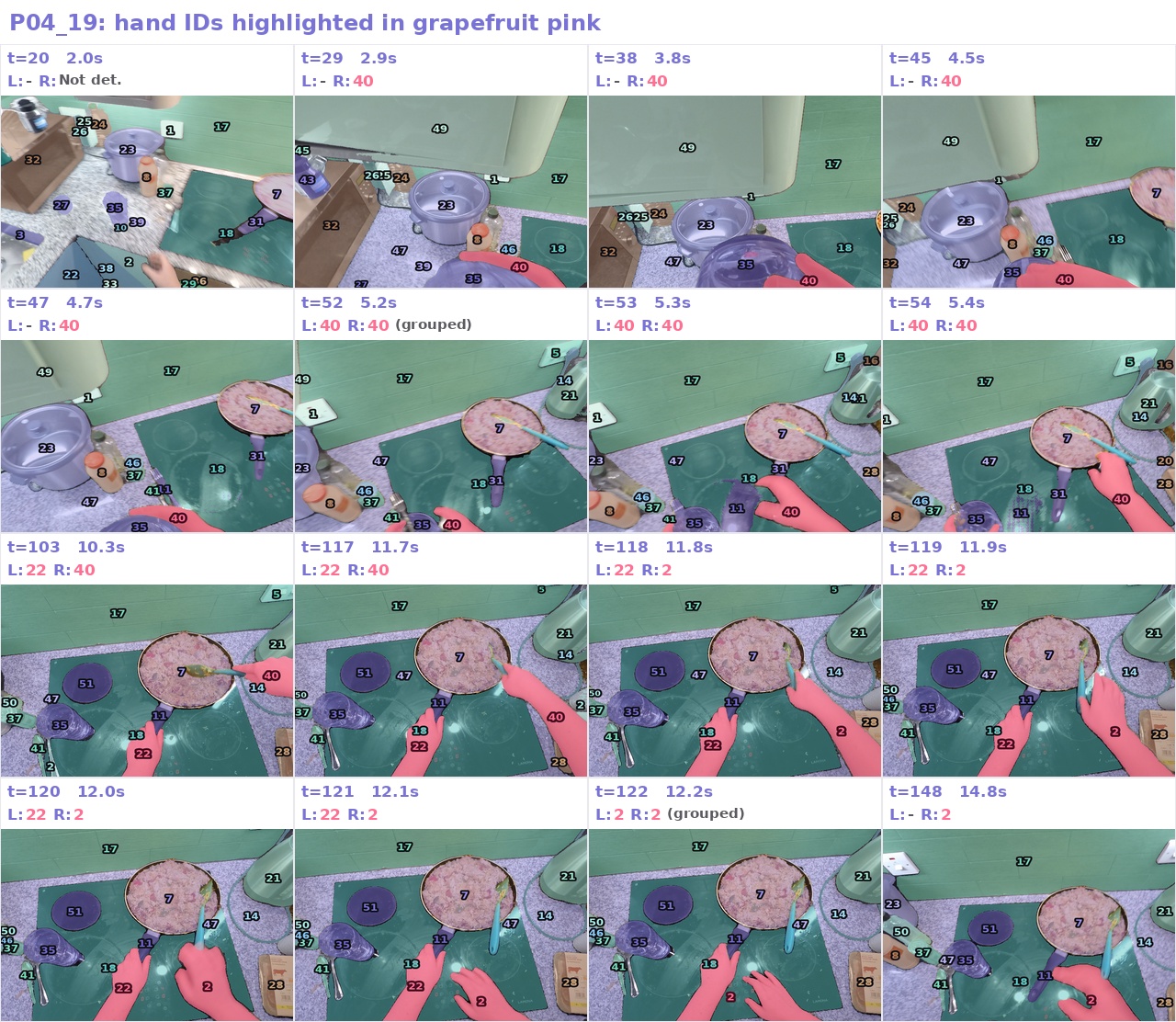}
  \caption{\small
  \textbf{Out-of-domain stress test on EPIC-KITCHENS: representative failure case.}
  In \texttt{P04\_19}, the right hand is tracked as identity~\texttt{40} from $t{=}29$ to $t{=}117$, then switches to identity~\texttt{2} at $t{=}118$ while remaining visible.
  The clip also shows a two-hand grouping event at $t{=}122$.
  These examples exhibit identity switching and nearby-instance grouping, extending the failure patterns in Fig.~\ref{fig:savvy_failure_cases} to egocentric activity video.}
  \label{fig:ood_epickitchens_failure}
\end{figure*}

We focus on hands because they are among the most challenging instances in this domain: they deform continuously, are often motion-blurred, appear near the camera boundary, and can become spatially adjacent with nearly identical appearance when both hands interact. In Figs.~\ref{fig:ood_epickitchens_success} and~\ref{fig:ood_epickitchens_failure}, hand masks are highlighted in pink, and the per-panel \texttt{L}/\texttt{R} labels denote Savvy identities read off by manual inspection, not ground truth hand labels.

\paragraph{Transfer under relatively clean appearance.}
Figure~\ref{fig:ood_epickitchens_success} shows a representative case from \texttt{P08\_25}. The right hand keeps identity~\texttt{12} from $t{=}16$ to $t{=}69$ across large camera motion and a changing background when the hand remains comparatively rigid and in focus. This suggests that the same appearance self-consistency and track-consolidation mechanisms that support stable indoor objects can also sustain hand identity out of domain when their input evidence remains stable. The brief two-hand grouping at $t{=}35$ illustrates the nearby-instance grouping mode discussed below.

\paragraph{Observed failures align with known modes.}
Figure~\ref{fig:ood_epickitchens_failure} shows a more difficult case from \texttt{P04\_19}. The right hand is tracked as identity~\texttt{40} over a long interval, but later switches to identity~\texttt{2} between $t{=}117$ and $t{=}118$ while the hand remains visible. At this transition, the hand exhibits blur, deformation, and articulation, and its predicted identity changes despite continuous visibility. This extends the identity-consistency challenges in Fig.~\ref{fig:savvy_failure_cases}(b,e) to continuously visible, deforming hands in egocentric activity video. The same clip also shows a two-hand grouping event at $t{=}122$, where the hands collapse into one identity. This is consistent with the local blobbing and similar-instance confusion in Fig.~\ref{fig:savvy_failure_cases}(c,d), now induced by adjacent hands with similar appearance.

\begin{figure*}[t]
  \centering
  \includegraphics[width=\linewidth]{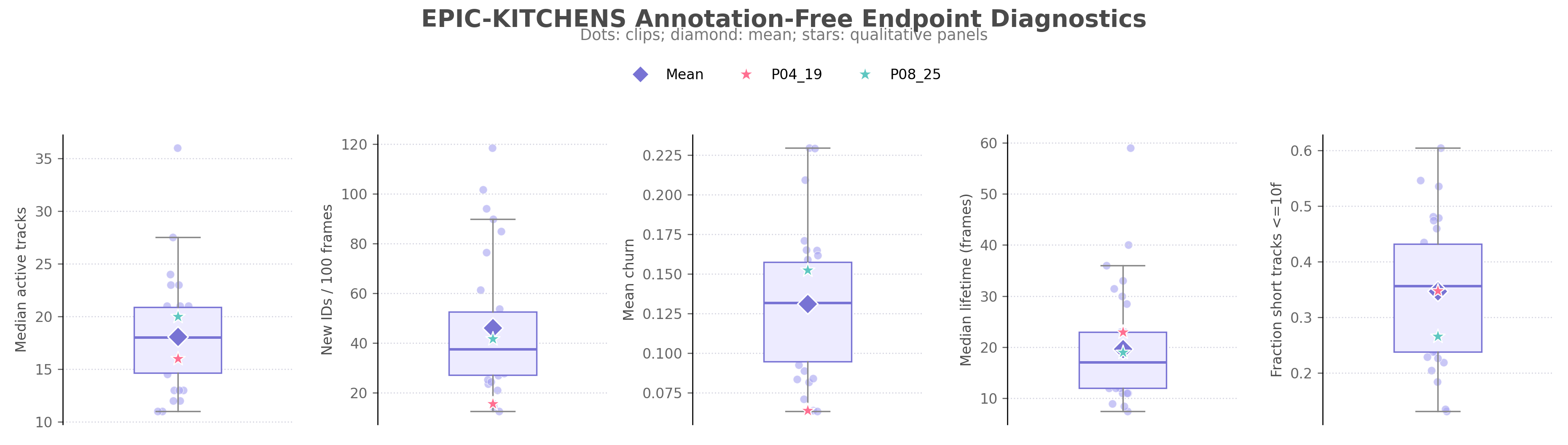}
  \caption{\small
  \textbf{Annotation-free endpoint diagnostics on EPIC-KITCHENS.}
  Each dot is a clip; the diamond marks the mean; stars mark the two qualitative sequences.
  Across $30$ clips from $12$ participants, the active set remains bounded while identities accumulate at a low, roughly steady rate.
  This profile is consistent with a mixture of benign granularity variation and genuine re-association or grouping errors, with consolidation and active-track control preventing unbounded active-set growth.}
  \label{fig:ood_epickitchens_diagnostics}
\end{figure*}

\paragraph{Annotation-free aggregate diagnostics.}
We also compute annotation-free diagnostics over $30$ clips from $12$ participants (Fig.~\ref{fig:ood_epickitchens_diagnostics}). These diagnostics characterize active-state usage and cumulative identity growth. The active-track count remains in a bounded range, while cumulative identities grow at a low, approximately steady rate. This pattern is consistent with the runtime behavior observed on ScanNet in Appendix~\ref{app:runtime_memory}: memory pruning, active-track control, and consolidation keep the active set from compounding even as new identities are introduced over time.

Cumulative identity growth can reflect new-object discovery, granularity variation, and identity errors such as re-association failures or local grouping and splitting. These annotation-free diagnostics describe their aggregate effect; distinguishing their relative contributions requires instance annotations. The bounded active-track count shows that cumulative identities can grow without corresponding growth in the active predictor set.

\paragraph{Takeaway.}
These out-of-domain examples suggest that Savvy's lifecycle mechanisms can remain useful on egocentric activity video, but they also expose the same system-side limitations discussed in Appendix~\ref{app:limitations}: granularity instability, re-association difficulty, identity leakage, and confusion among nearby similar instances. These qualitative results motivate linking part- and whole-level masks across granularity levels so that different views of the same physical object are related rather than accumulated as unrelated identities. Such grouping would complement stronger re-identification and granularity stabilization in highly dynamic open-world videos.